\documentclass[oneside,final, letterpaper]{ucr}

\begin{document}

\widowpenalty10000
\clubpenalty10000

\title{Towards a Near Universal Time Series Data Mining Tool:\\Introducing the Matrix Profile}
\author{Chin-Chia Michael Yeh}
\degreemonth{September}
\degreeyear{2018}
\degree{Doctor of Philosophy}
\chair{Dr.  Eamonn Keogh}
\othermembers{Dr. Stefano Lonardi\\
Dr. Evangelos E. Papalexakis\\
Dr. Christian R. Shelton}
\numberofmembers{4}
\field{Computer Science}
\campus{Riverside}

\maketitle
\copyrightpage{}
\approvalpage{}

\degreesemester{Summer}

\begin{frontmatter}

\begin{acknowledgements}
I would like to take this opportunity to express my deepest gratitude to my adviser, Dr. Eamonn Keogh, for his invaluable guidance, mentorship, and generous support throughout my doctoral study.
I am extremely fortunate and grateful to have received guidance on both data mining and general scientific research from such a brilliant professor.
Moreover, he gracefully encouraged me to pursue my own interests.
I am also very grateful to my committee members, Dr. Stefano Lonardi, Dr. Evangelos E. Papalexakis, and Dr. Christian R. Shelton for their generous advice in compiling this thesis, and Dr. Sara C. Mednick for being a committee member in my oral qualifying exam.

Additionally, I express gratitude to the wonderful colleagues I came across during my time at the UCR Data Mining Lab for their invaluable support and friendship:
Alireza Abdoli,
Dau Hoang Anh,
Dr. Nurjahan Begum,
Yifei Ding,
Shaghayegh Gharghabi,
Shima Imani,
Kaveh Kamgar,
Frank Madrid,
Nader S. Senobari,
Dr. Mohammad Shokoohi-Yekta,
Dr. Diego F. Silva,
Dr. Shailendra Singh,
Dr. Liudmila Ulanova,
Yan Zhu,
Zachary Zimmerman,
and alumni of the lab whom I frequently ran into at conferences: Dr. Jessica Lin and Dr. Abdullah Mueen.

Finally, I am grateful to my mentor, Dr. Yi-Hsuan Yang, during my time as a Research Assistant at Academia Sinica, Taiwan.
His guidance and support through the initial phase of my research career was important and invaluable. 
I am also thankful for the friends I made during this period:
Chih-Ming Chen,
Dr. Tak-Shing Thomas Chan,
Szu-Yu Chou,
Ping-Keng Jao,
Hsin-Ming Lin,
Sung-Yen Liu,
Dr. Jen-Yu Liu,
Dr. Li Su,
Yuan-Ching Teng,
Dr. Ju-Chiang Wang,
and Li-Fan Yu

\end{acknowledgements}

\begin{dedication}
\null\vfil
{\large
\begin{center}
To my dear and supportive wife, Shih-Chen Kuo.
\end{center}}
\vfil\null
\end{dedication}

\begin{abstract}
The last decade has seen a flurry of research on \emph{all-pairs-similarity-search} (or, self-join) for text, DNA, and a handful of other datatypes, and these systems have been applied to many diverse data mining problems. 
Surprisingly, however, little progress has been made on addressing this problem for \emph{time series subsequences}.
In this thesis, we have introduced a near universal time series data mining tool called \emph{matrix profile} which solves the all-pairs-similarity-search problem and caches the output in an easy-to-access fashion.
The proposed algorithm is not only \emph{parameter-free}, \emph{exact} and \emph{scalable}, but also applicable for both single and multidimensional time series.
By building time series data mining methods on top of matrix profile, many time series data mining tasks (e.g., motif discovery, discord discovery, shapelet discovery, semantic segmentation, and clustering) can be efficiently solved.
Because the same matrix profile can be shared by a diverse set of time series data mining methods, matrix profile is \emph{versatile} and \emph{computed-once-use-many-times} data structure.
We demonstrate the utility of matrix profile for many time series data mining problems, including motif discovery, discord discovery, weakly labeled time series classification, and representation learning on domains as diverse as seismology, entomology, music processing, bioinformatics, human activity monitoring, electrical power-demand monitoring, and medicine.
We hope the matrix profile is not the \emph{end} but the \emph{beginning} of many more time series data mining projects.
\end{abstract}

\tableofcontents
\listoffigures
\listoftables
\end{frontmatter}


\chapter{Introduction}
Data mining is the process of discovering and extracting knowledge from data~\cite{aggarwal2015data, han2011data}.
Data mining practitioners tend to use fast unsupervised methods in the early stages of data mining process.
For example, motif discovery, discord discovery, clustering, and segmentation are widely used in mining time series data~\cite{chandola2009anomaly, gharghabi2017icdm, gharghabi2018dmkd, mueen2009exact, keogh2005hot, ulanova2016clustering}.
If we can pre-compute and store certain information that can be used across all aforementioned tasks, it will greatly expedite the data mining process.
In this thesis, we propose a near universal time series data mining tool called \emph{matrix profile}\footnote{Matrix profile is formally defined in Chapter~\ref{ch2}.}~\cite{yeh2016icdm, yeh2017dmkd}.
The matrix profile computes and stores the all-pairs-similarity-search information in an efficient and easy-to-access fashion, and this information can be used in a variety of data mining tasks ranging from well-defined tasks (e.g., motif discovery) to more open-ended tasks (e.g., representation learning).

\begin{figure}[htb]
\centering
\includegraphics[trim={8cm 6.5cm 8cm 6.5cm}, clip, width=1\textwidth,page=1]{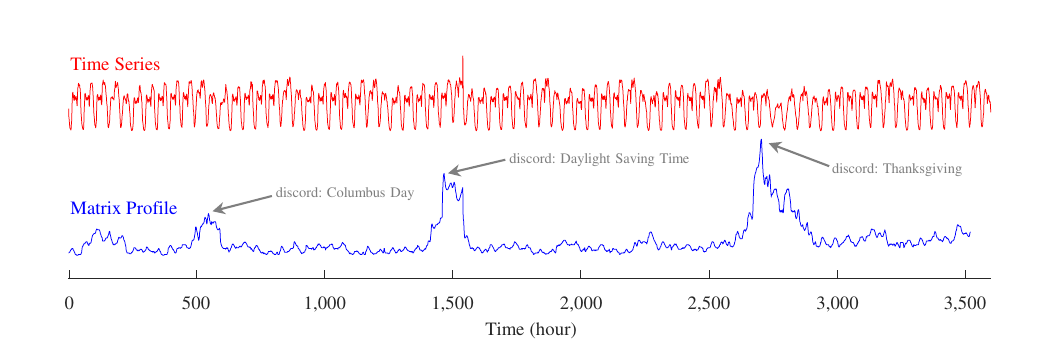}
\caption{
\emph{top}) The hourly average of the number of NYC taxi passengers over 75 days in Fall of 2014~\cite{lavin2015evaluating, rong2017asap}.
\emph{bottom}) The associated matrix profile reveals discords/anomalies from the data.
}
\label{ch1figtaxi}
\end{figure}

To give the reader a more concrete idea about what we are trying to achieve with matrix profile, consider the time series shown in Figure~\ref{ch1figtaxi}.\emph{top}.
Imaging an analyst is asked to examine the time series for ``interesting'' knowledge in the taxi ridership time series~\cite{lavin2015evaluating, rong2017asap}.
Without any domain knowledge, the analyst may try to identify the time series motifs using MK algorithm~\cite{mueen2009exact}.
Despite the returned time series motifs being well-conserved, the result does not tell an interesting story about the data as well-conserved patterns are presented in almost all of the data.
Time series discords~\cite{chandola2009anomaly, keogh2005hot} (i.e., unique patterns) should be mined instead of time series motifs.
By visually examining the matrix profile (Figure~\ref{ch1figtaxi}.\emph{bottom}), the analyst can quickly realize that he or she should look for time series discords as the matrix profile values can be interpreted as the uniqueness for time series subsequences~\cite{yeh2016icdm, yeh2017dmkd}.
Even if the analyst made the \emph{wrong} decision in his or her initial analysis (by choosing to examine time series motifs), the analyst can quickly locate the time series discords without much (computational) effort because the matrix profile is already computed.

\begin{figure}[htb]
\centering
\includegraphics[trim={8cm 5.5cm 8cm 5.5cm}, clip, width=1\textwidth,page=3]{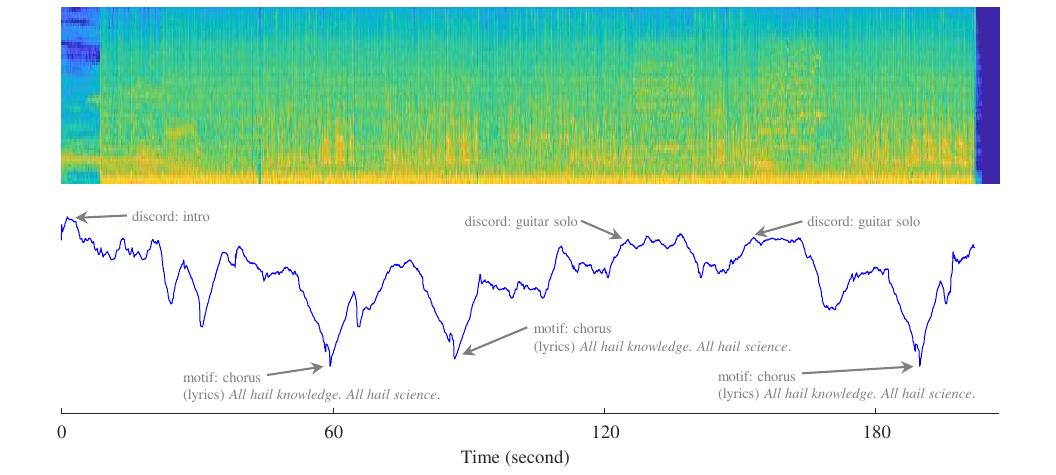}
\caption{
\emph{top}) The Mel-spectrogram of the song \emph{All Hail Science}~\cite{allegaeon2016all} by American death metal band, Allegaeon.
\emph{bottom}) The associated matrix profile can help users identify intro (discord), chorus (motif), and guitar solo (discord).
}
\label{ch1figallhail}
\end{figure}

It is also common that the user is interested in \emph{both} time series motifs and discords instead of \emph{only} time series motifs or discords.
Let us say a musician is analyzing the song structure of the song \emph{All Hail Science}~\cite{allegaeon2016all} by American death metal band, Allegaeon in the Mel-spectrogram space (Figure~\ref{ch1figallhail}.\emph{top}).
Because time series motif usually corresponds to the chorus, while time series discord usually corresponds to improvisation segments~\cite{silva2016ismir, silva2018fast}, the user may need to identify both time series motifs and discords from the music recording for such analysis.
Matrix profile (Figure~\ref{ch1figallhail}.\emph{bottom}) not only conveniently provides both types information \emph{exactly}, but also gives a visually intuitive way to display such information.
The musician can quickly identify the time series motifs by examining the subsequences with low matrix profile values and can identify time series discord by examining the subsequences with high matrix profile values.

This thesis contains an introductory summary for the matrix profile research. 
Our contribution includes:

\begin{itemize}
\item We introduce \emph{parameter-free}, \emph{exact}, and \emph{scalable} matrix profile algorithm for single and multidimensional time series.
\item We provide an \emph{efficient} and \emph{accurate} motif identification framework that is capable of discovering subdimensional motifs in multidimensional time series.
\item By adopting matrix profile, we demonstrate \emph{interpretable} models can be learned from time series with \emph{weak} labels.
\item We show that combining matrix profile with autoencoder yields more \emph{powerful} representations comparing to the standard autoencoder.
\end{itemize}

The rest of the thesis is organized as follows.
In Chapter~\ref{ch2}, we introduce the idea of matrix profile and algorithms computing it.
Chapter~\ref{ch3} shows the application of matrix profile in motif discovery (specifically for subdimensional motifs in multidimensional time series).
Then, in Chapter~\ref{ch4} and Chapter~\ref{ch5}, we demonstrate the usefulness of matrix profile in weakly labeled time series classification and representation learning.
Finally, we offer conclusions and directions for future work in Chapter~\ref{ch6}.

\chapter{Matrix Profile}
\label{ch2}
The basic problem statement for \emph{all-pairs-similarity-search} (also known as \emph{similarity join}) problem is this: \emph{Given a collection of data objects, retrieve the nearest neighbor for every object}. 
In the text domain, the dozens of algorithms which have been developed to solve the similarity join problem (and its variants) have been applied to an increasingly diverse set of tasks, such as community discovery, duplicate detection, collaborative filtering, clustering, and query refinement~\cite{agrawal1993efficient}.
However, while virtually all text processing algorithms have analogues in time series data mining~\cite{mueen2009exact}, there has been surprisingly little progress on Time Series subsequences All-Pairs-Similarity-Search (TSAPSS).

It is clear that a scalable TSAPSS algorithm would be a versatile building block for developing algorithms for many time series data mining tasks (e.g., motif discovery, shapelet discovery, semantic segmentation and clustering). 
As such, the lack of progress on TSAPSS stems not from a lack of interest, but from the daunting nature of the problem. 
Consider the following example that reflects the needs of an industrial collaborator: A boiler at a chemical refinery reports pressure once a minute. 
After a year, we have a time series of length 525,600. 
A plant manager may wish to do a similarity self-join on this data with week-long subsequences (10,080) to discover operating regimes (summer vs. winter or light-distillate vs. heavy-distillate etc.) 
The obvious nested loop algorithm requires 132,880,692,960 Euclidean distance computations. 
If we assume each one takes 0.0001 second, then the join will take 153.8 days. 
The core contribution of this study is to show that we can reduce this time to 1.2 hours, using an off-the-shelf desktop computer. 
Moreover, we show that this join can be computed and/or updated incrementally. 
Thus, we could \emph{maintain} this join essentially forever on a standard desktop, even if the data arrival frequency was much faster than once a minute.

In this chapter, we are introducing a novel idea called \emph{Matrix Profile}, which uses an ultra-fast similarity search algorithm under $z$-normalized Euclidean distance as a subroutine, exploiting the redundancies between overlapping subsequences to achieve its dramatic speedup and low space overhead. 

The matrix profile has the following advantages/features:

\begin{itemize}
\item \textbf{It is exact}: 
Our method provides no false positives or false dismissals. 
This is an important feature in many domains. 
For example, a recent paper has addressed the TSAPSS problem in the special case of earthquake telemetry~\cite{yoon2015earthquake}. 
The method \emph{does} achieve speedup over brute force, but allows false dismissals. 
A single high-quality seismometer can cost \$10,000 to \$20,000~\cite{alibaba}, and the installation of a seismological network can cost many millions. 
Given that cost and effort, users may not be willing to miss a single nugget of exploitable information, especially in a domain with implications for human life.
\item \textbf{It is simple and parameter-free}: 
In contrast, the more general metric space APSS algorithms typically require building and tuning spatial access methods and/or hash functions~\cite{luo2012efficient,ma2016parallel,yoon2015earthquake}.
\item \textbf{It is space efficient}: 
Our algorithm requires an inconsequential space overhead, just $O(n)$ with a small constant factor. 
In particular, we avoid the need to actually \emph{extract} the individual subsequences~\cite{luo2012efficient,ma2016parallel} something that would increase the space complexity by two or three orders of magnitude, and as such, force us to use a disk-based algorithm, further reducing time performance.
\item \textbf{It is an anytime algorithm}:
While our \emph{exact} algorithm is extremely scalable, for extremely large datasets we can compute the results in an anytime fashion~\cite{assent2012anyout,ueno2006anytime}, allowing ultra-fast \emph{approximate} solutions.
\item \textbf{It is incrementally maintainable}: 
Having computed the similarity join for a dataset, we can incrementally update it very efficiently. 
In many domains this means we can effectively maintain exact joins on streaming data forever. 
\item \textbf{It does not require the user to set a similarity/distance threshold}: 
Our method provides \emph{full} joins, eliminating the need to specify a similarity \emph{threshold}, which as we will show, is a near impossible task in this domain. 
\item \textbf{It can leverage hardware}: 
Our algorithm is embarrassingly parallelizable, both on multicore processors and in distributed systems.
\item \textbf{It has time complexity that is constant in subsequence length}:
This is a very unusual and desirable property; virtually all time series algorithms scale poorly as the subsequence length grows~\cite{ding2008querying,mueen2009exact}.
\item \textbf{It takes deterministic time}:
This is also unusual and desirable property for an algorithm in this domain. 
For example, even for a fixed time series length, and a fixed subsequence length, all other algorithms we are aware of can radically different times to finish on two (even \emph{slightly}) different datasets. 
In contrast, given only the length of the time series, we can predict precisely how long it will take our finish in advance.
\end{itemize}

Given all these features, our algorithm may have implications for many time series data mining tasks~\cite{chandola2009detecting, hao2012visual,mueen2009exact, yoon2015earthquake}.

In recent work, we have introduced several matrix profile algorithms and exemplify applications in various domains.
Matrix profile was first proposed in~\cite{yeh2016icdm} as a data structure which holds essential information (i.e., the similarity of subsequences) for time series.
The paper outlines a simple algorithm for computing the matrix profile using Mueen's Algorithm for Similarity Search (MASS)~\cite{masspage} and demonstrates the usefulness of matrix profile on various basic time series data mining tasks~\cite{yeh2016icdm}.
A more efficient algorithm for computing matrix profile was later introduced in~\cite{yeh2017dmkd, zhu2016icdm, zhu2018exploiting} in which the matrix profile is computed in parallel utilizing Graphics Processing Unit (GPU).
In~\cite{yeh2017icdm}, the author generalized the original matrix profile definition to explain the subspace similarity\footnote{Similarity obtained only using a subset of dimensions rather than all dimension like in~\cite{silva2016ismir,silva2018fast}.} for multidimensional time series.
The matrix profile has shown effective in many different applications of time series data mining, including music information retrieval~\cite{silva2018summarizing,silva2016ismir,silva2018fast}, weakly labeled classification in time series~\cite{yeh2017vldb}, time series data visualization~\cite{yeh2016icdm2}, semantic segmentation~\cite{gharghabi2017icdm, gharghabi2018dmkd}, time series chain discovery~\cite{zhu2017icdm}, augmented time series motif discovery~\cite{dau2017kdd}, and variable-length motif discovery~\cite{linardi2018matrix}.
We refer interested reader to the matrix profile project website~\cite{mppage} for the more up-to-date information.

The rest of the chapter is organized as follows.
Section~\ref{ch2secrelated} and Section~\ref{ch2secdef} review related work and introduce the necessary background materials and definitions. 
In Section~\ref{ch2secmpsts}, we outline our algorithm and its anytime and incremental variants for single dimensional time series.
Then, in Section~\ref{ch2secmpmts}, we extend the aforementioned single dimensional matrix profile algorithm to compute multidimensional matrix profile.
Finally, in Section~\ref{ch2secconclude} we offer conclusions and directions for future work.

\section{Background and Related Work}
\label{ch2secrelated}
The particular variant of \emph{similarity join} problem we wish to solve is: Given a collection of data objects, retrieve the nearest neighbor for \emph{every} object. 
We believe this is the most basic version of the problem, and any solution for this problem can be easily extended to other variants of similarity join problem.

Other common variants include retrieving the top-K nearest neighbors or the nearest neighbor for each object if that neighbor is within a user-supplied threshold, $\tau$. (Such variations are trivial generalizations of our proposed algorithm, so we omit them from further discussion). 
The latter variant results in a much easier problem, provided that the threshold is reasonably small. 
For example, Agrawal et al.~\cite{agrawal1993efficient}  notes that virtually all research efforts ``\emph{exploit a similarity threshold more aggressively in order to limit the set of candidate pairs that are considered.. [or] ...to reduce the amount of information indexed in the first place}".

\begin{figure}[htb]
\centering
\includegraphics[trim={9.5cm 7.5cm 9.5cm 7.5cm}, clip, width=1\textwidth,page=17]{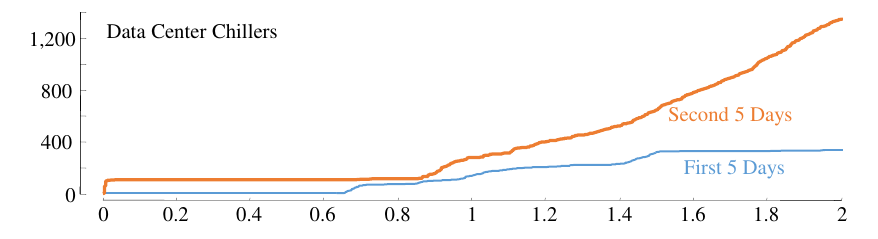}
\caption{
Output size vs. threshold for data center chillers~\cite{patnaik2009sustainable}. 
Values beyond 2.0 are truncated for clarity (see~\cite{yeh2017dmkd} for the full data). 
For a large range of thresholds (from 0 to 0.65) the difference in the selectivity is enormous, from {\color{orange}zero} to {\color{blue}one-hundred and eight}.
}
\label{ch2figrelated}
\end{figure}

This critical dependence on $\tau$ is a major issue for text joins, as it is known that ``\emph{join size can change dramatically depending on the input similarity threshold}"~\cite{lee2011similarity}. 
However, this issue is even more critical for time series for two reasons. 
First, unlike \emph{similarity} (which is bounded between zero and one), the Euclidean distance is effectively unbounded, and generally not intuitive. 
For example, if two heartbeats have a Euclidean distance of 17.1, are they similar? 
Even if you are a domain expert and know the sampling rate and the noise level of the data, this is not obvious. 
Second, a single threshold can produce radically different output sizes, even for datasets that are very similar to the human eye.  
Consider Figure~\ref{ch2figrelated} which shows the output size vs. threshold setting for the first and second halves of a ten-day period monitoring data center chillers~\cite{patnaik2009sustainable}. 
For the first five days a threshold of 0.6 would return zero items, but for the second five days the same setting would return 108 items.  
This shows the difficulty in selecting an appropriate threshold. 
Our solution is to have \emph{no} threshold and do a \emph{full} join. 
After the join is computed, the user may then use any ad-hoc filtering rule to give the result set she desires. 
For example, the \emph{top-fifty matches}, or \emph{all matches in the top-two-percent}. 
Moreover, as demonstrated by Yeh et al.~\cite{yeh2017dmkd}, people may be interested in the \emph{bottom-fifty matches}, or \emph{all matches in the bottom-two-percent}. 
To the best of our knowledge, no research effort in time series joins can support such primitives, as all techniques explicitly exploit pruning strategies based on ``nearness"~\cite{luo2012efficient, ma2016parallel}.

A handful of efforts have considered joins on time series, achieving speedup by (in addition to the use of MapReduce) converting the data to lower-dimensional representations such as PAA~\cite{luo2012efficient} or SAX~\cite{ma2016parallel} and exploiting lower bounds and/or Locality Sensitive Hashing (LSH) to prune some calculations. 
However, the methods are very complex, with many (10-plus) parameters to adjust.
As Luo et al.~\cite{luo2012efficient} acknowledge with admirable candor, ``\emph{Reasoning about the optimal settings is not trivial}". 
In contrast, our proposed algorithm has zero parameters to set.

A very recent research effort~\cite{yoon2015earthquake} has tackled the scalability issue by converting the real-valued time series into discrete ``fingerprints" before using a LSH approach, much like the text retrieval community~\cite{agrawal1993efficient}. 
They produced impressive speedup, but they also experienced false negatives. 
Moreover, the approach has several parameters that need to be set; for example, they set the threshold to a very precise 0.818. 
In passing, we note that one experiment they performed offers confirmation of the pessimistic ``153.8 days" example we gave in the introduction. 
A brute-force experiment they conducted with slightly longer time series but much shorter subsequences took 229 hours, suggesting a value of about 0.0002 seconds per comparison, just twice our estimate (see~\cite{yeh2017dmkd} for analysis). 
We will revisit this work in Section~\ref{ch2secanytime}.

As we shall show, our algorithm allows both \emph{anytime} and \emph{incremental} (i.e. streaming) versions. 
While a streaming join algorithm for \emph{text} was recently introduced~\cite{de2016streaming} we are not aware of any such algorithms for time series data or general metric spaces. 
More generally, there is a huge volume of literature on joins for text and DNA processing~\cite{agrawal1993efficient}. 
Such work is interesting, but of little direct utility given our constraints, data type and problem setting. 
We are working with real-valued data, not discrete data. 
We require full joins, not threshold joins, and we are unwilling to allow the possibility, no matter how rare, of false negatives.

\section{Definitions and Notation}
\label{ch2secdef}
We begin by defining the data type of interest, \emph{time series}:

\begin{definition}
\normalfont
A \emph{time series} $T\in \mathbb{R}^n$ is a sequence of real-valued numbers $t_i\in \mathbb{R}\colon T=[t_1,t_2,\cdots,t_n]$ where $n$ is the length of $T$.
\end{definition}

For motif discovery, we are not interested in the global properties of a time series, but in the local \emph{subsequences}:

\begin{definition}
\normalfont
A \emph{subsequence} $T_{i,m}\in \mathbb{R}^m$ of a $T$ is a continuous subset of the values from $T$ of length $m$ starting from position $i$. Formally, $T_{i,m} =[t_i,t_{i+1},\cdots,t_{i+m-1}]$.
\end{definition}

The particular local properties that we seek to capture are \emph{time series motifs}:

\newpage
\begin{definition}
\normalfont
A \emph{time series motif} is the most similar subsequence pair of a time series. Formally, $T_{a,m}$ and $T_{b,m}$is the motif pair iff $dist(T_{a,m},T_{b,m})\leq dist(T_{i,m},T_{j,m} )\forall i,j\in [1,2,\cdots,n-m+1]$, where $a\neq b$ and $i\neq j$, and dist is a function that computes the $z$-normalized Euclidean distance between the input subsequences.
\end{definition}

We store the distance between a subsequence of a time series with all the other subsequences from the same time series in an ordered array called \emph{distance profile}.

\begin{definition}
\label{ch2defdp}
\normalfont
A \emph{distance profile} $D\in \mathbb{R}^{n-m+1}$ of a time series $T$ and a subsequence $T_{i,m}$ is a vector that stores $dist(T_{i,m},T_{j,m})\forall j\in[1,2,\cdots,n-m+1]$.
\end{definition}

\begin{figure}[htb]
\centering
\includegraphics[trim={8.5cm 7.5cm 8.5cm 6.5cm}, clip, width=1\textwidth,page=1]{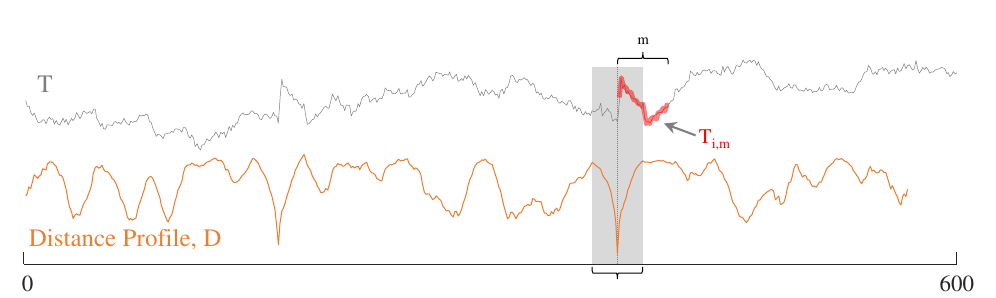}
\caption{
Distance profile of $T$ and $T_{i,m}$.
The lowest points on $D$ correspond to the querying subsequence ($T_{i,m}$).
}
\label{ch2figdp}
\end{figure}

The distance profile can be computed efficiently by using a convolution-based method such as Mueen's Algorithm for Similarity Search (MASS)~\cite{masspage}.
Figure~\ref{ch2figdp} shows the distance profile of $T$ and $T_{i,m}$. 

Note that by definition, the distance profile must be zero at the location of $T_{i,m}$, and close to zero just before and just after. 
Such matches are called \emph{trivial matches} in the literature~\cite{mueen2009exact}, and are avoided by ignoring an exclusion zone (shown as a gray region in Figure~\ref{ch2figdp}) of $m/2$ before and after the location of $T_{i,m}$.

The most efficient method of locating time series motifs exactly, is to compute the \emph{matrix profile}~\cite{yeh2016icdm}.

\begin{definition}
\label{ch2defmp}
\normalfont
A \emph{matrix profile} $P\in \mathbb{R}^{n-m+1}$ of a time series $T$ is a meta time series that stores the $z$-normalized Euclidean distance between each subsequence and its nearest neighbor, where $n$ is the length of $T$, and $m$ is the given subsequence length. 
The time series motif can be found by locating the two lowest values in $P$ (they will have tying values). 
Note that other definitions of motifs (range motifs, top-$K$ motifs etc.) can also be extracted trivially from the matrix profile~\cite{yeh2016icdm, yeh2017dmkd}.
\end{definition}

\begin{figure}[htb]
\centering
\includegraphics[trim={8.5cm 7.5cm 8.5cm 7.5cm}, clip, width=1\textwidth,page=2]{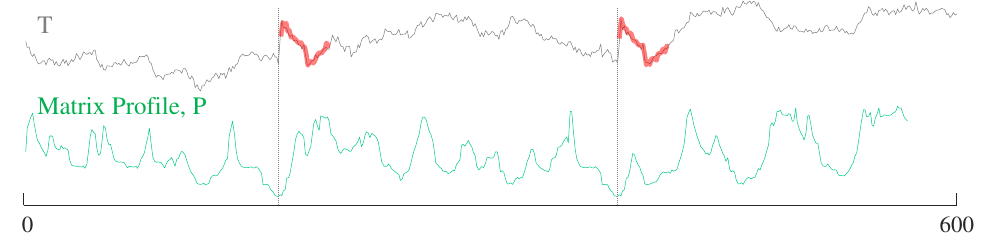}
\caption{
Matrix profile of $T$. 
The two lowest points on $P$ correspond to the locations of embedded motif pair ({\color{red}red}).
}
\label{ch2figmp}
\end{figure}

The time complexity to compute $P$ is $O(n^2)$~\cite{zhu2016icdm}. 
This may seem unscalable, but this is mitigated by the following facts. 
The time complexity does not depend on the length of the motifs\footnote{
The word ``dimensionality'' is overloaded for multidimensional time series. 
It is used both to refer to the number of time series and to the number of data points in a subsequence. 
For clarity, we only use it in the former sense.}. 
In contrast,~\cite{balasubramanian2016discovering, minnen2007detecting, tanaka2005discovery, vahdatpour2009toward} all scale poorly for longer motif lengths. 
Moreover, the matrix profile can be computed with a variety of algorithms/computational frameworks, including STAMP~\cite{yeh2016icdm, yeh2017dmkd}, STAMPI~\cite{yeh2016icdm, yeh2017dmkd}, STOMP~\cite{yeh2017dmkd, zhu2016icdm, zhu2018exploiting}, and GPU-STOMP~\cite{yeh2017dmkd, zhu2016icdm, zhu2018exploiting}, which can exploit both the available computational resources and domain constraints for optimal performance. 
Even without resorting to high-performance hardware, our algorithm is at least two orders of magnitude faster than~\cite{balasubramanian2016discovering, minnen2007detecting, tanaka2005discovery, vahdatpour2009toward}. 
Figure~\ref{ch2figmp} shows the matrix profile of $T$. 

\newpage
Although the motif pair ({\color{red}red}) is visually similar to the background random walk (black), the matrix profile still reveals the locations of the motif pair by strongly minimizing at the appropriate locations.
Since the matrix profile can be interpreted as a way to store the \emph{1-nearest-neighbor-graph} of subsequences (each node is a subsequence and each edge is the nearest neighbor relationship), Definition~\ref{ch2defmp} can be easily generalized to \emph{k-nearest-neighbor-graph}, if such information is required by the application.

The $i$th element in the matrix profile tells us the $z$-normalized Euclidean distance to the nearest neighbor of the subsequence of $T$, starting at $i$.  
However, it does not reveal \emph{where} that neighbor is located. 
This information is recorded in the \emph{matrix profile index}.

\begin{definition}
\label{ch2defmpi}
\normalfont
A \emph{matrix profile Index} $I\in \mathbb{R}^{n-m+1}$ of a time series $T$ is a meta time series that stores the identity (in terms of index) of each subsequence's nearest neighbor, where $n$ is the length of $T$, and $m$ is the given subsequence length.
\end{definition}

By storing the neighboring information this way, we can efficiently retrieve the nearest neighbor of $T_{i,m}$ by accessing the $i$th element in the matrix profile index.
In addition to the special case of a single dimensional time series, we generalize and extend the matrix profile (Definition~\ref{ch2defmp}) to find motifs in \emph{multidimensional time series}.

\begin{definition}
\normalfont
A \emph{multidimensional time series} $\mathbf{T}\in \mathbb{R}^{d\times n}$ is a set of co-evolving time series $T^{(i)}\in \mathbb{R}^n\colon \mathbf{T}=[T^{(1)},T^{(2)},\cdots,T^{(d)} ]^T$ where $d$ is the dimensionality of $\mathbf{T}$ and $n$ is the length of $\mathbf{T}$.
\end{definition}

Similarly, the definition of a subsequence in multidimensional setting becomes the following:

\begin{definition}
\normalfont
A \emph{multidimensional subsequence} $\mathbf{T}_{i,m}\in \mathbb{R}^{d\times m}$ of a multidimensional time series $\mathbf{T}$ is a set of univariant subsequences from $\mathbf{T}$ of length $m$ starting from position $i$. Formally, $\mathbf{T}_{i,m}=[T_{i,m}^{(1)},T_{i,m}^{(2)},\cdots,T_{i,m}^{(d)}]^T$.
\end{definition}

Using all dimensions for motif discovery is generally guaranteed to fail (A similar observation, but for time series \emph{classification}, is forcefully made in~\cite{hu2013classification}). 
In general, only a subset of all dimensions should be used for multidimensional motif discovery.

We refer to such subsets of subsequences \emph{subdimensional subsequences}.

\begin{definition}
\normalfont
A \emph{subdimensional subsequence} $\mathbf{T}_{i,m} (X)\in \mathbb{R}^{k\times m}$ is a multidimensional subsequence for which only a subset of dimension is selected, where $X$ is an indicator vector that shows which dimension is included, and $k$ is the number of dimension included (i.e., $\|X\|_0=k$).
\end{definition}

We want to compute the distance between two multidimensional subsequences by using only their corresponding subdimensional subsequences. 
The distance function that measures this relation is called \emph{k-dimensional distance}.

\begin{definition}
\label{ch2defkdp}
\normalfont
A \emph{k-dimensional distance profile} $D\in \mathbb{R}^{n-m+1}$ of a time series $\mathbf{T}$ and a subsequence $\mathbf{T}_{i,m}$ is a vector that stores $dist^{(k)} (\mathbf{T}_{i,m},\mathbf{T}_{j,m})\forall j\in [1,2,\cdots,n-m+1]$.
\end{definition}

Multidimensional motifs must also be redefined slightly to allow for representing within subdimensional setting.

\begin{definition}
\label{ch2defkmotif}
\normalfont
A \emph{k-dimensional motif} is the most similar subdimensional subsequence pair of a multidimensional time series when the distance is computed by using the $k$-dimensional distance function. 
Formally, $\mathbf{T}_{a,m}$ and $\mathbf{T}_{b,m}$ is the $k$-dimensional motif pair iff $dist^{(k)}(\mathbf{T}_{a,m},\mathbf{T}_{b,m})\leq dist^{(k)}(\mathbf{T}_{i,m},\mathbf{T}_{j,m})\forall i,j\in[1,2,\cdots,n-m+1]$, where $a\neq b$ and $i\neq j$.
\end{definition}

To find the $k$-dimensional motif, we modify the matrix profile for the $k$-dimensional motif problem.

\begin{definition}
\label{ch2defmmp}
\normalfont
A \emph{k-dimensional matrix profile} $P\in \mathbb{R}^{n-m+1}$ of a multidimensional time series $\mathbf{T}$ is a meta time series that stores the $z$-normalized Euclidean distance between each subsequence and its nearest neighbor (the distance is computed using $k$-dimensional distance function), where $n$ is the length of $\mathbf{T}$, $d$ is the dimensionality of $\mathbf{T}$, $k$ is the given number of dimension, and m is the given subsequence length. 
Formally, the ith position in $P$ stores $dist^{(k)}(\mathbf{T}_{i,m},\mathbf{T}_{j,m})\forall j\in[1,2,\cdots,n-m+1]$, where $i\neq j$.
The $k$-dimensional motif can be found by locating the two lowest values in $P$ (these two lowest values must be a tie~\cite{yeh2016icdm, yeh2017dmkd}).
\end{definition}

\begin{figure}[htb]
\centering
\includegraphics[trim={8.5cm 5cm 8.5cm 5cm}, clip, width=1\textwidth,page=3]{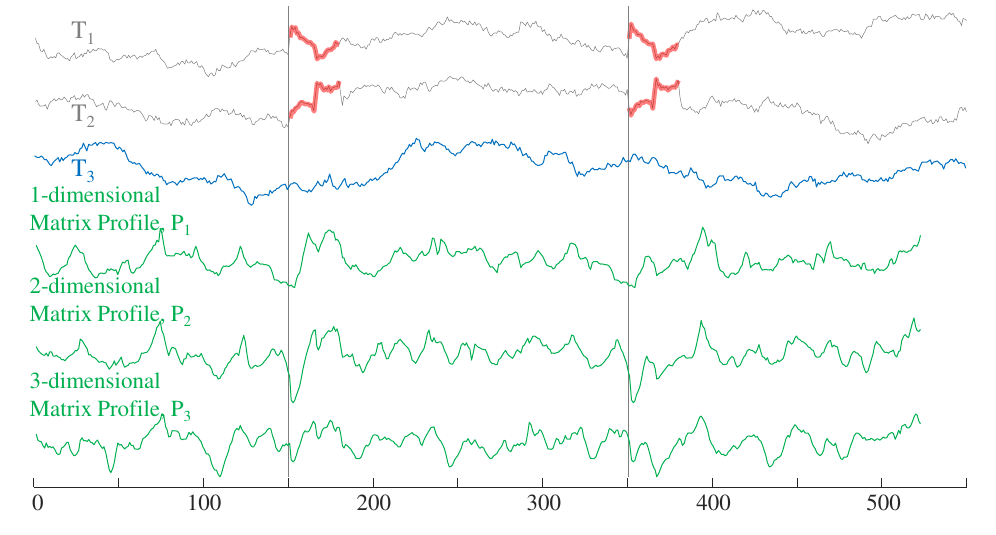}
\caption{
\emph{top)} The multidimensional time series $\mathbf{T}=[T_1, T_2, T_3]^T$.
\emph{Bottom)} The multidimensional matrix profiles of various subsets of the data. 
Note that the (implanted) semantically meaningful motif can be spotted visually by inspecting the lowest points of the 1-dimensional or 2-dimensional matrix profile, but the 3-dimensional case has the motifs in an effectively \emph{random} location.
}
\label{ch2figmmp}
\end{figure}

Figure~\ref{ch2figmmp} shows the $k$-dimensional matrix profile of the running example for all possible settings of $k$.

Note, the correct motif pair only appears in $P_1$ and $P_2$  (as the lowest point in the curve), since the inserted motif is 1-dimensional and 2-dimensional motif by definition.
The identity of each subsequence's nearest neighbor is stored in the \emph{k-dimensional matrix profile index} similar to Definition~\ref{ch2defmpi}.

A $k$-dimensional matrix profile only reveals the location of motifs in time, but it fails to reveal which $k$ out of the $d$ dimension contains the motif pair. 
To store this information, we define another meta time series called the \emph{k-dimensional matrix profile subspace}.

\begin{definition}
\normalfont
A \emph{k-dimensional matrix profile subspace} $\mathbf{S}\in\mathbb{R}^{k\times n-m+1}$ is a multidimensional meta time series that stores the selected $k$ dimension for each subsequence when computing the distance with others.
\end{definition}

\newpage
With these definitions formalized, we are ready to introduce our algorithms. 
Before continuing, we wish to clarify our claimed contributions. 
Our algorithm \emph{is} orders of magnitude faster than existing works~\cite{balasubramanian2016discovering, minnen2007detecting, tanaka2005discovery, vahdatpour2009toward}; however, this is simply a property we inherit from the use of the matrix profile~\cite{yeh2016icdm, yeh2017dmkd}, which is \emph{not} a claimed original contribution. 
Our contribution is in producing semantically meaningfully multidimensional motifs on a subset of a large MTS, which may comprise \emph{mostly} of irrelevant and spurious data.

\section{Matrix Profile for Single dimensional Time Series}
\label{ch2secmpsts}
We are finally in a position to explain our algorithms. 
We begin by stating the fundamental intuition, which stems from the relationship between distance profiles and the matrix profile. 
As Figure~\ref{ch2figdp} and Figure~\ref{ch2figmp} visually suggest, all distance profiles (excluding the trivial match region) are upper bound approximations to the matrix profile. 
More critically, if we compute \emph{all} the distance profiles, and take the minimum value at each location, the result is the matrix profile.

This tells us that if we have a fast way to compute the distance profiles, then we also have a fast way to compute the matrix profile. 
As we shall show in the next section, we \emph{do} have such an ultra-fast way to compute the distance profiles.

\subsection{The MASS algorithm}
\label{ch2secmass}
We begin by introducing a novel ultra-fast Euclidean distance similarity search algorithm called \emph{MASS} (Mueen's Algorithm for Similarity Search)~\cite{masspage} for time series data. 
The algorithm does not just find the nearest neighbor to a query and return its distance; it returns the distance to every subsequence.
In particular, it computes the \emph{distance profile}, as shown in Figure~\ref{ch2figdp}. The algorithm requires just $O(n\log n)$ time by exploiting the fast Fourier transform (FFT) to calculate the dot products between the query and all subsequences of the time series. 

We need to carefully qualify the claim of ``ultra-fast". There are dozens of algorithms for time series similarity search that utilize index structures to efficiently locate neighbors~\cite{ding2008querying}.
While such algorithms can be faster in the \emph{best case}, all of these algorithms degenerate to brute force search in the worst case\footnote{
There are \emph{many} such worse case scenarios, including high levels of noise blurring the distinction between closest and furthest neighbors and thus rendering triangular-inequality pruning and early abandoning worthless.} (actually, much worse than brute force search due to the overhead of the index). 
Likewise, there are index-free methods that achieve speed-up using various early abandoning tricks~\cite{rakthanmanon2012searching} but they too degrade to brute force search in the worst case. 
In contrast, the performance of the algorithms outlined in Algorithm~\ref{ch2algsliding} and Algorithm~\ref{ch2algmass} is completely \emph{independent} of the data.

\begin{algorithm}[htb]
\caption{Calculation of sliding dot products.}
\label{ch2algsliding}
\begin{algorithmic}[1]
\Statex \textbf{Procedure} $SlidingDotProduct(Q,T)$ \vspace{-1em} 
\Statex \textbf{Input:} A query $Q$, and a user provided time series $T$ \vspace{-1em} 
\Statex \textbf{Output:} The dot product between $Q$ and all subsequences in $T$ \vspace{-1em} \
\State $n \gets Length(T)$, $m \gets Length(Q)$ \vspace{-1em} 
\State $T_a \gets$ Append $T$ with $n$ zeros \vspace{-1em} 
\State $Q_r \gets Reverse(Q)$ \vspace{-1em} 
\State $Q_{ra} \gets$ Append $Q_r$ with $2n-m$ zeros \vspace{-1em} 
\State $Q_{raf} \gets FFT(Q_{ra})$, $T_{af} ← FFT(T_a)$ \vspace{-1em} 
\State $QT \gets InverseFFT(ElementwiseMultiplication(Qraf, Taf))$ \vspace{-1em}
\State \textbf{return} $QT$
\end{algorithmic}
\end{algorithm}

\begin{figure}[htb]
\centering
\includegraphics[trim={10cm 7cm 10cm 7cm}, clip, width=1\textwidth,page=4]{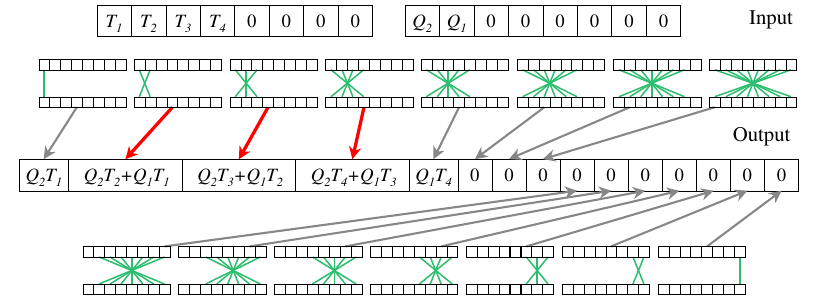}
\caption{
A toy example of convolution operation being used to calculate sliding dot products for time series data. 
Note the \emph{reverse} and \emph{append} operation on $T$ and $q$ in the input. 
Fifteen dot products are calculated for every slide. 
The cells $m = 2$ to $n = 4$ from left ({\color{red}red}/bold arrows) contain valid products. 
Algorithm~\ref{ch2algmass} takes this subroutine and uses it to create a distance profile (see Definition~\ref{ch2defdp}).
}
\label{ch2figqt}
\end{figure}

Line 1 determines the length of both the time series $T$ and the query $Q$. 
In line 2, we use that information to append $T$ with an equal number of zeros.
In line 3, we obtain the mirror image (i.e. Reverse) of the original query. 
Reverse of a sequence $[x_1, x_2, x_3, \cdots, x_n]$ is $[x_n , x_{n-1}, x_{n-2},\cdots, x_1]$. 
Reversing a sequence takes only linear time.
Typically, the query time series is small, and the cost of reversing is so small it is difficult to reliably measure.
This reversing ensures that a convolution (i.e. ``crisscrossed" multiplication) essentially produces in-order alignment. 
Because we require both vectors to be the same length, in line 4 we append enough zeros to the (now reversed) query so that, like $T_a$, it is also of length $2n$.
In line 5, the algorithm calculates Fourier transforms of the appended-reversed query ($Q_{ra}$) and the appended time series $T_a$. 
Note that we use FFT algorithm which is an $O(n\log n)$ algorithm.
The $Q_{raf}$ and the $T_{af}$ produced in line 5 are vectors of complex numbers representing frequency components of the two time series.
The algorithm calculates the element-wise multiplication of the two complex vectors and performs inverse FFT on the product.
Lines 5-6 are the classic convolution operation on two vectors~\cite{wikiconvolution}. 
Figure~\ref{ch2figqt} shows a toy example of the sliding dot product function in work. 
Note that the algorithm time complexity does not depend on the length of the query ($m$).

In line 1 of Algorithm~\ref{ch2algmass}, we invoke the dot products code outlined in Algorithm~\ref{ch2algsliding}. 
The formula to calculate the $z$-normalized Euclidean distance $D[i]$ between two time series subsequence $Q$ and $T({i,m}$ using their dot product, $QT[i]$ is~\cite{yeh2016icdm, yeh2017dmkd}:
\begin{equation}
\label{ch2eqqt}
D[i] = \sqrt{2m\left(1-\frac{QT[i]-m\mu_QM_T[i]}{m\sigma_Q\Sigma_T[i]}\right)}
\end{equation}
where $m$ is the subsequence length, $\mu_Q$ is the mean of $Q$, $M_T[i]$ is the mean of $T_{i,m}$, $\sigma_Q$ is the standard deviation of $Q$, and  $\Sigma_T[i]$ is the standard deviation of $T_{i,m}$.
Normally, it takes $O(m)$ time to calculate the mean and standard deviation for every subsequence of a long time series. 
However, here we exploit a technique noted in~\cite{rakthanmanon2012searching} in a different context. 
We cache cumulative sums of the values and square of the values in the time series. 
At any stage the two cumulative sum vectors are sufficient to calculate the mean and the standard deviation of any subsequence of any length.

\begin{algorithm}[htb]
\caption{The MASS algorithm.}
\label{ch2algmass}
\begin{algorithmic}[1]
\Statex \textbf{Procedure} $MASS(Q,T)$ \vspace{-1em} 
\Statex \textbf{Input:} A query $Q$, and a user provided time series $T$ \vspace{-1em} 
\Statex \textbf{Output:} A distance profile $D$ of the query $Q$ \vspace{-1em} 
\State $QT\gets SlidingDotProducts(Q, T)$\Comment{Algorithm~\ref{ch2algsliding}} \vspace{-1em} 
\State $\mu Q, \sigma Q, Μ_T, \Sigma_T\gets ComputeMeanStd(Q, T)$\Comment{\cite{rakthanmanon2012searching}} \vspace{-1em} 
\State $D\gets CalculateDistanceProfile(QT, \mu Q, \sigma Q, Μ_T, \Sigma_T)$\Comment{Equation~\ref{ch2eqqt}} \vspace{-1em} 
\State \textbf{return} $D$
\end{algorithmic}
\end{algorithm}

Unlike the dozens of time series KNN search algorithms in the literature~\cite{ding2008querying}, this algorithm calculates the distance to \emph{every} subsequence, i.e. the \emph{distance profile} of time series $T$. 
Alternatively, in join nomenclature, the algorithm produces one full row of the all-pair similarity matrix. 
Thus, as we show in the next section, our join algorithm is little more than a loop that computes each full row of the all-pair similarity matrix and updates the current ``best-so-far" matrix profile when warranted.

\subsection{The STAMP Algorithm}
We call our join algorithm STAMP, Scalable Time series Anytime Matrix Profile. 
The algorithm is outlined in Algorithm~\ref{ch2algstamp}. 
In line 1, we extract the length of $T$. 
In line 2, we allocate memory and initial matrix profile $P$ and matrix profile index $I$. 
From lines 3 to line 6, we calculate the distance profiles $D$ using each subsequence $T[idx]$ in the time series $T$ and the time series $T$. 
Then, we perform pairwise minimum for each element in $D$ with the paired element in $P$ (i.e., $min(D[i], P[i])$ for $i = 0$ to $length(D) - 1$).
We also update $I[i]$ with $idx$ when $D[i] \leq P[i]$ as we perform the pairwise minimum operation.
Trivial matches is  ignored in $D$ when performing $ElementWiseMin$ in line 5.
Finally, we return the result $P$ and $I$ in line 7.

Note that Algorithm~\ref{ch2algstamp} computes the matrix profile for the self-similarity join.
Please refer to~\cite{yeh2017dmkd} for the STAMP that computes the general similarity join matrix profile.

\begin{algorithm}[htb]
\caption{The STAMP algorithm.}
\label{ch2algstamp}
\begin{algorithmic}[1]
\Statex \textbf{Procedure} $STAMP(T,m)$ \vspace{-1em} 
\Statex \textbf{Input:} A user provided time series $T$ and interested subsequence length $m$ \vspace{-1em} 
\Statex \textbf{Output:} A matrix profile $P$ and matrix profile index $I$ \vspace{-1em} 
\State $n\gets Length(T)$\vspace{-1em} 
\State $P\gets infs$, $I\gets zeros$, $idxes\gets 0:n-m$\vspace{-1em}
\For{\textbf{each} $idx$ \textbf{in} $idxes$} \Comment{in any order, but random for anytime algorithm}\vspace{-1em} 
\State $D\gets MASS(T[idx], T)$ \Comment{Algorithm~\ref{ch2algmass}} \vspace{-1em} 
\State $P, I\gets ElementWiseMin(P, I, D, idx)$ \vspace{-1em} 
\EndFor \vspace{-1em} 
\State \textbf{return} $P, I$
\end{algorithmic}
\end{algorithm}

To parallelize the STAMP algorithm for multicore machines, we simply distribute the indexes to secondary process run in each core, and the secondary processes use the indexes they received to update their own $P$ and $I$. 
Once the main process returns from all secondary processes, we use a function similar to $ElementWiseMin$ to merge the received $P$ and $I$.

\subsection{On the Anytime Property of STAMP}
\label{ch2secanytime}
While the exact algorithm introduced in the previous section is extremely scalable, there will always be datasets for which time needed for an \emph{exact} solution is untenable. 
We can mitigate this by computing the results in an \emph{anytime} fashion, allowing fast \emph{approximate} solutions~\cite{zilberstein1995approximate}. 
To add the anytime nature to the STAMP algorithm, all we need to do are to ensure a randomized order when we select subsequences from $T$ in line 2 of Algorithm~\ref{ch2algstamp}.

\begin{figure}[htb]
\centering
\includegraphics[trim={10cm 8cm 10cm 7.5cm}, clip, width=1\textwidth,page=5]{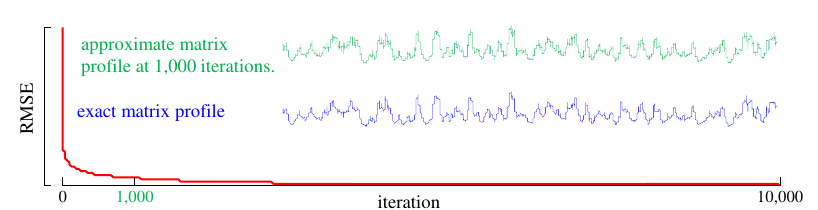}
\caption{
\emph{main}) The decrease in RMSE as the STAMP algorithm updates matrix profile with the distance profile calculated at each iteration.
\emph{inset}) The approximate matrix profile at the 10\% mark is visually indistinguishable from the final matrix profile.
}
\label{ch2figanytime}
\end{figure}

We can compute a (post-hoc) measurement of the quality of an anytime solution by measuring the Root-Mean-Squared-Error (RMSE) between the true matrix profile and the current best-so-far matrix profile. 
As Figure~\ref{ch2figanytime} suggests, with an experiment on random walk data, the algorithm converges very quickly.

Zilberstein and Russell~\cite{zilberstein1995approximate} give a number of desirable properties of anytime algorithms, including \emph{Low Overhead}, \emph{Interruptibility}, \emph{Monotonicity}, \emph{Recognizable Quality}, \emph{Diminishing Returns}, and \emph{Preemptability} (the meanings of these properties are mostly obvious from their names, but full definitions are at~\cite{zilberstein1995approximate}).

Because each subsequence's distance profile is bounded below by the exact matrix profile, updating an approximate matrix profile with a distance profile with pairwise minimum operation either drives the approximate solution closer the exact solution or retains the current approximate solution. 
Thus, we have guaranteed \emph{Monotonicity}. 
From Figure~\ref{ch2figanytime}, the approximate matrix profile converges to the exact matrix profile superlinearly; therefore, we have strong \emph{Diminishing Returns}. 
We can easily achieve \emph{Interruptibility} and \emph{Preemptability} by simply inserting a few lines of code between lines 5 and 6 of Algorithm~\ref{ch2algstamp} that read:

\begin{algorithm}[htb]
\begin{algorithmic}[1]
\If{$CheckForUserInterrupt$ \textbf{is} $True$}\vspace{-1em} 
\State \textbf{return} $P, I$ \Comment{return an approximate answer} \vspace{-1em} 
\EndIf \vspace{-1em}
\If{$isFurtherRefine$ \textbf{is not} $True$}\vspace{-1em}
\State \textbf{break} \vspace{-1em}
\EndIf
\end{algorithmic}
\end{algorithm}

The space and time overhead for the anytime property is effectively zero; thus, we have \emph{Low Overhead}. 
This leaves only the property of \emph{Recognizable Quality}. 
Here we must resort to a probabilistic argument.  
The convergence curve shown in Figure~\ref{ch2figanytime} is very typical, so we could use past convergence curves to predict the quality of solution when interrupted on similar data.

\subsection{The Utility of Anytime Matrix Profile}
\begin{figure}[htb]
\centering
\includegraphics[trim={10cm 7cm 10cm 7cm}, clip, width=1\textwidth,page=6]{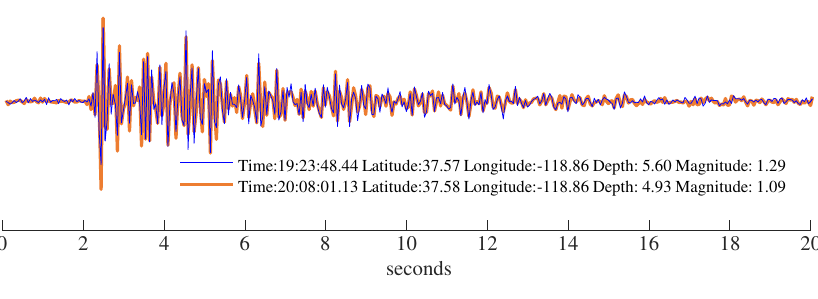}
\caption{
A set of doublets extracts from the seismic data recorded at a station near Mammoth Lakes on February 17th, 2016. 
One occurrence (fine/{\color{blue}blue}) is overlaid on top of another occurrence (bold/{\color{orange}orange}) that happened about 45 minutes later. 
}
\label{ch2figdoublet}
\end{figure}

In the early 1980's it was discovered that in telemetry of seismic data recorded by the same instrument from sources in given region there will be many similar seismograms~\cite{geller1980four}.
Geller and Mueller~\cite{geller1980four} suggested that ``\emph{The physical basis of this clustering is that the earthquakes represent repeated stress release at the same asperity, or stress concentration, along the fault surface}". 
These repeated patterns are call “doublets” in seismology, and exactly correspond to the more general term ``time series motifs". 
Figure~\ref{ch2figdoublet} shows an example of doublets from seismic data. 
A more recent paper notes that many fundamental problems in seismology can be solved by joining seismometer telemetry in search of these doublets~\cite{yoon2015earthquake}, including the discovery of foreshocks, aftershocks, triggered earthquakes, swarms, volcanic activity and induced seismicity (we refer the interested reader to the original paper for details). 
However, the paper notes a join with a query length of 200 on a data stream of length 604,781 requires 9.5 days. 
Their solution, a clever transformation of the data to allow Locality-Sensitive Hashing (LSH) based techniques, does achieve significant speedup, but at the cost of false negatives and the need for significant parameter tuning.

\begin{figure}[htb]
\centering
\includegraphics[trim={10cm 7.5cm 10cm 7.5cm}, clip, width=1\textwidth,page=7]{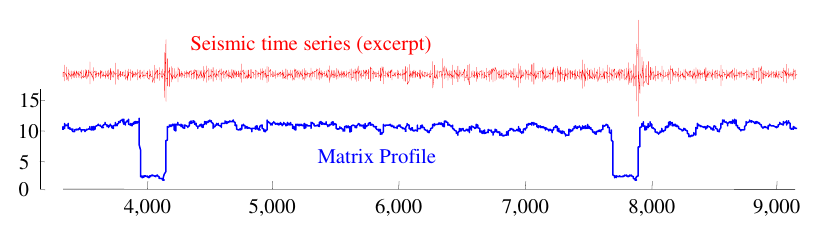}
\caption{
\emph{top}) An excerpt of a seismic time series aligned with its matrix profile (\emph{bottom}).
The ground truth provided by Yoon et al.~\cite{yoon2015earthquake} requires that the events occurring at time 4,050 and 7,800 match. 
}
\label{ch2figanyseismic}
\end{figure}

The authors kindly shared their data and, as we hint at in Figure~\ref{ch2figanyseismic}, confirmed that our STOMP approach does not have false negatives.

\begin{figure}[htb]
\centering
\includegraphics[trim={10cm 7.5cm 10cm 7.5cm}, clip, width=1\textwidth,page=8]{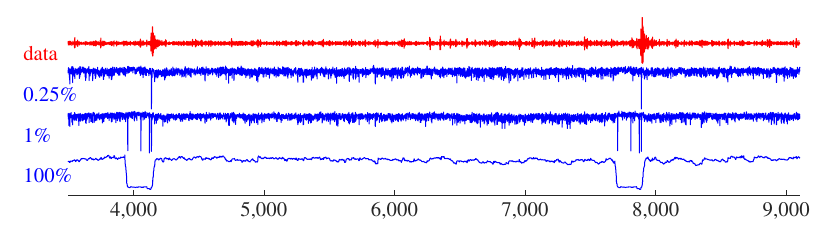}
\caption{
\emph{top}) An excerpt of the seismic data that is also shown in Figure~\ref{ch2figanyseismic}. 
\emph{top-to-bottom}) The approximations of the matrix profile for increasing interrupt times. 
By the time we have computed just 0.25\% of the calculations required for the full algorithm, the minimum of the matrix profile points to the ground truth.
}
\label{ch2figanyseismic2}
\end{figure}

We repeated the $n = 604,781, m = 200$ experiment and found it took just 1.7 hours to finish. 
As impressive as this is, we would like to claim that we can do even better.  
The seismology dataset offers an excellent opportunity to demonstrate the utility of the anytime version of our algorithm. 
The authors of~\cite{yoon2015earthquake} revealed their long-term ambition of mining even larger datasets~\cite{beroza2016personal}. 
In Figure~\ref{ch2figanyseismic2} we repeated the experiment with the snippet shown in Figure~\ref{ch2figanyseismic}, this time reporting the \emph{best-so-far} matrix profile reported by the STAMP algorithm at various milestones. 
Even with just 0.25\% of the distances computed (that is to say, 400 times faster), the correct answer has emerged. 

Thus, we can provide the correct answers to the seismologists in just minutes, rather than the 9.5 days originally reported.

To show the generality of this anytime feature of STAMP, we consider a very different dataset. 
As shown in Algorithm~\ref{ch2algdna}, it is possible to convert DNA to a time series~\cite{rakthanmanon2012searching}. 
We converted the Y-chromosome of the Chimpanzee (\emph{Pan troglodytes}) this way.

\begin{algorithm}[htb]
\caption{An algorithm for converting DNA string sequence to DNA time series.}
\label{ch2algdna}
\begin{algorithmic}[1]
\Statex \textbf{Procedure} $ConvertDNAStringToTimeSeries(chromosome)$ \vspace{-1em} 
\Statex \textbf{Input:} DNA string sequence $chromosome$ \vspace{-1em} 
\Statex \textbf{Output:} DNA sequence in form of time series $T$ \vspace{-1em} 
\State $T[0]\gets 0$\vspace{-1em} 
\For{$i\gets 0$ \textbf{to} $Length(chromosome)$}\vspace{-1em}
\State \textbf{if} $chromosome[i]$ \emph{is} {\color{green}$A$}, \textbf{then} $T[i + 1]\gets T[i] + 2$\vspace{-1em} 
\State \textbf{if} $chromosome[i]$ \emph{is} {\color{yellow}$G$}, \textbf{then} $T[i + 1]\gets T[i] + 1$\vspace{-1em} 
\State \textbf{if} $chromosome[i]$ \emph{is} {\color{blue}$C$}, \textbf{then} $T[i + 1]\gets T[i] - 1$\vspace{-1em} 
\State \textbf{if} $chromosome[i]$ \emph{is} {\color{red}$T$}, \textbf{then} $T[i + 1]\gets T[i] - 2$\vspace{-1em} 
\EndFor \vspace{-1em} 
\State \textbf{return} $T$
\end{algorithmic}
\end{algorithm}

\begin{figure}[htb]
\centering
\includegraphics[trim={10cm 6.5cm 10cm 6.5cm}, clip, width=1\textwidth,page=9]{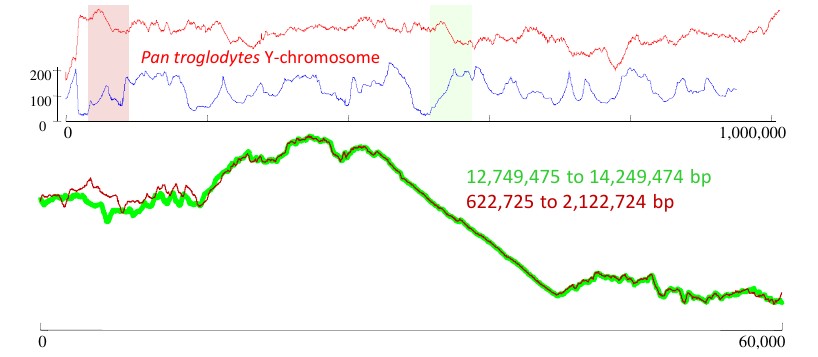}
\caption{
\emph{top}) The Y-chromosome of the Chimp in time series space with its matrix profile. 
\emph{bottom}) A zoom-in of the top motif discovered using anytime STAMP, we believe it to be an \emph{amplicon}~\cite{hughes2010chimpanzee}.
}
\label{ch2figanydna}
\end{figure}

While the original string is of length 25,994,497, we downsampled by a factor of twenty-five to produce a time series that is little over one-million in length. 
We performed a self-join with $m = 60,000$.  
Figure~\ref{ch2figanydna}.\emph{bottom} shows the best motif is so well conserved (ignoring the first 20\%), that it must correspond to a recent (in evolutionary time) gene duplication event.
In fact, in a subsequent analysis we discovered that ``\emph{much of the Y (Chimp chromosome) consists of lengthy, highly similar repeat units, or `amplicons'}"~\cite{hughes2010chimpanzee}.

This demanding join would take just over a day of CPU time (see Figure~\ref{ch2figspeed1}). 
However, using anytime STAMP we have the result shown above after doing just 0.021\% of the computations, in about 18 seconds. 
We have videos~\cite{anytimeStamp} that intuitively show the rapid convergence of the anytime variant of STAMP.

\subsection{The STOMP Algorithm}
\label{ch2secstomp}
As impressive as STAMP's time efficiency is, we can create an \emph{even faster} algorithm if we are willing to sacrifice one of STAMP's features: its anytime nature. 
This is a compromise that many users may be willing to make.
Because this variant of STAMP performs an \emph{ordered} (not \emph{random}) search, we call it STOMP, Scalable Time series Ordered-search Matrix Profile.

As we will see, the STOMP algorithm is very similar to STAMP, and at least in principle it is \emph{still} an anytime algorithm. 
However, because STOMP must compute the distance curves in a left-to-right order, it is vulnerable to an ``adversarial'' time series dataset which has motifs only towards the right side, and random data on the left side. 
For such a dataset, the convergence curve for STOMP will similar to Figure~\ref{ch2figanytime}, but the best motifs will not be discovered until the final iterations of the algorithm. 
This is important because we expect the most common use of the matrix profile will be in supporting \emph{motif discovery}, given that motif discovery has emerged as one of the most commonly used time series analysis tools in recent years~\cite{brown2013dictionary, mueen2009exact, shao2012temporal, yoon2015earthquake}.  
In contrast, STAMP is not vulnerable to such an ``adversarial arrangement of motifs'' argument as it computes the distance profiles in random order (Algorithm~\ref{ch2algstamp}, line 3). 
With this background stated, we are now in a position to explain how STOMP works.

In Section~\ref{ch2secmass} we introduced a formula to calculate the $z$-normalized Euclidean distance of two time series subsequences $Q$ and $T_{i,m}$ using their dot product. 
Note that the query $Q$ is also a subsequence of a time series; let us call this time series the Query Time Series, and denote it $T^Q$ ($T^Q = T$ if we are calculating self-join). 
To better explain the STOMP algorithm, here we denote query $Q$ as $T^Q_{j,m}$, where $j$ is the starting position of $Q$ in $T^Q$. 
We denote the $z$-normalized Euclidean distance between $T^Q_{j,m}$ and $T_{i,m}$ as $D_{j,i}$, and their dot product as $QT_{j,i}$. 
Equation~\ref{ch2eqqt} in Section~\ref{ch2secmass} can then be rewritten as:
\begin{equation}
\label{ch2eqqt2}
D_{j,i} = \sqrt{2m\left(1-\frac{QT_{j,i}-m\mu_j\mu_i}{m\sigma_j\sigma_i}\right)}
\end{equation}
where $m$ is the subsequence length, $\mu_j$ is the mean of $T^Q_{j,m}$, $\mu_i$ is the mean of $T_{i,m}$, $\sigma_j$ is the standard deviation of $T^Q_{j,m}$, and  $\sigma_i$ is the standard deviation of $T_{i,m}$.

The technique introduced in~\cite{rakthanmanon2012searching} allows us to obtain the means and standard deviations with $O(1)$ time complexity; thus, the time required to compute $D_{j,i}$ depends mainly on the time required to compute $QT_{j,i}$. 
Here we claim that $QT_{j,i}$ can also be computed in $O(1)$ time, once $QT_{j-1,i-1}$ is known.

Note that $QT_{j-1,i-1}$ can be decomposed as:
\begin{equation}
\label{ch2eqqtdecomp}
QT_{j-1,i-1}=\sum\limits_{k=0}^{m-1} T^Q_{j-1+k}T_{i-1+k}
\end{equation}
and $QT_{j,i}$ can be decomposed as:
\begin{equation}
\label{ch2eqqtdecomp2}
QT_{j,i}=\sum\limits_{k=0}^{m-1} T^Q_{j+k}T_{i+k}
\end{equation}
Thus we have
\begin{equation}
\label{ch2eqqtupdate}
QT_{j,i}=QT_{j-1,i-1}-T^Q_{j-1}T_{i-1}+T^Q_{j+m-1}T_{i+m-1}
\end{equation}
Our claim is thereby proved.

The relationship between $QT_{j,i}$ and $QT_{j-1,i-1}$ indicates that once we have the distance profile of time series $T$ with regard to $T^Q_{j-1,m}$, we can obtain the distance profile with regard to $T^Q_{j,m}$ in just $O(n)$ time, which removes an $O(\log n)$ complexity factor from Algorithm~\ref{ch2algmass} (MASS algorithm).

However, we will not be able to benefit from the relationship between $QT_{j,i}$ and $QT_{j-1,i-1}$ when $j=1$ or $i=1$. 
This problem is easy to solve: we can simply pre-compute the dot product values in these two special cases with MASS algorithm in Algorithm~\ref{ch2algmass}. 
Concretely, we use $MASS(T^Q_{1,m}, T)$ to obtain the dot product vector when $j=1$, and we use $MASS(T_{1,m}, T^Q)$ to obtain the dot product vector when $i=1$. 
The two dot product vectors are stored in memory and used when needed.

We call this algorithm the STOMP algorithm, as it exploits the fact that we evaluate the distance profiles in-Order.
The algorithm is outlined in Algorithm~\ref{ch2algstomp}.

\begin{algorithm}[htb]
\caption{The STOMP algorithm.}
\label{ch2algstomp}
\begin{algorithmic}[1]
\Statex \textbf{Procedure} $STOMP(T,m)$ \vspace{-1em} 
\Statex \textbf{Input:} A user provided time series $T$ and interested subsequence length $m$ \vspace{-1em} 
\Statex \textbf{Output:} A matrix profile $P$ and matrix profile index $I$ \vspace{-1em} 
\State $n\gets Length(T)$\vspace{-1em} 
\State $M_T, \Sigma_T\gets ComputeMeanStd(T)$\Comment{\cite{rakthanmanon2012searching}} \vspace{-1em} 
\State $D, QT\gets MASS(T_{1,m}, T)$ \Comment{Algorithm~\ref{ch2algmass}}\vspace{-1em} 
\State $QT_{first}\gets QT$\vspace{-1em} 
\State $P\gets D$, $I\gets ones$\vspace{-1em}
\For{$i\gets1$ \textbf{to} $n-m$}\vspace{-1em} 
\For{$j\gets n-m$ \textbf{down to} $1$} \Comment{Equation~\ref{ch2eqqtupdate}}\vspace{-1em} 
\State $QT[j]\gets QT[j-1] - T[i-1] \times T[j-1] + T[i+m-1] \times T[j+m-1]$  \vspace{-1em} 
\EndFor  \vspace{-1em} 
\State $QT[0]\gets QT_{first}[i]$  \vspace{-1em} 
\State $D\gets CalculateDistanceProfile(QT,i,M_T, \Sigma_T)$ \Comment{Equation~\ref{ch2eqqt2}} \vspace{-1em} 
\State $P, I\gets ElementWiseMin(P, I, D, i)$ \vspace{-1em} 
\EndFor \vspace{-1em} 
\State \textbf{return} $P, I$
\end{algorithmic}
\end{algorithm}

The algorithm begins in Line 1 by evaluating the time series length $n$.
In line 2, the mean and standard deviation for each subsequence is computed with the technique introduced in~\cite{rakthanmanon2012searching}.
Line 3 calculates the first distance profile and stores the corresponding dot product in vector $QT$. 
Note that in lines 3, we require the similarity search algorithm (i.e., Algorithm~\ref{ch2algmass}) to not only return the distance profile $D$, but also the vector $QT$ in its first line. 
In Line 4 the algorithm store the dot product in another vector $QT_{first}$ for later use. 
Line 5 initializes the matrix profile and matrix profile index. 
The loop in lines 6-13 evaluates the distance profiles of the subsequences of $T$ in sequential order, with lines 7-9 updating $QT$ according to the mathematical formula in Equation~\ref{ch2eqqtupdate}. 
Line 10 updates $QT[i]$ with the pre-computed result from line 4. 
Finally, lines 10-12 evaluate distance profile and update matrix profile.

The time complexity of STOMP is $O(n^2)$; thus, we have an achieved a $O(\log n)$ factor speedup over STAMP. 
Moreover, it is clear that $O(n^2)$ is optimal for any full-join algorithm in the general case. 
The $O(\log n)$ speedup clearly make little difference for small datasets, for instance those with just a few tens of thousands of datapoints. 
However, as we tackle the datasets with millions of datapoints, something on the wish list of seismologists for example~\cite{beroza2016personal, yoon2015earthquake} this $O(\log n)$ factor begins to produce a very useful order-of-magnitude speedup.

As noted above, unlike the STAMP algorithm, STOMP is not really a good anytime algorithm, even though in principle we could interrupt it at any time and examine the current best-so-far matrix profile. 
The problem is that closest pairs (i.e. the motifs) we are interested in might be clustered at the end of the ordered search, defying the critical \emph{diminishing returns} property~\cite{zilberstein1995approximate}.
In contrast, STAMP's random search policy will, with very high probability, stumble on these motifs early in the search.

In fact, it may be possible to obtain the best of both worlds in meta-algorithm by interleaving periods of STAMP's random search with periods of STOMP's faster ordered search. 
This meta-algorithm would be slightly slower than pure STOMP, but would have the anytime properties of STAMP. 
For brevity we leave a fuller discussion of this issue to future work. 

\subsection{The STOMP\emph{I} Algorithm}
Up to this point we have discussed the batch version of matrix profile. 
By batch, we mean that the STAMP/STOMP algorithms need to see the entire time series $T$ before creating the matrix profile. 
However, in many situations it would be advantageous to build the matrix profile incrementally. 
Given that we have performed a batch construction of matrix profile, when new data arrives, it would clearly be preferable to incrementally adjust the current profile, rather than starting from scratch.

Because the matrix profile solves both the times series motif and the time series discord problems, an incremental version of STAMP/STOMP would automatically provide the first incremental versions of both these algorithms. 
In this section, we demonstrate that we \emph{can} create such an incremental algorithm.

By definition, an incremental algorithm sees data points arriving one-by-one in sequential order, which makes STOMP a better starting point than STAMP. 
Therefore we name the incremental algorithm STOMP\emph{I} (STOMP \emph{I}ncremental). 
For simplicity and brevity, Algorithm~\ref{ch2algstompi} only shows the algorithm to incrementally maintain the self-similarity join. 
The generalization to general similarity joins is obvious.

\begin{algorithm}[htb]
\caption{The STOMP\emph{I} algorithm.}
\label{ch2algstompi}
\begin{algorithmic}[1]
\Statex \textbf{Procedure} $STOMPI(T,m,t,P,I,QT,M_T,\Sigma_T)$ \vspace{-1em} 
\Statex \textbf{Input:} The original time series $T$, subsequence length $m$, a new data point $t$ \vspace{-1em} following $T$, the matrix profile $P$ and its associated matrix profile index $I$ of $T$, dot \vspace{-1em} product vector $QT$, mean vector $M_T$ and standard deviation vector $\Sigma_T$ \vspace{-1em} 
\Statex \textbf{Output:} The updated matrix profile $P_{new}$ and its matrix profile index $I_{new}$ \vspace{-1em} corresponding to the new time series $T_{new}$, the updated dot product vector $QT_{new}$, updated \vspace{-1em} mean vector $M_{T,new}$, and standard deviation vector $\Sigma_{T,new}$ \vspace{-1em} 
\State $n\gets Length(T), l\gets n-m+1, T_{new}\gets [T, t], S\gets T_{new}[l+1:n+1]$\vspace{-1em} 
\State $t_{drop}\gets T_{new}[l]$\Comment{$t_{drop}$ is the first item of the last subsequence of $T$}\vspace{-1em}
\For{$i\gets l$ \textbf{down to} $1$} \Comment{Equation~\ref{ch2eqqtupdate}}\vspace{-1em} 
\State $QT_{new}[j]\gets QT[j-1] - T_{new}[i-1] \times t_{drop} + T_{new}[i+m-1] \times t$  \vspace{-1em} 
\EndFor  \vspace{-1em} 
\For{$i\gets 0$ \textbf{to} $m-1$} \Comment{calculate the first dot product with simple brute-force}\vspace{-1em} 
\State $QT_{new}[0]\gets T_{new}[i]\times S[i]$  \vspace{-1em} 
\EndFor  \vspace{-1em} 
\State $\mu_{new} \gets M_T[l] + (t - t_{drop}) / m$  \Comment{update mean of $S$}\vspace{-1em} 
\State $\sigma_{new} \gets \Sigma_T[l]^2 + M_T[l]^2 + (t^2 - t^2_{drop}) / m - \mu^2_{new}$  \Comment{update standard deviation of $S$}\vspace{-1em} 
\State $M_{T,new} \gets [M_T, \mu_{new}]$, $\Sigma_{T,new} \gets [\Sigma_T, \sigma_{new}]$ \vspace{-1em} 
\State $D\gets CalculateDistanceProfile(QT_{new},M_{T,new}, \Sigma_{T,new})$ \Comment{Equation~\ref{ch2eqqt2}} \vspace{-1em} 
\State $P, I\gets ElementWiseMin(P, I, D[0:l], l)$ \vspace{-1em}
\State $p_{last}, i_{last}\gets FindMin(D)$ \vspace{-1em}
\State $P_{new}\gets [P, p_{last}], I_{new}\gets [I, i_{last}]$ \vspace{-1em}
\State \textbf{return} $P_{new}, I_{new}, QT_{new}, M_{T,new}, \Sigma_{T,new}$
\end{algorithmic}
\end{algorithm}

As a new data point $t$ arrives, the size of the original time series $T$ increases by one. 
We denote the new time series as $T_{new}$, and we need to update the matrix profile $P$ to $P_{new}$ and its associated matrix profile index $I$ to $I_{new}$. 
For clarity, note that the input variables $QT$, $M_T$ and $\Sigma_T$ are all vectors, where $QT[i]$ is the dot product of the $i$th and last subsequences of $T$; $M_T[i]$ and $\Sigma_T[i]$ are, respectively, the mean and standard deviation of the $i$th subsequence of $T$.

\newpage
In line 1, $S$ is a new subsequence generated at the end of $T_{new}$. 
Lines 2-5 evaluate the new dot product vector $QT_{new}$ according to Equation~\ref{ch2eqqtupdate}, where $QT_{new}[i]$ is the dot product of $S$ and the $i$th subsequence of $T_{new}$. 
Note that the length of $QT_{new}$ is one item longer than that of $QT$. 
The first dot product $QT_{new}[0]$ is a special case where Equation~\ref{ch2eqqtupdate} is not applicable, so lines 6-9 calculate with simple brute-force. 
In lines 10-12 we evaluate the mean and standard deviation of the new subsequence $S$, and update the vectors $M_{T,new}$ and $\Sigma_{T,new}$. 
After that we calculate the distance profile $D$ with regard to $S$ and $T_{new}$ in line 13. 
Then, similar to STAMP/STOMP, line 14 performs a pairwise comparison between every element in $D$ and the corresponding element in $P$ to see if the corresponding element in $P$ needs to be updated. 
Note that we only compare the first $l$ elements of $D$ here, since the length of $D$ is one item longer than that of $P$. 
Line 15 finds the nearest neighbor of $S$ by evaluating the minimum value of $D$. 
Finally, in line 16, we obtain the new matrix profile and associated matrix profile index by concatenating the results in line 14 and line 15.

The time complexity of the STOMPI algorithm is $O(n)$ where $n$ is the length of size of the current time series $T$. 
Note that as we maintain the profile, each incremental call of STOMP\emph{I} deals with a one-item longer time series, thus it gets \emph{very} slightly slower at each time step. 
Therefore, the best way to measure the performance is to compute the Maximum Time Horizon (MTH), in essence the answer to this question: ``\emph{Given this arrival rate, how long can we maintain the profile before we can no longer update fast enough?}''

Note that the subsequence length $m$ is not considered in the MTH evaluation, as the overall time complexity of the algorithm is $O(n)$, which is independent of $m$. 
We have computed the MTH for two common scenarios of interest to the community.

\begin{itemize}
\item \textbf{House Electrical Demand}~\cite{murray2015data}: 
This dataset is updated every eight seconds. 
By iteratively calling the STOMP\emph{I} algorithm, we can maintain the profile for at least twenty-five years. 
\item \textbf{Oil Refinery}:  
Most telemetry in oil refineries and chemical plants is sampled at once a minute~\cite{tucker2004bayesian}.  
The relatively low sampling rate reflects the ``inertia'' of massive boilers/condensers. 
Even if we maintain the profile for 40 years, the update time is only around 1.36 seconds. 
Moreover, the raw data, matrix profile and index would only require 0.5 gigabytes of main memory. 
Thus the MTH here is forty-plus years.
\end{itemize}

For both these situations, given projected improvements in hardware, these numbers effectively mean we can maintain the matrix profile forever.

As impressive as these numbers are, they are actually quite pessimistic. 
For simplicity we assume that \emph{every} value in the matrix profile index will be updated at each time step. 
However, empirically, much less than 0.1\% of them need to be updated. 
If it is possible to \emph{prove} an upper bound on the number of changes to the matrix profile index per update, then we could greatly extend the MTH, or, more usefully, handle much faster sampling rates. 
We leave such considerations for future work.

\subsection{STAMP and STOMP Allow a Perfect ``Progress Bar"}
Both the STAMP and STOMP algorithms have an interesting property for a motif discovery/join algorithm, in that they both take deterministic and predicable time. 
This is very unusual and desirable property for an algorithm in this domain. 
In contrast, the two most cited algorithms for motif discovery~\cite{li2015quick, mueen2009exact}, while they can be fast on average, take an unpredictable amount of time to finish. 
For example, suppose we observe that either of these algorithms takes exactly one hour to find the best motif on a particular dataset with $m = 500$ and $n = 100,000$. 
Then the following are all possible:

\begin{itemize}
\item Setting $m$ to be a single data point shorter (i.e. $m = 499$), could increase or decrease the time needed by over an order of magnitude.
\item Decreasing the length of the dataset searched by a single point (that is to say, a change of just 0.001\%), could increase or decrease the time needed by over an order of magnitude.
\item Changing a \emph{single} value in the time series (again, changing only 0.001\% of the data), could increase or decrease the time needed by over an order of magnitude~\cite{yeh2017dmkd}.
\end{itemize}

Moreover, if we actually made the above changes, we would have no way to know in advance how our change would impact the time needed. 

In contrast, for both STAMP and STOMP (assuming that $m \ll n$), given only $n$, we can predict how long the algorithm will take to terminate, completely independent of the value of $m$ and the data itself. 

To do this we need to do one calibration run on the machine in question. 
With a time series of length $n$, we measure $\delta$, the time taken to compute the matrix profile.
Then, for any new length $n_{new}$, we can compute $\delta_{required}$ the time needed as:
\begin{equation}
\label{ch2eqprogress}
\delta_{required} = \frac{\delta}{n^2} \times n^2_{new}
\end{equation}
So long as we avoid trivial cases, such as that $m ~ n$ or $n_{new}$ and/or $n$ are very small, this formula will predict the time needed with an error of less than 5\%.

\subsection{Scalability of STAMP and STOMP}
Because the time performance of STAMP is independent of the data quality or any user inputs (there are none except the choice of $m$, which does not affect the speed), our scalability experiments are unusually brief; for example, we do not need to show how different noise levels or different types of data can affect the results.

\begin{figure}[htb]
\centering
\includegraphics[trim={10cm 6cm 10cm 6cm}, clip, width=1\textwidth,page=10]{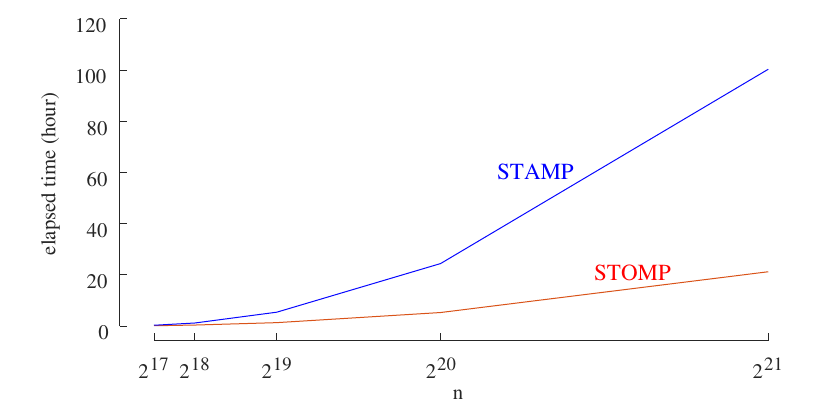}
\caption{
Time required for both STAMP and STOMP self-join with $m = 256$, varying~$n$.
}
\label{ch2figspeed1}
\end{figure}

In Figure~\ref{ch2figspeed1} we show the time required for a self-join with $m$ fixed to 256, for increasing long time series. 

\begin{figure}[htb]
\centering
\includegraphics[trim={10cm 6cm 10cm 6cm}, clip, width=1\textwidth,page=11]{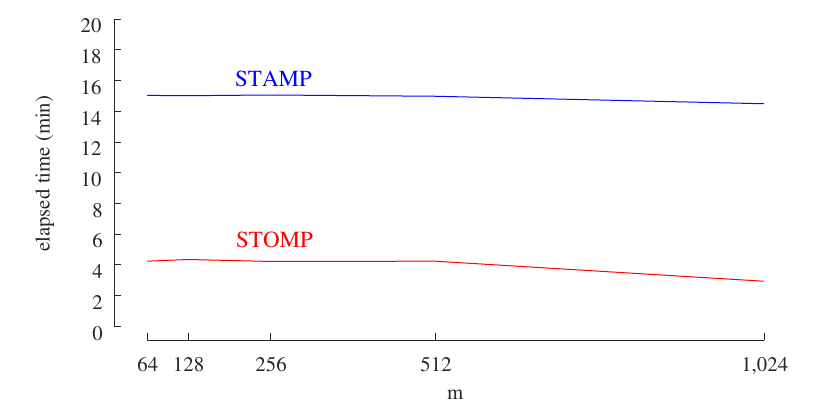}
\caption{
Time required for both STAMP and STOMP self-join with $n= 2^{17}$, varying~$m$.
}
\label{ch2figspeed2}
\end{figure}

In Figure~\ref{ch2figspeed2}, we show the time required for a self-join with $n$ fixed to $2^{17}$, for increasing long $m$. 
Again recall that unlike virtually all other time series data mining algorithms in the literature whose performance degrades for longer subsequences~\cite{ding2008querying, mueen2009exact} the running time of both STAMP and STOMP does not depend on $m$.

Note that the time required for the longer subsequences is slightly shorter. 
This is because the number of pairs that must be considered for a time series join is $(n-m)/2$, so as $m$ is becomes larger, the number of comparisons becomes slightly smaller.

We can further exploit the simple parallelizability of the algorithm by using four 16-core virtual machines on Microsoft Azure to redo the two-million join ($n = 2^{21}$ and $m = 256$) experiment. 
By scaling up the computational power, we have reduced the running time from 4.2 days to just 14.1 hours. 
This use of cloud computing required writing just few dozen lines of simple additional code.

In order to see the improvements of STOMP over STAMP, we repeated the last two sets of experiments. 
In Figure~\ref{ch2figspeed1} we also show the time required for a self-join with $m$ fixed to 256, for increasing long time series. 
As expected, the improvements are modest for smaller datasets, but much greater for the larger datasets, culminating in a $4.7X$ speedup for the ~2 million length time series.

In Figure~\ref{ch2figspeed2}, we show the time required for a self-join with $n$ fixed to $2^{17}$, for increasing long $m$. 
Once again note that the running time of STOMP does not depend on~$m$.

\section{Matrix Profile for Multidimensional Time Series}
\label{ch2secmpmts}
We can now explain our matrix profile algorithms for the more general case of multidimensional time series (Definition~\ref{ch2defmp}).
The modifications introduce in this section can extend either STAMP or STOMP.
Because our algorithm for multidimensional matrix profile improves upon our previous solutions, the fundamental intuition also stems from the relationship between distance profiles and the matrix profile.

\subsection{The \emph{m}STAMP Algorithm}
\label{ch2secmstamp}
Our definitions allow a na\"ive solution. 
We could compute the matrix profile (the multidimensional variant using all dimensions~\cite{silva2016ismir,silva2018fast}) to all $d$ choose $k$ combinations of dimensions and choose the best one under some ranking function. 
However, this na\"ive solution is only computable for trivially small datasets due to the combinatorial explosion inherent in this approach.

Fortunately, the combinatorial search space can be searched efficiently and admissibly in a greedy fashion. 
Our algorithm can compute the $k$-dimensional matrix profile for every possible setting of $k$ (i.e., 1 to $d$) simultaneously in $O(d \log ⁡d n^2)$  time and $O(dn)$ space. 
The algorithm is outlined in Algorithm~\ref{ch2algmstamp}.
We choose to extend the STAMP algorithm; however, the same modification can be trivially applied to the STOMP algorithm.
To simplify the presentation, we omit the operations related to storing of the $k$-dimensional matrix profile subspace. 
Before explaining the algorithm, we note that the source code of multidimensional STAMP (\emph{m}STAMP) and multidimensional STOMP (\emph{m}STOMP) in both MATLAB and Python is publicly available in~\cite{mstampcode} and that the correctness of the algorithm is formally demonstrated in Section~\ref{ch2seccorret}.

\begin{algorithm}[htb]
\caption{The \emph{m}STAMP algorithm.}
\label{ch2algmstamp}
\begin{algorithmic}[1]
\Statex \textbf{Procedure} $mSTAMP(\mathbf{T},m)$ \vspace{-1em} 
\Statex \textbf{Input:} Inputted time series $\mathbf{T}\in \mathbb{R}^{d\times n}$, interested subsequence length $m\in \mathbb{Z}$ \vspace{-1em} 
\Statex \textbf{Output:} A set of $k$-dimensional matrix profile $\mathbf{P}\in \mathbb{R}^{d\times n-m+1}$ \vspace{-1em} 
\State $\mathbf{P} \gets$ size $d\times n-m+1$ inf matrix\vspace{-1em}
\State $idxes \gets$ integers from $0$ to $n-m$ \vspace{-1em}
\For{\textbf{each} $idx$ \textbf{in} $idxes$}\Comment{random order if anytime}\vspace{-1em}
\State $\mathbf{D}\gets $ size $d\times n-m+1$ zero matrix\vspace{-1em}
\For{$i\gets 0$ \textbf{to} $d$} \vspace{-1em}
\State $Q\gets \mathbf{T}[i, idx:idx+m]$ \vspace{-1em}
\State $\mathbf{D}[i, :]\gets DistanceProfile(Q,T[i,:])$ \vspace{-1em}
\EndFor \vspace{-1em}
\State \vspace{-1em}
\State $\mathbf{D}\gets ColumnWiseAscendingSort(\mathbf{D})$ \vspace{-1em}
\State $D' \gets$ length $n-m+1$ zero array \vspace{-1em}
\For{$i\gets 0$ \textbf{to} $d$} \vspace{-1em}
\State $D' \gets D'+\mathbf{D}[i,:]$ \vspace{-1em}
\State $D" \gets D'/ (i + 1)$ \vspace{-1em}
\State $\mathbf{P}[i,:] \gets ElementWiseMin(\mathbf{P}[i,:], D")$ \vspace{-1em}
\EndFor \vspace{-1em}
\EndFor  \vspace{-1em} 
\State \textbf{return} $\mathbf{P}$
\end{algorithmic}
\end{algorithm}

In line 1, the memory for the $k$-dimensional matrix profile for each setting of $k$ is allocated and initialized as an array filled with infinity. 
For each iteration in the main loop (line 3 to line 17), we select one subsequence from $\mathbf{T}$ as the query for further processing.
The subsequences are selected in a random order if the anytime-algorithm property is desired~\cite{yeh2016icdm, yeh2017dmkd}. 
From line 5 to line 8, the dimension-wise distance profile using the query and $\mathbf{T}$ is computed and stored in matrix $\mathbf{D}$. 
If the query is selected in a random order, MASS~\cite{masspage} is used for the distance profile computation; otherwise, the method proposed by Zhu et al.~\cite{zhu2016icdm} is used for distance profile computation.
This is because that method (with time complexity of $O(n)$) is faster than MASS (with time complexity of $O(n \log ⁡n)$) but requires the subsequences to be selected in order (line 3), which nullifies the anytime-algorithm property.  
Next, in line 10, a column-wise sort in ascending order is applied to the matrix $\mathbf{D}$. 
Finally, from line 12 to line 15, each $k$-dimensional matrix profile is updated with the corresponding $k$-dimensional distance profile (i.e., $D"$) if the corresponding element in $D"$ is smaller.

\subsection{Demonstration of Correctness}
\label{ch2seccorret}
The basic strategy of \emph{m}STAMP is simple.
In each iteration of the main loop (line 3 to line 17 in Algorithm~\ref{ch2algmstamp}), the algorithm computes the $k$-dimensional distance profile for a given subsequence under every possible setting of $k$ (from 1 to $d$). 
Therefore, it is sufficient to justify the algorithm's overall correctness by demonstrating the correctness of the computed $k$-dimensional distance profile.

Given a multidimensional subsequence $\mathbf{T_{i,m}}$ and its parent time series $\mathbf{T}$, the algorithm first computes the distance profiles for each dimension independently and stores them in matrix $\mathbf{D}$ (line 4 to line 8 in Algorithm~\ref{ch2algmstamp}). 
In other words, the $(l,j)$ position of $\mathbf{D}$ stores the distance between  $T_{i,m}^{(l)}$ and $T_{j,m}^{(l)}$. 
Note that each row of $\mathbf{D}$ is the dimension-wise distance profile (Definition~\ref{ch2defdp}) instead of the $k$-dimensional distance profile (Definition~\ref{ch2defkdp}). 
Na\"ive, the $k$-dimensional distance profile can be produced by solving $\min_X⁡ \|\mathbf{D}[j,X]\|_0  \forall j\in[0,2,…,n-m]$ for each setting of $k$, where $X$ is an indicator vector that shows which dimensions are included ($\|X\|_0=k$). 
However, computing the $k$-dimensional distance by enumerating all possible combination would be extremely inefficient.

Because the $z$-normalized Euclidean distance is non-negative, every number in $\mathbf{D}$ is non-negative. 
By taking this fact into account, the 1-dimensional distance profile is the smallest value in each column of $\mathbf{D}$, the 2-dimensional distance profile is the two smallest values in each column of $\mathbf{D}$, and the rest can be solved trivially after $\mathbf{D}$ is sorted column-wise. 
As a result, applying column-wise sort (line 10 in Algorithm~\ref{ch2algmstamp}) and column-wise cumulative sum (line 13 in Algorithm~\ref{ch2algmstamp}) to $\mathbf{D}$ can produce the $k$-dimensional distance profile. 
Therefore, the algorithm ultimately computes the correct $k$-dimensional matrix profile.

\subsection{The Expressiveness of Multidimensional Matrix Profile}
\begin{figure}[htb]
\centering
\includegraphics[trim={9cm 7cm 9cm 7cm}, clip, width=1\textwidth,page=12]{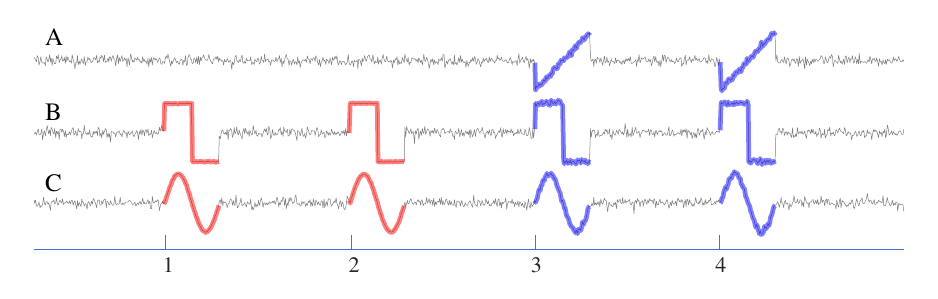}
\caption{
When the 2-dimensional motif and 3-dimensional motif are extracted using the multidimensional matrix profile, the 2-dimensional motif may or may not be a subset of the 3-dimensional motif.
In this example, the motif with lower dimensionality is not a subset of the higher dimensional motif.
}
\label{ch2figexpress}
\end{figure}

With the correctness of the algorithm demonstrated, now we are ready to discuss the expressiveness of the discovered multidimensional motifs. 
It may seem counterintuitive, but as demonstrated in Figure~\ref{ch2figexpress},  the lower dimensional motif may or may not be a subset of the higher dimensional motif, since the lower dimensional motif pair could be closer than any subset of dimensions in the higher dimensional motif pair.

For clarity, here the best 3-dimensional motif pair is the patterns occurring at times `3' and `4' of all three time series, but the best 2-dimensional motif pair is the patterns occurring at times `1' and `2' of just B and C.

This property is unfortunate, since it excludes the possibility to use various pruning and dynamic programming techniques to speed up the computation. 
However, as we will see, it is this expressiveness that allows the discovery of semantically meaningful motifs in high-dimensional data. 

\subsection{Scalability of \emph{m}STAMP and \emph{m}STOMP}
The multidimensional matrix profile algorithm can be built on top of either the STAMP or STOMP algorithm; therefore, it inherits all the positive characteristics from its parent algorithm, including:

\begin{itemize}
\item The runtime does not depend on data's properties (noise, stationarity, periodicity etc.), only it length $n$.
\item The runtime does not depend on the \emph{subsequence} length, $m$.
\item The algorithm is easy to parallelize.
\item The algorithm can be cast as an anytime algorithm~\cite{yeh2016icdm, yeh2017dmkd}. 
\end{itemize}

To empirically confirm the aforementioned characteristics, we have performed a scalability test.
All experiments are performed on a server with Intel(R) Xeon(R) CPU E5-2620 v3 @ 2.40GHz, and the algorithm is implemented with MATLAB. 
However, we also have a Python version of our algorithm freely available in~\cite{mstampcode}.

\begin{figure}[htb]
\centering
\includegraphics[trim={9cm 7cm 9cm 7cm}, clip, width=1\textwidth,page=13]{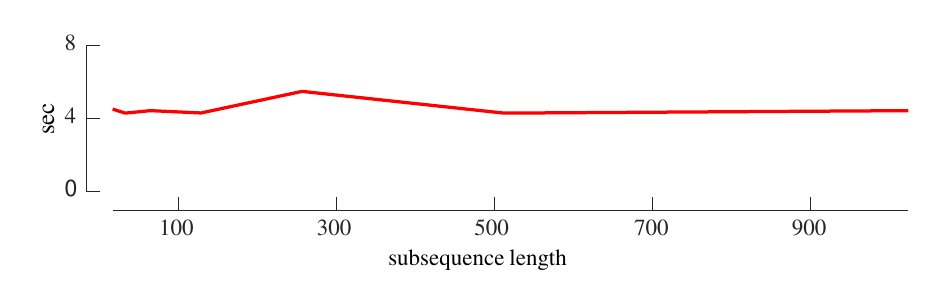}
\caption{
The runtime does not vary significantly as we change the subsequence length.
}
\label{ch2figscalability1}
\end{figure}

We begin by testing scalability of the \emph{m}STOMP algorithm with a randomly generated 4-dimensional time series of length $2^{14}$ with multiple subsequence lengths.
The resulting runtimes are shown in Figure~\ref{ch2figscalability1}. 
Unsurprisingly, the change of subsequence length does not impact the runtime, concurring with the claims of both STAMP~\cite{yeh2016icdm, yeh2017dmkd} and STOMP~\cite{yeh2017dmkd, zhu2016icdm, zhu2018exploiting}.

Before moving on, it is worth reminding ourselves how remarkable and unexpected this property is. 
We can perform motif search with complete freedom from the curse of \emph{dimensionality} (unlike everywhere else in this section, here the term dimensionality is used to denote subsequence length) that plagues all other approaches~\cite{balasubramanian2016discovering, berlin2012detecting, patel2002mining, tanaka2005discovery, vahdatpour2009toward}.

\begin{figure}[htb]
\centering
\includegraphics[trim={9cm 7cm 9cm 7cm}, clip, width=1\textwidth,page=14]{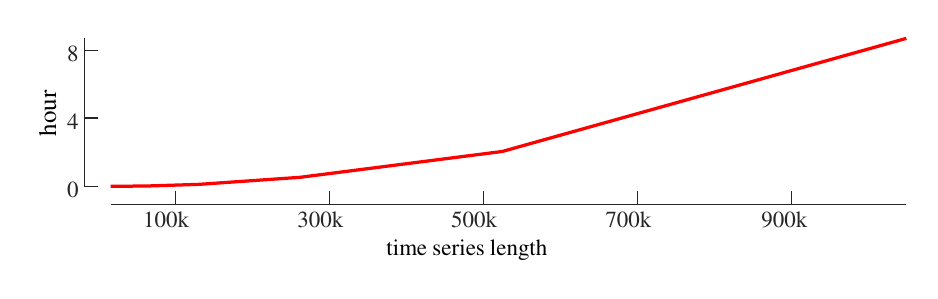}
\caption{
The runtime increases quadratically with the length of the time series.
}
\label{ch2figscalability2}
\end{figure}

\newpage
Next, we fix the subsequence length to 256 and test the \emph{m}STOMP on a 4-dimensional time series of increasing lengths. 
As shown in Figure~\ref{ch2figscalability2}, the runtime grows quadratically with time series length, which coincides with the claimed time complexity of the parent algorithm, STOMP~\cite{yeh2017dmkd, zhu2016icdm, zhu2018exploiting}.

Before further mitigating this time complexity, it is worth noting that it may already be fast enough for most applications.
For example, an oil distillation column may have four dimensions, say \texttt{[TEMP, PRESS, FLOW-RATE, REFLUX-RATE]} and be sampled once a minute. 
Figure~\ref{ch2figscalability2} indicates that it will take about two hours of CPU time to find motifs in a full year of historical data (525,600 data points). 
This is almost certainly acceptable in this domain; given the potential cost savings an actionable motif could lead to.

\begin{figure}[htb]
\centering
\includegraphics[trim={9cm 6cm 9cm 6cm}, clip, width=1\textwidth,page=15]{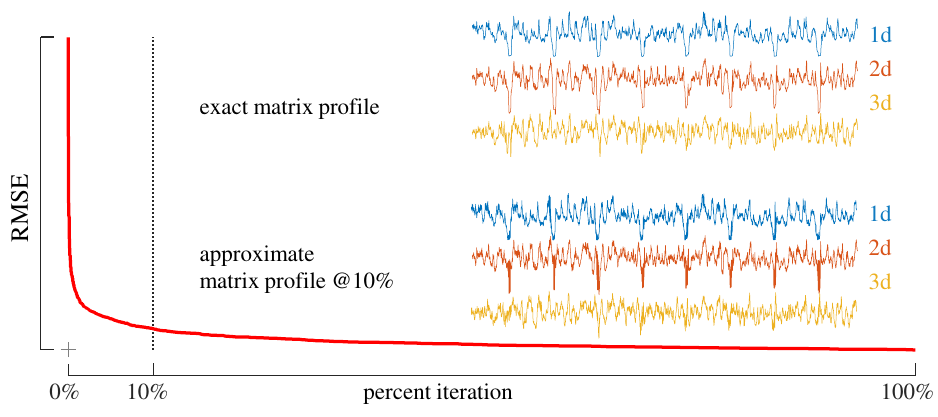}
\caption{
Like its parent STAMP, \emph{m}STAMP converges quickly. 
The approximated multidimensional matrix profile achieves a low Root Mean Square Error (RMSE) when just 10\% of the iteration are completed. 
The inset images are the multidimensional matrix profiles.
}
\label{ch2figanytime2}
\end{figure}

Nevertheless, we can offer the user a further significant speed-up by processing the data in an anytime fashion. 
Like one of its parent algorithm STAMP~\cite{yeh2016icdm, yeh2017dmkd}, \emph{m}STAMP can be trivially modified to be an effective anytime algorithm. 
Figure~\ref{ch2figanytime2} shows the convergence rate of \emph{m}STAMP on a 3-dimensional time series with 2-dimensional motifs embedded. 
The Root Mean Squared Error (RMSE) decreases quickly in the first few percent of iterations. 
After only 10 percent of the computations have been completed, the current ``best-so-far" matrix profile is not only visually similar to the exact matrix profile (the inset images in Figure~\ref{ch2figanytime2}), but the RMSE is also very low. 
This property is useful for interactive data exploration as the user can terminate the algorithm early when satisfied by the discovered motifs using the current approximate matrix profile~\cite{yeh2016icdm, yeh2017dmkd}.

\begin{figure}[htb]
\centering
\includegraphics[trim={9cm 7cm 9cm 7cm}, clip, width=1\textwidth,page=16]{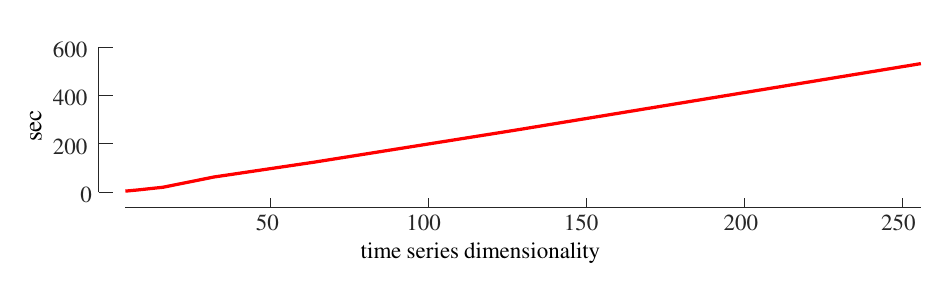}
\caption{
The runtime increases linearithmically with the dimensionality of time series.
}
\label{ch2figscalability3}
\end{figure}

Because the input time is multidimensional, we need to test the scalability of \emph{m}STAMP when we vary the dimensionality of the input time series. 
Here, we fixed the time series length to $2^{14}$ and subsequence length to 256. 
The runtime shown in Figure~\ref{ch2figscalability3} confirms our claim in Section~\ref{ch2secmstamp} as the runtime has a linearithmic relationship with the time series dimensionality.

\section{Discussion and Conclusion}
\label{ch2secconclude}
In this chapter, we have defined matrix profile for both single dimensional and multidimensional time series.
We have introduced several scalable algorithms for matrix profile. 
Our algorithms are simple, fast, parallelizable and parameter-free, and can be incrementally updated for moderately fast data arrival rates.
We will showcase applications of matrix profile in motif discovery (Chapter~\ref{ch3}), weakly labeled time series classification (Chapter~\ref{ch4}), and representation learning (Chapter~\ref{ch5}).
The information regarding other applications of matrix profile like music information retrieval~\cite{silva2018summarizing,silva2016ismir,silva2018fast}, time series data visualization~\cite{yeh2016icdm2}, semantic segmentation~\cite{gharghabi2017icdm, gharghabi2018dmkd}, time series chain discovery~\cite{zhu2017icdm}, augmented time series motif discovery~\cite{dau2017kdd}, shapelet discovery~\cite{yeh2017dmkd}, and variable-length motif discovery~\cite{linardi2018matrix} can be found in their corresponding papers.
Our code, including MATLAB interactive tools, a C++ version, and a Python version, are freely available for the community to confirm, extend, and exploit our ideas.
The interested reader may refer to the matrix profile project website~\cite{mppage} for more information.

\chapter{Matrix Profile for Motif Discovery}
\label{ch3}
Time series motifs are approximately repeating patterns in real-value data, Figure~\ref{ch3figtoy} shows some examples highlighted in the top two time series. 
They are useful in exploratory data mining. 
If a time series pattern is \emph{conserved}, we may assume that there is some high-level atomic mechanism/behavior that causes that pattern to be conserved. 
That behavior may be desirable in certain cases (e.g., a perfect badminton shot~\cite{tanaka2005discovery}) or undesirable in others (e.g., the cough of a sick pig~\cite{exadaktylos2008time}). 
In either case, the discovery of motifs is often the first step in various kinds of higher-level time series analytics~\cite{yeh2016icdm, yeh2017dmkd}.

\begin{figure}[htb]
\centering
\includegraphics[trim={12.5cm 7.5cm 12.5cm 7.5cm}, clip, width=0.8\textwidth,page=1]{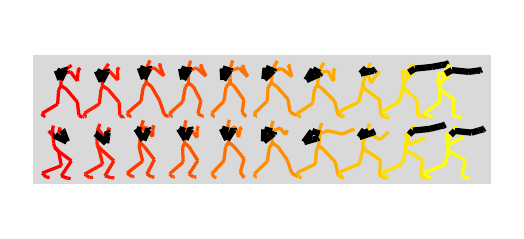}
\caption{
Two motion-capture traces~\cite{cmumocap}. 
While the right-hand punch is nearly identical in both moves, the boxer in the top trace is throwing a cross. 
In contrast, the boxer in the bottom trace is throwing a one-two combo. 
A video of these motions is available in~\cite{mmpwebsite}.
}
\label{ch3figcmu}
\end{figure}

\begin{figure}[htb]
\centering
\includegraphics[trim={12.5cm 6cm 12.5cm 6cm}, clip, width=0.8\textwidth,page=2]{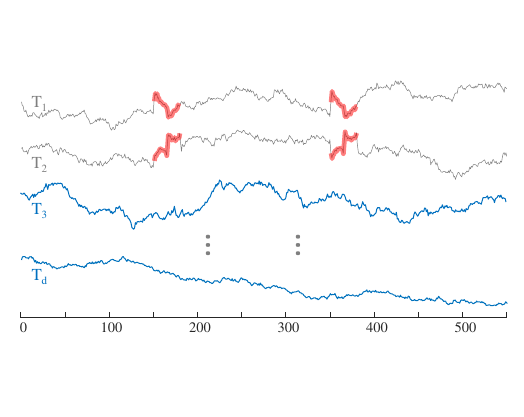}
\caption{
A running example of a multidimensional time series. 
Both of the first two dimensions have a motif of length 30 embedded at locations 150 and 350. 
All remaining time series (just two are shown above) are simply random walks.
}
\label{ch3figtoy}
\end{figure}

Since the introduction of the first motif discovery algorithm for univariate time series in 2002~\cite{patel2002mining}, many researchers have generalized motifs to the multidimensional case~\cite{balasubramanian2016discovering, berlin2012detecting, tanaka2005discovery, vahdatpour2009toward}.
However, almost all of these efforts attempt to find motifs on \emph{all} dimensions. 
We believe that using all dimensions will generally not produce meaningful motifs, except in the most contrived situations. 
To see this in an intuitive setting, consider Figure~\ref{ch3figcmu}.

If we focus solely on the boxer's dominant hand, the two behaviors are almost identical. 
However, if we look that the full set of Mo-Cap markers on all of the limbs, the differences in the non-dominant hand and in the footwork ``swamp" the similarity of the punch, making this repeated behavior impossible to find with the classic multidimensional motif discovery algorithms, that use all the available dimensions~\cite{balasubramanian2016discovering, berlin2012detecting, tanaka2005discovery, vahdatpour2009toward}.

To see why this is, consider the multidimensional time series shown in Figure~\ref{ch3figtoy} (we will formalize our definitions in Section~\ref{ch2secdef}).

If we run the classic \emph{single dimensional} motif discovery [26] on either of the first two dimensions, we correctly find the visually obvious motifs at locations 150 and 350. 
If we generalize the motif definition to \emph{Multidimensional Time Series data} (MTS), and consider the best motif in the two dimensions $\{T1,T2\}$, then unsurprisingly, we still find the same best motif location. 
However, what will happen as we add in some random walks to the \emph{multidimensional} dataset we consider? 
With just one random walk added to create a three-dimensional times series, we can still robustly find the correct motif locations; the signal of the true subset $\{T1,T2\}$ is strong enough to resist the irrelevant information added by a single random walk. 
However, empirically averaging over 100 trials, we have found that if there are eight additional irrelevant dimensions, then we do about as well as random chance. 
Moreover, the above motifs make up about 5\% of the data. However, motifs are often much rarer, which accelerates the rate at which increasing dimensionality masks motifs that exist in a subspace of the data.

This illustrates a problem that is ubiquitous in medicine, science, and industry. 
The analyst suspects that there are motifs in some subset of the time series, but does not know \emph{which} dimensions are involved, or even \emph{how many} dimensions are involved. 
Doing motif search on all dimensions is almost guaranteed to produce meaningless results, even if a subset of dimensions has clear and unambiguous motifs.

Informally, we would like any multidimensional motif framework to be able to support all the following types of queries. 
Given a large $k$-dimensional time series:

\begin{itemize}
\item \textbf{Guided Search}: 
Find the best motif on $k$ dimensions, where the integer $k$ is given by the user, but which $k$ dimensions to use is unspecified.
\item \textbf{Constrained Search}: 
Find the best motif on $k$ dimensions, but explicitly include (or exclude) a given subset of dimensions.
\item \textbf{Unconstrained Search}: 
Find the best motif on $k$ dimensions, where $k$ is not given by the user but is the ``natural" subset of the data that has motifs.
\end{itemize}

The first two tasks mostly reduce to questions of speed and scalability; the last task is subtler, requiring us rank different tentative solutions and return the most natural one.

The need for such tools is based on our collaborations with domain experts. 
For example, in the oil and gas industry, a single distillation column typically has well over a hundred time series (\emph{Tags}, in the parlance of the industry) monitoring various aspects of the system. 
However, motifs typically appear in just a handful of dimensions. As a concrete example, consider a known motif known to appear on distillation columns in Texas. 
Between April and September, Texas often has brief thunderstorms with large amounts of rain falling within short periods of time. 
This falling rain cools the distillation column, reducing the pressure inside, and invokes a change in flowrate, or some other part of the system that attempts to compensate for the reduced pressure. 
Thus the ``rainstorm" motif may \emph{only} show up on the \texttt{$\{$temp, pressure, flowrate$\}$} tags.

Before leaving this example, it is worth noting that the important dimensions for the motif depend on the user-specified motif length. 
In such datasets, a motif query of one hour may turn up the thunderstorm example, but a motif query of length one day may find the motif representing a monthly calibration/cleaning run, which affects many more dimensions.

The rest of the chapter is organized as follows.
In Section~\ref{ch3secrelated} we discuss related work. 
Section~\ref{ch3seframework} outlines our matrix profile based motif discovery framework.
Then, we provide a rigorous empirical evaluation and case studies in Section~\ref{ch3secexp}.
Finally, in Section~\ref{ch3secconclude}, we offer conclusions and directions for future work.

\section{Background and Related Work}
\label{ch3secrelated}
There is a large and growing body of work on single time series motif discovery~\cite{patel2002mining, yeh2016icdm2, yeh2016icdm, yeh2017dmkd}; however, there is much less work on the multidimensional case~\cite{balasubramanian2016discovering, minnen2007detecting, tanaka2005discovery, vahdatpour2009toward}.

The work of Minnen et al.~\cite{minnen2007detecting} is the closest in spirit to our work.
Their work was the first to note the detrimental impact of irrelevant dimensions on multidimensional motif search, and they introduced a method that is shown to be somewhat robust for a small number of \emph{smooth}, but irrelevant dimensions, or \emph{just} one noisy irrelevant dimension. 
However, the algorithm introduced is \emph{approximate}. 
Even in an ideal case, with just six dimensions, they report ``\emph{with no noise, (our approach) achieves roughly 80\% accuracy}". 
We want to consider much higher dimensionalities, with a much greater fraction of irrelevant dimensions, and we are unwilling to compromise accuracy. 
The work was notable at the time for being much faster than a brute-force search, but since the advent of the Matrix Profile, that advantage has narrowed or disappeared \cite{yeh2016icdm, yeh2017dmkd}.

Tanaka et al.~\cite{tanaka2005discovery} propose to perform multidimensional motif discovery by ``\emph{transforming multi-dimensional time-series data into 1-dimensional time-series data}". 
The idea is attractive for its simplicity, but it requires all (or at least \emph{most}) of the dimensions to be relevant, as the algorithm is brittle to even a handful of irrelevant dimensions. ddd
Moreover, both the speed and accuracy of Tanaka's algorithm depend on careful tuning of five parameters.

In a series of papers, Vahdatpour and colleagues introduce an MTS motif discovery tool and apply it to a variety of medical monitoring applications~\cite{vahdatpour2009toward}. 
Their approach is based on computing time series motifs for each individual dimension and using clustering to ``stitch" together various dimensions. 
However, even when the motifs are quite obvious, the problems are small and simple, and at most three irrelevant dimensions are considered, they never achieved greater than 85\% accuracy on the three domains in which they were tested. 
To be sure, this is much better than the 17\% they achieve with the strawman of only considering a single dimension. 
But given that seven parameters need to be tuned to achieve this result, accuracy is likely to be further compromised in more challenging data sets.

It is worth restating that the multidimensional motif discovery algorithms in which we are aware have the weakness of being \emph{approximate}. 
For example,~\cite{balasubramanian2016discovering, minnen2007detecting, tanaka2005discovery} and~\cite{vahdatpour2009toward} all achieve scalability by searching over a reduced time resolution/reduced cardinality symbolic approximation of the original data, and~\cite{berlin2012detecting} achieves scalability by searching over a piecewise linear approximation of the data. 
While it is known that such methods \emph{can} produce high precision results in the univariate case, with carefully chosen parameters, on relatively smooth data, it is less clear how well they work in the more general case. 
In contrast to these approaches, our multidimensional matrix profile algorithm is \emph{exact}; thus, we can ignore such considerations.

To summarize, all current multidimensional motif discovery algorithms in the literature are slow, approximate, and brittle to irrelevant dimensions.
In contrast, we desire an algorithm that is fast, exact, and robust to hundreds of irrelevant dimensions.

\subsection{Dismissing Apparent Solutions}
Before continuing, we will take the time to dismiss some \emph{apparent} solutions to our problem. 

It may appear that we could use the correlation (or some other measure of mutual dependence) between the times series to guide our search for subsets of dimensions likely to yield $k$-dimensional motifs (Definition~\ref{ch2defkmotif}). 
However, this is not the case. 
Recall $\{T1, T2\}$ from Figure~\ref{ch3figtoy}. 
Their correlation is effectively zero (-0.0052). 
However, if we create 10 random walks of the same length, then on average, we expect that about 22 of the 45 pairwise combinations will have a higher correlation. 
We are interested in repeated \emph{local} patterns; statistics about \emph{global} tendencies are unlikely to be informative.

\section{The Motif Discovery Framework}
\label{ch3seframework}
The matrix profile based motif discovery framework is designed for the multidimensional matrix profile (see Section~\ref{ch2secmstamp}). 
Similar to the original matrix profile~\cite{yeh2016icdm, yeh2017dmkd}, the multidimensional matrix profile may be computed through multiple algorithms and can be adopted in various time series data miming tasks with appropriate modification and/or postprocessing.
The specific algorithm, modification, and postprocessing described in this section is just one realization for using multidimensional matrix profile in motif discovery.
As the guided (motif) search can be trivially achieved by the multidimensional matrix profile, we only introduced \emph{constrained search} and \emph{unconstrained search} in this section.

\subsection{Constrained Search}
There are two types of constraints that are useful in multidimensional motif searches: \emph{exclusion} and \emph{inclusion}. 
The exclusion constraint ``blacklists" a predetermined set of dimensions from the search; therefore, no motif can span the excluded dimensions. 
Conversely, the inclusion constraint ``whitelists" a predetermined set of dimensions, and all motifs must span the included dimensions. 
The implementation of exclusion is simple; we simply remove the blacklisted dimension before calling $m$STAMP (Algorithm~\ref{ch2algmstamp}). 
The implementation of inclusion is slightly more complicated, as we must move the distance computed by using whitelisted dimensions up to the front after a column wise-ascending sort has been applied (see line 10 in Algorithm~\ref{ch2algmstamp}).

These constraints are similar to the ``\emph{must-link}" and ``\emph{cannot-link}" operators in constrained clustering~\cite{wagstaff2001constrained}. 
They allow the user to give domain specific ``hints" to the algorithm. 
We developed this tool in collaboration with Dr. John Criley (UCLA School of Medicine), who gave us the following example. 
The reader may not understand the intricacies of the following examples, but our main point is that domain experts \emph{will} appreciate the ability to do constrained search.

Dr. John Criley noted that a cardiologist searching a heavily telemetered archive of sleep studies for evidence of predictors of \emph{Pulsus Paradoxus} might need to insist on the inclusion \emph{RESPIRATION}, but be agnostic as to which other time series could be a part of a motif~\cite{guntheroth1967effect}. 
In contrast, a neurosurgeon searching the same dataset may wish to exclude explicitly \emph{one} of the two \emph{ELECTROOCULOGRAM} (EOG) time series (eye movement). 
Because the two eyes typically move in tandem, they are redundant, and the pairing of $\{EOG_{left}, EOG_{right}\}$ will tend to report a strong, but spurious 2-dimenional motif.

We envision that domain experts in other areas will be interested in experimenting with similar domain-based constraints, based on their experience and knowledge.

\subsection{Unconstrained Search}
\label{ch3secunconstrained}
It is possible that a user knows, even if only approximately, the ``expected" dimensionality of patterns in her domain. 
For example, suppose the user wishes to find repeated saxophone elements in a musical performance that is represented in twelve-dimensional Mel-frequency cepstral coefficient (MFCCs) space. 
The user can be sure that the motif will span about three dimensions, but \emph{which} three depends on whether the instrument is a soprano, alto, tenor, baritone, or bass saxophone~\cite{loughran2008use}. 

\newpage
However, it is also possible that a user exploring a dataset has little idea about the plausible dimensionality of the repeated structure in their time series; therefore, it is necessary to support an \emph{unconstrained} search for multidimensional motif search. 

To be clear, by \emph{unconstrained} search, we mean that multidimensional matrix profile (Algorithm~\ref{ch2algmstamp}) searches the full $d$ dimension space and returns the multidimensional motif on $k$ dimensions, with $1 \leq k \leq d$, and typically $k \ll d$; where $k$ is not a user input, but it is chosen by an algorithm as the ``natural'' dimensionality of a repeated structure in the data. 
Because the multidimensional matrix profile algorithm searches for motifs in all possible subsets of dimensions of a given multidimensional time series, the problem of an \emph{unconstrained} search reduces to selecting the best motif of all possible $k$-dimensional motifs. 

Before describing our selection method for choosing the ``natural'' motif dimensionality in a dataset, we note that since all $k$ multidimensional motifs are found by the time the selection method is invoked by the user. 
If the user is not satisfied by the output of the selection method, finding it to be too conservative, or too liberal, the user can ``nudge'' the solution to examine the other possibilities without any significant (re)computational effort.

Our selection method is inspired by the elbow (or knee) finding method~\cite{thorndike1953belongs}, which is commonly used for model selection, for example choosing between alterative clusterings. 
We visually or algorithmically locate the inflection point when we plot the ``score'' for each $k$-dimensional motif. 
By adopting an elbow-finding framework, we further reduce the problem to which statistics about the motifs can be used as the score. 
We claim that the matrix profile value for each $k$-dimensional motif is a convenient and suitable score for this purpose. 

Let us revisit the toy example shown in Figure~\ref{ch3figtoy}, with the number of random walk time series set to four in addition to the two random walks that have an embedded motif. 
We note in passing that even this simple and small example is not trivial for humans to process. 
In~\cite{mmpwebsite}, we remove the color clue that helps in Figure~\ref{ch3figtoy} and shuffled the order of the time series. 
We invite the reader to see how difficult it is to find the correct answer by visual inspection.

\begin{figure}[htb]
\centering
\includegraphics[trim={12.5cm 7.5cm 12.5cm 7.5cm}, clip, width=0.8\textwidth,page=3]{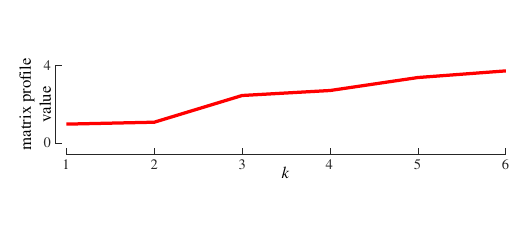}
\caption{
The matrix profile value for each $k$-dimensional motif. 
Notice how the value dramatically increases when $k$ is greater than 2 (the natural dimensionality of the embedded motif).
}
\label{ch3figelbow}
\end{figure}

We locate all $k$-dimensional motifs by using multidimensional matrix profile and plot their corresponding matrix profile values in Figure~\ref{ch3figelbow}. 
The matrix profile value for 3-dimensional motif is noticeably greater than the 2-dimensional motif's matrix profile value; therefore, the figure has suggested that the natural dimensionality is 2, coinciding with the ground truth dimensionality of the embedded motif.

Beyond the visual inspection used above, there are multiple suggestions in the literature on how to automatically locate the turning point in an elbow plot~\cite{salvador2004determining}. 
We use the Minimum Description Length (MDL) principle~\cite{rissanen1978modeling} to determine the most preferable $k$. 
In essence, the MDL principle states that the model, that allows the observed data to be compressed the most, is likely to be the true model. 
In other words, the MDL principle has cast the elbow-finding problem into a maximum compression (or minimum model size) finding problem.

The compression (encoding) technique we consider is similar to the difference-encoding scheme used in~\cite{yeh2016icdm2}. 
We encode a given time series $T$ by storing the difference between $T$ and the reference time series $T_r$. 
For example, given two discrete time series $T$ and $T_r$ (with 4-bit integers):
\begin{center}
\begin{tabular}{p{0.5cm}rrrrrrrrrr}
$T=$ & ~1 & ~2 & ~0 & 12 & ~4 & ~5 & ~2 & ~1 & 10 & 15
\end{tabular}
\end{center}
\begin{center}
\begin{tabular}{p{0.5cm}rrrrrrrrrr}
$T=$ & ~1 & ~2 & ~0 & 11 & ~4 & ~5 & ~1 & ~0 & 10 & 15
\end{tabular}
\end{center}
This would take 80 bits to store, as there are 20 4-bit integers. 
We can compute the difference $\Delta=T-T_r$:
\begin{center}
\begin{tabular}{p{0.5cm}rrrrrrrrrr}
$\Delta=$ & ~0 & ~0 & ~0 & ~1 & ~0 & ~0 & ~1 & ~1 & ~0 & ~0
\end{tabular}
\end{center}
Since $\Delta$ only contains 0s and 1s, we can use 10 1-bit integers to store $\Delta$, and compression can be achieved by storing the same information indirectly with $T_r$ and $\Delta$ (which requires 50 bits to store) instead of storing $T$ and $T_r$ directly.

\begin{figure}[htb]
\centering
\includegraphics[trim={12.5cm 7.5cm 12.5cm 7.5cm}, clip, width=0.8\textwidth,page=4]{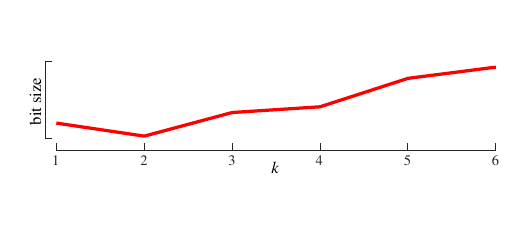}
\caption{
The required bit value for storing each $k$-dimensional motif. 
Notice the 2-dimensional motif required the minimal bit to store.
}
\label{ch3figmdl}
\end{figure}

The MDL principle can be applied trivially in this problem. 
We compute the number of bits required to store each of the $k$-dimensional motifs by compressing the subsequence pair that spans the motif subspace suggested by the $k$-dimensional matrix profile. 
Figure~\ref{ch3figmdl} shows the bit information of the same $k$-dimensional motifs (the motifs used to plot Figure~\ref{ch3figelbow}), and the embedded motif (i.e., 2-dimensional motif) can be identified by looking for the minimum point in the bit information curve.

In the case where multiple semantically meaningful $k$-dimensional motifs are presented in the multidimensional time series (e.g., Figure~\ref{ch2figexpress}), we can just interactively apply the MDL-based method to discover the motif. 
There are two steps in each iteration: 1) apply the MDL-based method to find the $k$-dimensional motif with the minimum bit size and 2) remove the found $k$-dimensional motif by replacing the matrix profile values of the found motif (and its trivial match) to infinity. 
If we apply the two steps above to the time series shown in Figure~\ref{ch2figexpress}, the 3-dimensional motif would be discovered in the first iteration, and the 2-dimensional motif would be discovered in the second iteration. 
In terms of the terminal condition for the iterative method, it can be either be an input for the user or a more advanced technique could be applied. 
Due to space limitations, we will have to leave the discussion on termination condition to future works. 
An example of applying such iterative algorithm on real-world physical activity monitoring time series is shown in Section~\ref{ch3secphysical}.

\section{Experimental Evaluation}
\label{ch3secexp}
We begin by stating our experimental philosophy. 
We have designed all experiments in a manner such that they are easily reproducible; we have built a web page~\cite{mmpwebsite} that contains all datasets and code used in this chapter.

\subsection{Synthetic Dataset}
\begin{figure}[htb]
\centering
\includegraphics[trim={12.5cm 7.5cm 12.5cm 7.5cm}, clip, width=0.8\textwidth,page=5]{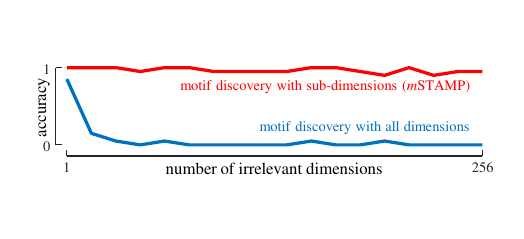}
\caption{
The accuracy of the MDL-based unconstrained motif search algorithm as we vary the number of irrelevant dimensions while keeping the number of relevant dimension (i.e., 4) fixed. 
The results are averaged over forty trials. 
The method is robust against irrelevant dimensions.
}
\label{ch3figsyn}
\end{figure}

In this section, we use a synthetic dataset to test the \emph{accuracy} of the unconstrained motif search. 
Note that the $m$STAMP algorithm does compute the multidimensional matrix profile exactly. 
However, the unconstrained motif search could still fail to find the semantically correct motifs. 
For example, this could happen if the motifs are subtle and the large number of irrelevant dimensions happens to produce a spuriously similar pair of subsequences.  
Thus, here we test the MDL-based heuristic's ability to find an embedded 4-dimensional motif among a set of multidimensional random walks. 
Figure~\ref{ch3figsyn} shows the average accuracy as we increase the number of irrelevant dimensions for both the MDL-based method and the original matrix profile by using all dimensions. 
Note the latter is an upper bound for the performance of all known rival methods~\cite{tanaka2005discovery} that use \emph{all} dimensions, since they are using \emph{all} dimensions, \emph{and} are approximate.

The MDL-based algorithm almost always finds the correct embedded motif, while the \emph{all} dimensions algorithm failed in most cases. 
Even if we increase the number of irrelevant dimensions to 64 times the number of relevant dimensions, the accuracy is still near perfect. 

Becase the multidimensional matrix profile is already computed exactly for the MDL-based algorithm, a manual inspection of the matrix profile value curve (see Figure~\ref{ch3figelbow}) could also be performed as a safeguard measure.

\subsection{Motion Capture Case Study}
The creation of \emph{motion graphs} is a fundamental problem in computer animation/gaming~\cite{kovar2008motion}. 
The task is as follows: Given a large corpus of motion capture data, automatically construct a directed graph called a \emph{motion graph} that encapsulates connections among the database.  
This allows a finite repertoire of motions to be synthesized into an infinite set of plausible motions, which can be ``steerable'' to some goal, or adaptive to changing inputs~\cite{beaudoin2008motion, kovar2008motion} (This video~\cite{beaudoin2008video}, which accompanies~\cite{beaudoin2008motion}, offers a more visual and intuitive explanation of motion graphs).

We demonstrate how $m$STAMP can help the user create higher quality motion graphs by discovering subdimensional motifs, rather than being forced to consider all dimensions. 
We applied the $m$STAMP algorithm to subject 13's motion capture recording (where the subject performs various boxing moves for 40 seconds) from the CMU Motion Capture Database~\cite{cmumocap}. 
The recording consists of a multidimensional time series with 38 dimensions, each corresponding to the motion of a given joint.

First, we visually examined the video snippet corresponding to the motif pair discovered by using \emph{all} 38 dimensions. 
We found that the subject is performing an uppercut punch in one of the snippets, but the other snippet consists of blocking/dodging motion. 
In retrospect, the results shown in Figure~\ref{ch3figsyn} make this result unsurprising. 
This finding offers support for our claim that sometimes an algorithm needs to ignore a significant fraction of the dimensions to discover semantically meaningful motifs in multidimensional time series.

\begin{figure}[htb]
\centering
\includegraphics[trim={12.5cm 7.5cm 12.5cm 7.5cm}, clip, width=0.8\textwidth,page=6]{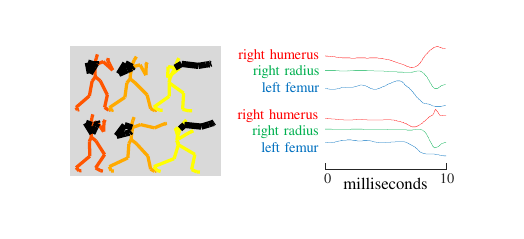}
\caption{
(see also Figure~\ref{ch3figcmu}) \emph{top}) The subject is throwing a cross. 
\emph{bottom}) The subject is throwing a one-two combo (jab cross combo). 
The right arm is highlighted with black. 
Our algorithm is capable of discovering the cross in the one-two combo, because it explores the subspace rather than \emph{all} dimensions. 
}
\label{ch3figcmu2}
\end{figure}

Next, we examine the video snippet corresponding to the 3-dimensional motif discovered by $m$STAMP. 
Here, the motif pair discovered consists of the subject performing a cross and a one-two combo. 
Our algorithm matches a simple cross with the cross in a one-two combo, and the three matching dimensions are from joints in the right humerus (right upper arm), right radius (right forearm), and left femur (left upper lag). 
The motif discovered within the subspace is much more meaningful comparing to the motif discovering using all dimension, and allows the construction of a seamless motion graph after blending all other limbs~\cite{kovar2008motion}. 
We have plotted the motions as a sequence of stick figures in Figure~\ref{ch3figcmu2}. 
Note how the right arm of the subject is in a different position in latter frames within different occurrences of the motif. 
We invite the interested reader to refer to the supporting website for the motif pairs shown in the form of video~\cite{mmpwebsite}.

\subsection{Music Processing Case Study}
\begin{figure}[htb]
\centering
\includegraphics[trim={12.5cm 7.5cm 12.5cm 7.5cm}, clip, width=0.8\textwidth,page=7]{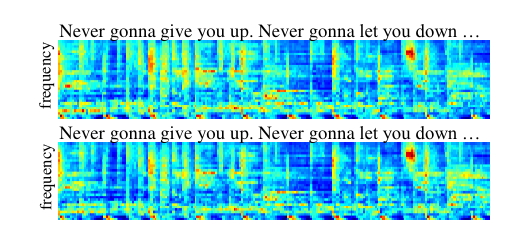}
\caption{
When the motif is found using all dimensions, the chorus is discovered.
This visualization of the data is compact and intuitive, but note that our algorithm is still operating on the raw time series signals. 
}
\label{ch3figrick1}
\end{figure}

The original matrix profile has been shown to be useful for music information retrieval (MIR)~\cite{silva2016ismir,silva2018fast}. 
To demonstrate the potential utility of our enhanced multidimensional variant of matrix profile for MIR, we have performed a simple motif discovery experiment on the Mel-spectrogram of the song \emph{Never gonna give you up}~\cite{astley2016all} by Rick Astley. 
The Mel-spectrogram is extracted with the following parameters, which are commonly used in MIR: 46 milliseconds short time Fourier transform (STFT) window, 23 milliseconds STFT hop, and 32 Mel-scale triangular filters. 
When we apply the matrix profile to music by using all dimensions with a five-second subsequence length, it is unsurprising that the motif we discovered is the chorus of the song~\cite{silva2016ismir,silva2018fast}. 
The discovered motif is shown in Figure~\ref{ch3figrick1}. Note how the extracted pairs match each other in all dimensions.

\begin{figure}[htb]
\centering
\includegraphics[trim={12.5cm 7.5cm 12.5cm 7.5cm}, clip, width=0.8\textwidth,page=8]{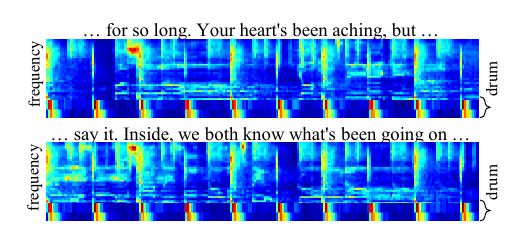}
\caption{
When the motif is found by using only the two best dimensions, the repeated drum pattern in lowest two frequency bands is dissevered. 
The lowest two frequency bands are enlarged for better visibility.
}
\label{ch3figrick2}
\end{figure}

Next, we applied $m$STAMP to discover motifs in subspaces, ranging from 1 dimension to 32 dimensions. 
We discovered that while most of the high dimensional motifs are parts of the chorus, both the 1-dimensional and 2-dimensional motif pair only represents the drum pattern. 
When we examine the exact subspace to which these lower dimensional motifs span, the motif pairs are in the space spanned by the two lowest frequency bands (i.e., typical frequency range for percussion), which confirms our intuition. 
Figure~\ref{ch3figrick2} shows the 2-dimensional motif pair. 
Note how the lowest two frequency bands are matched, while the other frequency bands differ significantly.

This example showcases one of the advantages of our method: Once the multidimensional matrix profile is computed, users can explore the matrix profile for different dimensionalities without additional computational cost. 
In other words, the users can quickly explore the motifs mined from each matrix profile and decide the correct number of dimensions for the users' specific task at hand; whether it is audio thumbnailing (as in Figure~\ref{ch3figrick1}) or generating infinite playlists (Figure~\ref{ch3figrick2})~\cite{bohra2015segmenting}.

\subsection{Electrical Load Measurement Case Study}
To illustrate the unconstrained search functionality of our motif search method, we tested our method on an electrical load measurement dataset~\cite{murray2015data}. 
The dataset consists of electrical load measurements for individual appliances (and an \emph{aggregated} load) from households in United Kingdom. 
Five appliances are considered: fridge-freezer, freezer, tumble dryer, dishwasher, and washing machine. 
The data was collected from April 19, 2014 to May 15, 2014, where the length is 17,000. 
The subsequence length was set to 4 hours.

\begin{figure}[htb]
\centering
\includegraphics[trim={12.5cm 7.5cm 12.5cm 7.5cm}, clip, width=0.8\textwidth,page=9]{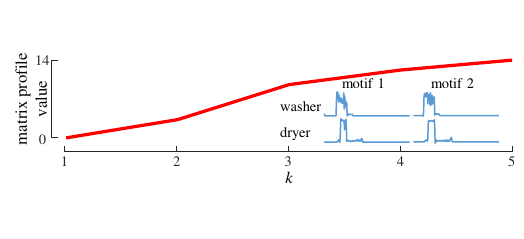}
\caption{
The natural dimensionality of the multidimensional motif is 2 as suggested by this figure. 
The discovered motifs (inset) correspond to the electrical load from using a washer, followed by dryer.
}
\label{ch3figelec}
\end{figure}

As shown in Section~\ref{ch3secunconstrained}, we can determine the natural dimensionality of a given multidimensional time series' motif by examining the matrix profile values of the $k$-dimensional motifs. 
Figure~\ref{ch3figelec} shows the matrix profile values for the motifs found in the electrical load measurement time series. 
According to the figure, it is likely that the natural dimensionality of the multidimensionality motif is 2. 
To confirm that this suggested dimensionality is semantically meaningful, we have examined the dimensions spanned by the 2-dimensional motif. 
The relevant dimensions are the electrical load measurements of tumble dryer and washing machine. 
Since both machines are typically used one after another in a short window of time, it is not surprising that the discovered 2-dimensional motif spanned the use of these related appliances.

\subsection{Physical Activity Monitoring Case Study}
\label{ch3secphysical}
Multiple types of human motion (e.g., walking, running, and rope jumping) can occur within a single recording session of physical activity, and the problem of extracting meaningful patterns is often formulated as a motif discovery problem~\cite{balasubramanian2016discovering, minnen2007detecting, tanaka2005discovery}. 
As noted in Section~\ref{ch3secunconstrained}, the MDL-based motif discovery algorithm can be applied multiple times to the precomputed multidimensional matrix profile to iteratively discover all top-K motifs (see~\cite{mueen2009exact} for definition of top-K motif). 
To showcase the effectiveness of the MDL-based dimension selection algorithm on such tasks, we consider the first subject (i.e., subject 101) of PAMAP2 dataset~\cite{reiss2012iswc}.

The dataset consists of multidimensional time series capturing both a heart rate monitor and three inertial measurement units (IMUs). 
The three IMUs are placed on the subject's wrist, chest, and ankle; each measures the temperature, 3-d acceleration data, 3-d gyroscope, and 3-d magnetometer while the subject is performing various physical activities~\cite{reiss2012iswc}. 
The activities performed by subject 101 during the recording are: lying, sitting, standing, walking, running, cycling, Nordic walking, ascending stairs, descending stairs, vacuum cleaning, ironing, and rope jumping~\cite{reiss2012iswc}.

\begin{figure}[htb]
\centering
\includegraphics[trim={12.5cm 8cm 12.5cm 8cm}, clip, width=0.8\textwidth,page=10]{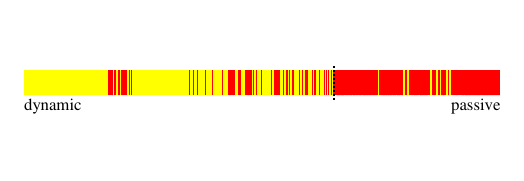}
\caption{
The MDL-based algorithm prioritizes more active and meaningful motifs. 
If we stop the retrieving process at the dashed line, the $F$-measure for the retrieval would be 0.88.
}
\label{ch3figactivity}
\end{figure}

Within the list of activities, the first three activities (i.e., lying, sitting, and standing) are more about the subject's passive posture rather than his or her action. 
As there are little or no repeated motion when the subject is not moving, the motif pairs that exist within these temporal regions should be less similar (and less meaningful) compared to the motif pairs occur during more dynamic activities. 
In other words, if our MDL-based method retrieves the motifs based on the similarity (i.e., from high similarity to low similarity), then we would expect the motifs from more dynamic events to rank higher than the more passive events. 
Figure~\ref{ch3figactivity} shows the extracted motif pairs' class (i.e., dynamic versus passive) ordered based on the order in which they were retrieved, and the result largely coincides with our speculation.

\newpage
To give a more quantitative evaluation on the motif retrieval result, we have computed the $F$-measure for each iteration (the MDL-based algorithm retrieves one item per iteration). 
The optimal stopping iteration is marked with dashed line in Figure~\ref{ch3figactivity}, and the corresponding $F$-measure is 0.88. 
Although the result $F$-measure is impressive given such simple MDL-based method, we cannot know the optimal stopping iteration without consulting the ground truth label. 
The $F$-measure provided here is for gauging the potential of the matrix profile based motif discovery framework.

\section{Discussion and Conclusion}
\label{ch3secconclude}
We have shown that if the time series motif discovery is blindly applied to the multidimensional case, the results are likely to be unsatisfactory. 
To address this, we have introduced a matrix profile based multidimensional motif discovery frame work, that solves this problem by returning the motifs that exist in natural subspaces of the higher dimensional data. 
The returned motifs are actionable, and they suggest at non-obvious latent structures in the data. 
We built our system on top of the recently introduced Matrix Profile and inherit all of its desirable properties, including anytime and incremental compatibility, low memory footprint, and scalability to large datasets~\cite{yeh2017dmkd, zhu2016icdm, zhu2018exploiting}.

\chapter{Matrix Profile for Weakly Labeled Time Series Classification}
\label{ch4}
\begin{figure}[htb]
\centering
\includegraphics[trim={12.5cm 7.5cm 12.5cm 7.5cm}, clip, width=0.8\textwidth,page=1]{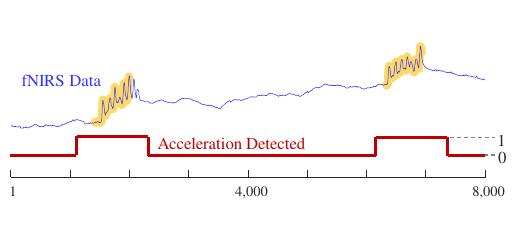}
\caption{
A two-dimensional time series. 
(\emph{top}) A real-valued fNIRS time series from a patient. 
(\emph{bottom}) A Boolean time series representing the detection of movement by the patient.  
}
\label{ch4figtoy}
\end{figure}

Much of the considerable progress in time series classification in recent years has ignored many of the pragmatic issues facing practitioners. 
To make progress, the community has typically manually contrived data to fit into the ``flat file'' format used in the machine learning community (i.e. ARFF format)~\cite{eibe1999weka}. 
The ready availability of such resources, including the UCR Time Series Archive~\cite{ucrarchive} and the more general UCI Archive~\cite{dua2017mlrepository}, has been a boon to researchers; however, it has isolated the academic community from the intricacies of time series classification as it presents itself in many industrial settings. 
To help the reader appreciate how the task-at-hand typically manifests itself in many industrial and medical settings, consider the two-dimensional time series shown in Figure~\ref{ch4figtoy}. 
We will define this ``learning from weakly labeled data'' problem more formally in Section~\ref{ch4secdef}.

\newpage
One dimension is a real-valued, functional Near-Infrared Spectroscopy (fNIRS) time series, and the other is a Boolean time series, which can be viewed as an ``annotation'' to the former. 
In a more general context, a `1' in this time series may represent a rare desirable or undesirable state. 
Here, it represents an undesirable patient movement that introduces artifacts into the recordings~\cite{sweeney2013use}. 

The weakly labeled time series learning task-at-hand reduces to the following: 
\emph{Suppose that we are given such training data ahead of time, but in the future, the Boolean time series will become unavailable (perhaps for some technical or privacy issue). 
Can we reconstruct the Boolean time series given just the real-valued signal?}

In some domains, this task can be trivial. For example, suppose the real-valued time series is \texttt{Patient Temperature (PT)}, and the Boolean time series is \texttt{HasFever (HF)}. 
Then a simple threshold rule, {\color{purple}If $PT > 100.4^\circ F$ then $HF\gets TRUE$}, would work, and we could robustly learn this rule even from a small dataset. 

\begin{figure}[htb]
\centering
\includegraphics[trim={12.5cm 7.5cm 12.5cm 7.5cm}, clip, width=0.8\textwidth,page=2]{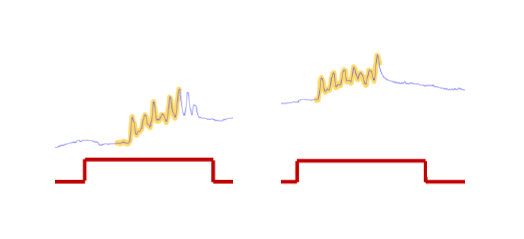}
\caption{
A zoom-in of where Figure~\ref{ch4figtoy} indicated the positive class for the Boolean time series ({\color{red}red}). 
The approximately repeated shapes in the fNIRS time series (highlighted in {\color{yellow}yellow}) are suggestive of a mechanism to solve the task-at-hand.
}
\label{ch4figtoy2}
\end{figure}

\newpage
However, note that no such threshold-based rule would work for the example in Figure~\ref{ch4figtoy}, where the height of the real-valued time series is unrelated to the Boolean value. 
Nevertheless, this toy problem does seem solvable based on alterative features. 
For example, the local variance of the time series seems to be higher at the relevant locations. 
However, in many datasets the variance, and/or other \emph{statistical} features are also a poor indicator of the Boolean variable. 
In this study, we proposed to use \emph{shape} features. 
As the zoom-in of the relevant sections demonstrates, shown in Figure~\ref{ch4figtoy2}, the local shape features may offer clues to the Boolean class labels.

Similar problems where \emph{both} time series are Boolean (or categorical) have been addressed in the literature~\cite{fung1994k}; however, the real-valued/Boolean task-at-hand here is significantly more difficult for the following reasons~\cite{sweeney2012methodology}:

\begin{itemize}
\item \textbf{Noisy Labels}: 
The Boolean annotation may be noisy. 
That is to say, it may have some false positives and/or false negatives. 
In our running example, an electrical spike in the recording device may give use an \texttt{Acceleration-Detected = TRUE} even if there was no actual movement by the patient.
\item \textbf{Label Slop}: 
As hinted at in Figure~\ref{ch4figtoy2}, the Boolean labels may only be approximately aligned with the real-value patterns. 
This problem is common in manufacturing. 
It may be that the Boolean time series is some measure of quality (acceptable/unacceptable) that can only be measured after some time lag, for example by a once-a-shift stoichiometry test~\cite{jain2001book}. 
Therefore, a failed test can only be loosely associated with the entire last eight-hour period. 
\item \textbf{Class Skew}: 
In our running example, the \texttt{Acceleration-Detected} variable was \texttt{TRUE} about a quarter of the time. 
However, more generally, the minority Boolean class may be vanishingly rare. 
Again, this is typically true by definition. In medicine and industry, we often want to learn to detect events that we hope are vanishingly rare, such as epileptic fits or catastrophic overpressurization~\cite{large2004book}. 
Thus, we expect that for a huge fraction of the time, a classifier will report ``\texttt{class unknown}''.
\item \textbf{Scale}: 
We would like to (indeed, because of class skew and the rarity of targeted events, \emph{need to}), be able to learn from very large datasets, with at least tens of millions of data points.
\item \textbf{Multi-Scale Polymorphic Patterns}: 
Most research assumes that time series patterns are of fixed length~\cite{bagnall2015time}. 
However, there is no reason to expect this to be true in real world applications. 
For example, suppose the Boolean label \texttt{TRUE} denotes an unacceptable yield quality in a chemical process. 
This might have been caused by a \texttt{flow-rate} that is increasing too quickly, is oscillating as it increases, or is increasing in discrete steps due to a sticky valve etc.~\cite{sanders2015chemical}.  
Not only do these single class root-cause patterns look different (they are \emph{polymorphic}), they can be of very different lengths. 
\end{itemize}

As the reader, will now appreciate, the data in the UCR and UCI archives are a poor proxy for learning from weakly labeled data on all the points above. 
While existing research on time classification tells us much about appropriate distance measures~\cite{bagnall2015time}, the importance of data normalization etc., to the best of our knowledge, there is currently no system that tackle the challenges above.

We note that beyond high classification accuracy, our solution to this problem also has a very desirable side-effect. 
The classification dictionaries we learn can (at least in principle) sometimes tell us something unexpected about the data/domain. 
For example, that the Asian citrus psyllid insect has two modes of eating (Section~\ref{ch4seccase1}) and humans react more strongly to images of faces than to images of houses (Section~\ref{ch4seccase2}). 
We suspect this secondary use of our algorithm may actually be more important in many domains.

The rest of this chapter is organized as follows. 
In Section~\ref{ch4secrelated} we discuss related work. 
Section~\ref{ch4secdef} introduces the necessary definitions and notations. 
We introduce our algorithm, SDTS (Scalable Dictionary learning for Time Series) in Section~\ref{ch4secsdts} and provide a rigorous empirical evaluation in Section~\ref{ch4secexp}. 
Finally, in Section~\ref{ch4secconclusion}, we discuss limitations of our work, and offer directions for future work.

The proposed matrix profile based method may also be applied, with modification, to conventional time series classification problems (e.g., UCR archive~\cite{ucrarchive}). 
Please see Appendix~\ref{appendixa} for examples on applying matrix profile based method in conventional time series classification.
We enclosed it in the appendix to enhance the understandability of this chapter.

\section{Background and Related Work}
\label{ch4secrelated}
The general literature on time series classification is vast; we refer the reader to~\cite{bagnall2015time, wang2013experimental} and the references therein. 
In the last decade, the majority of such research efforts consider \emph{only} data from the UCR archive~\cite{ucrarchive}. 
While this diverse set of datasets has been a useful resource to compare distance measures~\cite{wang2013experimental} and classification algorithms~\cite{bagnall2015time}, it tends to mask the practical issues of real-world deployments. 
The format of the UCR Archive is the antitheses of our assumptions, which are enumerated in the last section. 
In all eighty-five datasets, the ground-truth labels are all correct, there is no label slop, the classes are highly balanced, and the sizes are relatively small (i.e., the training sets have an average of just 454 exemplars). 
It is unclear if the datasets are polymorphic\footnote{
One of the current authors created or edited about one third of the datasets in the archive, and thus has some insights into this question. 
There are a handful of datasets that \emph{are} polymorphic. 
For example, for \texttt{Gun-Point}, both classes are performed by two actors of very different heights and holstering styles.
However, we believe that at least 90\% of the datasets are \emph{not} polymorphic.}, but each dataset only has patterns of a single fixed length.

The most limiting assumption of the literature is that the universe consists of $K$ well defined classes, and everything belongs to one such class. 
However, as our assumptions presage, we assume the universe consists of $K-1$ well defined classes, but there is an \texttt{other} class that is ill-defined and unstructured, and moreover, the vast majority of objects are belong to the \texttt{other} other. 
As a result of these mismatched assumptions, to the best of our knowledge, there is no technique in the literature~\cite{bagnall2015time, wang2013experimental} we can apply to this problem.

There are a handful of research efforts that have noted the label slop problem in a slightly different context. 
The first work that specifically addresses the problem is~\cite{stikic2011weakly}. 
They have cast the problem to the multi-instance learning framework by treating consecutive data points with uniform labels as \emph{bags}. 
\emph{Instances} are generated by first applying a sliding window within each bag, then conventional time series features are extracted within each sliding window. 
They use a multi-instance support vector machine to learn the correspondence between instances and labels. 
Recently, Guan et al.~\cite{guan2016efficient} has proposed a multi-instance learning graphical model based on Auto-Regressive Hidden Markov Model (ARHMM), which addresses the same problem. 
They improve upon~\cite{stikic2011weakly} by explicitly modeling the temporal dynamics of time series using ARHMM.

There is a large body of work on \emph{prognostics} and \emph{precursor} search~\cite{janakiraman2016discovery}, some of which have goals that are similar to ours (see also Section~\ref{ch4secfuture}). 
However, virtually all such work is highly domain specific. 
For example,Janakiraman et al.~\cite{janakiraman2016discovery} only considers a particular type of aviation evasive maneuvers, and Cheong~\cite{cheong2011extracting} only investigates a single type of earthquake. 
Likewise, there is a vast body of work devoted \emph{just} to the case when the time series comes from rotating machinery. 
The ability to inform/constrain an algorithm with first-principle models from aerodynamics, geology, or dynamics is clearly useful. 
However, it is contrary to our desire to have a parameter-free, domain-agnostic exploratory tool, that can work ``out-of-the-box".

\newpage
The core subroutine of our algorithm is \emph{subsequence similarity search}~\cite{rakthanmanon2012searching}, which we need to perform perhaps millions of times, in a (main memory) dataset that may also be millions of data points in length. 
This single fact may explain why we are the first to develop our rather straightforward algorithm. 
Until recently, the state-of-the-art for the similarity search task was the classic sliding-window similarity search, which must extract every subsequence, $z$-normalize it, then compute the distance~\cite{rakthanmanon2012searching}. 
While this can be accelerated in several ways (omitting the square root step of Euclidean distance, early abandoning etc.~\cite{rakthanmanon2012searching}), it is still $O(nm)$, with $n$ the query length and $m$ the dataset length. 
Note that it generally cannot be accelerated by caching the $z$-normalize subsequences, as this increases the memory footprint by a factor of $n$, and $n$ may be in the thousands.

The MASS algorithm recently introduced by Mueen and colleagues has reduced the time needed for subsequence similarity search to $O(n\log n)$~\cite{masspage}. 
Moreover, here the big O notation masks an at-least one order of magnitude additional difference. 
Unlike classic similarity search, the MASS algorithm has an extremely low constant factor. 
Moreover, it exploits FFT computation, which is the typically the most optimized algorithm in any software platform and is often accelerated by co-processors or other hardware optimizations. 
The practical implication of this is difficult to overstate. 
For example, in Section~\ref{ch4seccase2} we learn a model in 41 minutes, but this would have taken us at least many hours, perhaps days, if the state-of-the-art that that existed prior to MASS was used instead.

\section{Definitions and Notation}
\label{ch4secdef}
We are interested in the case which the real-valued time series $T$ is accompanied by a Boolean time series.

\begin{definition}
\normalfont
Given a time series $T$, a Boolean \emph{time series} $B\in \{0,1\}$ which annotated $T$ is a sequence of binary values $b_i\in \{0,1\}∶B= [b_1,b_2,\cdots,b_n]$ where $n$ is the length of $B$ and the length of $T$.
\end{definition}

Note that in some domains, the Boolean time series may be produced natively, for example by a quality control technician annotating the yield quality as \texttt{accept/reject}~\cite{jain2001book},  or by an attending physician annotating a patient's record as \texttt{tamponade/normal}~\cite{ozturk2014evaluation}. 
However, in other domains it may be the case that the analysist could convert a real-valued time series into a Boolean time series with a simple thresholding rule. 
In fact, as shown in Figure 4, this was how we produced the annotation for our running example.

\begin{figure}[htb]
\centering
\includegraphics[trim={12.5cm 6cm 12.5cm 6cm}, clip, width=0.8\textwidth,page=3]{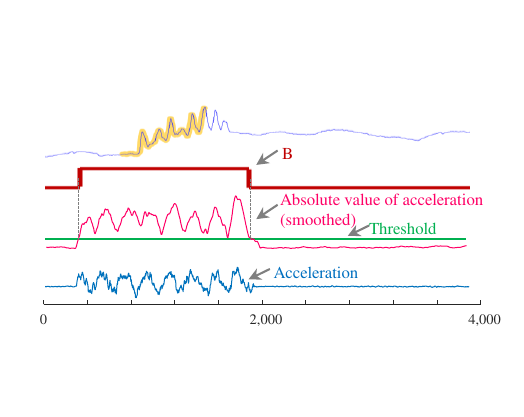}
\caption{
\emph{bottom-to-top}) We took the acceleration from a fNIRS sensor and used it to produce a new time series containing the smoothed absolute value of acceleration. 
By thresholding this new vector, we produced the Boolean vector $B$ that annotates the raw fNIRS (see Figure~\ref{ch4figtoy}).
}
\label{ch4figannotate}
\end{figure}

In many industrial domains the conversion may be even easier. 
For example, for a distillation column that is supposed to be able to produce at least 50 liters of material per second~\cite{large2004book, sanders2015chemical}, we could convert the real-valued flow rate to a Boolean measure of quality by the trivial formula: $B_{low-yield} =$ \texttt{flow-rate} $< 49$.

Note that for consistency with the literature, we refer to the \texttt{TRUE} labels as \emph{positive}, and the \texttt{FALSE} labels as \emph{negative}, without any reference to the desirability of the state. 
For example, \texttt{chemical-leak} or \texttt{EEG-seizure} may be positive. 
Here \emph{positive} just means the (typically rare) state we are attempting to predict.

\newpage
\begin{definition}
\normalfont
The \emph{weakly labeled time series problem} is the task of generating the binary time series $B'$ of a given real-valued time series $T'$ using knowledge (e.g., rules) acquired from the previously seen binary time series $B$ and real-valued time series $T$.
\end{definition}

Due to the class imbalance (and binary) nature of the problem, we use $F_\beta$-score instead of accuracy as the measure of success~\cite{powers2011evaluation}. 
We can set $\beta$ based on the relative importance of precision versus recall in the domain of interest. 
For example, $\beta$ can be set to 2 in cases where false alarms can be tolerated, while a failed alarm is more critical.

\newpage
Finally, we define the simple data structure that will allow us solve the problem-at-hand. 
We propose to solve the \emph{weakly labeled time series problem} by automatically learning a \emph{dictionary}.

\begin{definition}
\normalfont
A dictionary is a set of shapes $\mathbf{S}$ (possibly of different lengths), each with an associated threshold $H$. 
When used to monitor a new streaming time series $T'$, $B'$ is set to \texttt{TRUE} iff the current subsequence is within $h_i$ of $S_i$ ($h_i$ is $i$th member of $H$ and $S_i$ is $i$th member of  $\mathbf{S}$), else it remains \texttt{FALSE}.
\end{definition}

In the next section, we will show how we can automatically learn such dictionaries from the data. 

\section{The SDTS Algorithm}
\label{ch4secsdts}
With all the definitions and notation specified, we are finally prepared to explain our algorithm. 
Since the weakly labeled time series problem is a learning/predicting type of task, we first introduce the dictionary learning algorithm in Section~\ref{ch4seclearndic}, and subsequently show how to predict with the learnt dictionary in Section~\ref{ch4secusedic}.

\subsection{Learning the Dictionary}
\label{ch4seclearndic}
Having defined the dictionary in the previous section, and motivated the use of the $F_\beta$-score to evaluate it, how can we find the best dictionary for a given dataset? 
Even if we confine the patterns in the dictionary to come from the data itself, and limit the maximum dictionary size, say to just five entries, the number of possible dictionaries exceeds a trillion for a modestly sized dataset. 
As outlined in Algorithm~\ref{ch4alglearndic}, we propose to use an optimized greedy search to construct the dictionary. 

\begin{algorithm}[htbp]
\caption{Dictionary Learning Algorithm.}
\label{ch4alglearndic}
\begin{algorithmic}[1]
\Statex \textbf{Procedure} $Train(T, B, m)$ \vspace{-1em} 
\Statex \textbf{Input:} Time series $T$, annotation $B$, and subsequence length $m$ \vspace{-1em} 
\Statex \textbf{Output:} Dictionary (set of shape features $\mathbf{S}$ and thresholds $H$) \vspace{-1em}
\State $T' \gets ExtractPositiveSegment(T, B)$ \vspace{-1em}
\State $P \gets ComputeMP(T', m)$ \Comment{see Chapter~\ref{ch2}}\vspace{-1em} 
\State $\mathbf{C} \gets ExtractShapeCandi(T', P, m)$\vspace{-1em} 
\For{\textbf{each} $C$ \textbf{in} $\mathbf{C}$}\vspace{-1em} 
\State $H_C, F \gets FindThresholdEvalF(C, T, B)$\vspace{-1em} 
\EndFor \vspace{-1em}
\State $f \gets -\infty, \mathbf{S} \gets \emptyset, H \gets \emptyset$\vspace{-1em} 
\For{$i$ \textbf{from} $0$ \textbf{to} $|\mathbf{C}|/2$}\vspace{-1em} 
\State $f_{bsf} \gets -\infty, \mathbf{S_{bsf}} \gets \emptyset, H_{bsf} \gets \emptyset$\vspace{-1em} 
\For{\textbf{each} $(C, h)$ \textbf{in} $(\mathbf{C}, H_C)$}\vspace{-1em} 
\State $\mathbf{S_{new}}\gets \mathbf{S}\cup C$\vspace{-1em} 
\State $H_{new}\gets H\cup h$\vspace{-1em} 
\State $H_{new}, f_{new} \gets RefineThresholdEvalF(\mathbf{S_{new}}, H_{new}, T, B)$\vspace{-1em} 
\If{$f_{new} > f_{bsf}$} \vspace{-1em}
\State $f_{bsf} \gets f_{new}, \mathbf{S_{bsf}} \gets \mathbf{S_{new}}, H_{bsf} \gets H_{new}$\vspace{-1em} 
\EndIf \vspace{-1em}
\EndFor \vspace{-1em}
\If{$f_{bsf} > f$} \vspace{-1em}
\State $f \gets f_{bsf}, \mathbf{S} \gets \mathbf{S_{bsf}}, H \gets H_{bsf}$\vspace{-1em} 
\Else \vspace{-1em}
\State \textbf{break} \vspace{-1em}
\EndIf \vspace{-1em}
\EndFor \vspace{-1em}
\State \textbf{return} $\mathbf{S}, H$
\end{algorithmic}
\end{algorithm}

\begin{figure}[htb]
\centering
\includegraphics[trim={12.5cm 6cm 12.5cm 6cm}, clip, width=0.8\textwidth,page=4]{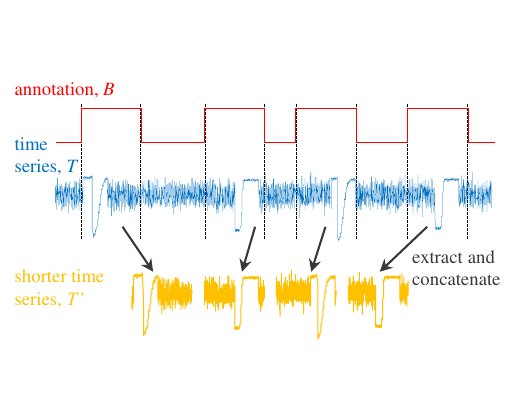}
\caption{
Positive segments are extracted and concatenated to form a shorter time series $T'$ for matrix profile computation. 
We link positive segments together in our algorithm; the space between each segment is added for visual clarity. 
Recall that `positive' just means Boolean \texttt{TRUE}, not necessary \emph{desirable}.
}
\label{ch4figconcate}
\end{figure}

In line 1, each segment that is marked positive in time series $T$ is extracted and concatenated to form another time series $T'$. 
This shorter time series $T'$ will allow us to limit the search space for shape features to place in our dictionary. 
Since the objective of the algorithm is to find a set of shape features used to predict positive segments, all possible shape feature candidates (according to $B$) are contained in $T'$. 
Our reason for concatenating all of the positive time series snippets into a single time series is more than a bookkeeping device; it allows us to extract the maximum speed-up from the STOMP algorithm (see Section~\ref{ch2secstomp}). 
Figure~\ref{ch4figconcate} shows how the shorter time series $T'$ is produced.

\begin{figure}[htb]
\centering
\includegraphics[trim={12.5cm 6cm 12.5cm 6cm}, clip, width=0.8\textwidth,page=5]{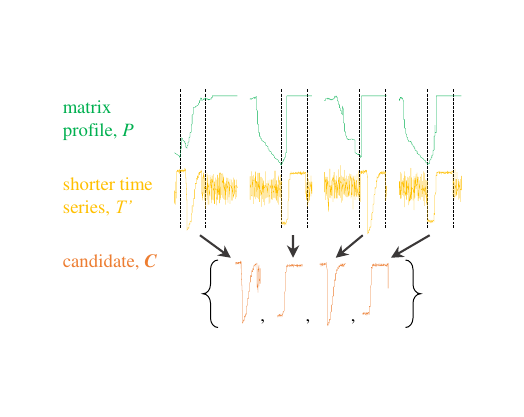}
\caption{
Candidate set $\mathbf{C}$ is selected from the shorter time series $T'$ based on the matrix profile $P$. 
The subsequences with smaller values in $P$ are selected and are added to $\mathbf{C}$.
}
\label{ch4figcandi}
\end{figure}

In line 2, the matrix profile $P$ of $T'$ is computed (recall Chapter~\ref{ch2}). 
Because $T'$ is generated by concatenating different segments of $T$, the discontinuity in time creates subsequences that do not exist in $T$ (similarly to the pseudo word `{\color{brown}cle}{\color{purple}an}' formed in the concatenation of \emph{Ora{\color{brown}cle}{\color{purple}an}omaly}). 
To avoid considering such nonexistent subsequences as a shape candidate, subsequences that cross discontinuity are ignored when computing $P$, and their corresponding values in $P$ are set to infinity. 
In line 3, a set of shape candidates $\mathbf{C}$ are selected based on their matrix profile values. 
For each positive segment in $T'$, the subseqence with lowest matrix profile value is extracted and added to $\mathbf{C}$, because subsequences with lower matrix profile values are repeated with greater fidelity than others (by definition).
Note: if there are a total of $b$ positive segments, the size of $\mathbf{C}$ is also $b$.
Figure~\ref{ch4figcandi} shows how the member of $\mathbf{C}$ is selected using $P$.

From lines 4 through 6, each shape feature $C$ in $\mathbf{C}$ is individually evaluated by finding the threshold that optimizes the $F_\beta$-score when used to perform a prediction on $T$.
Both the discovered threshold and corresponding $F_\beta$-score are stored in $H_C$ and $F$ respectively. 
The threshold is found efficiently by using the golden section search algorithm~\cite{goldenselection}. 
Although the thresholds found here are refined later in line 8 through 23, when the combination of shape features are considered, an initial set of thresholds is required as the initial condition for the coordinate ascent golden section search~\cite{goldenselection}.

In lines 8 through 23, the final shape features are selected using greedy search. 
Each shape candidate $C$ in $\mathbf{C}$ is tested by performing a prediction on $T$ when used in conjunction with previously selected candidates in $\mathbf{S}$. 
It is important that we evaluate the candidate in the context of previously added patterns; otherwise, the dictionary may fill up with redundant patterns that are only slight variants of each other.

When testing a given candidate $C$, first we refine the threshold setting for each shape feature by using the golden section search algorithm in a coordinate ascent fashion; as using multiple shape feature may require less strict thresholds. 
In the inner loop (lines 10 through 17), each candidate $C$ is tested independently with the previously selected shape feature, and the best one is stored in $\mathbf{S_{bsf}}$. 
From lines 18 through 22, if $\mathbf{S_{bsf}}$ improves the $F_\beta$-score, $\mathbf{S_{bsf}}$ is added to $\mathbf{S}$. 
Otherwise, the greedy search is terminated. 
To ensure that the candidates are tested on a sufficient amount of validation data, the number of shape features is limited to half of the number of candidates. 
Finally, the selected shape feature $\mathbf{S}$ and associated threshold $H$ are returned in line 24. 

To extend SDTS to allow dictionary elements of various lengths, we simply compute multiple matrix profiles using different settings of $m$ in line 2 and combine extracted candidate from each individual matrix profile in line 3. 
Note that while Euclidean distances of different lengths time series are not commensurate, the $F_\beta$-scores derived from different lengths time series pattern \emph{are} commensurate.

\begin{figure}[htb]
\centering
\includegraphics[trim={12.5cm 7.5cm 12.5cm 7.5cm}, clip, width=0.8\textwidth,page=6]{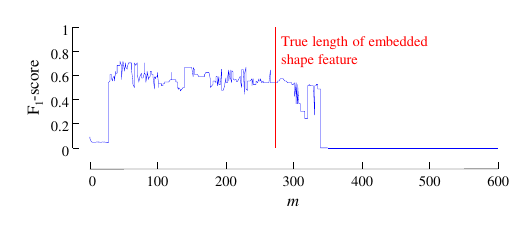}
\caption{
The performance of SDTS is relevantly insensitive to the settings of $m$. 
For an embedded pattern of length 275 (see Section~\ref{ch4secstress}), the $F_1$-score is about 0.6 for the large range of m greater than 50 and less than 300.
}
\label{ch4figsublen}
\end{figure}

Users can simply provide a set of m to SDTS, and SDTS will automatically select shape features with the appropriate lengths. 
SDTS is not particularly sensitive to the setting of $m$, as we demonstrate in Figure~\ref{ch4figsublen}. 
Given this, users can simply provide a coarse grid around the natural scale of the time series event. 
For example, if the user vaguely suspects that one hour is about the natural scale of the (sampled once a minute) data, the user can pass in a set of values for $m$ such as $[55, 60, 65]$ to bracket their intuition. 
The results of this search will almost certainly be as good as a search over increments of one second or finer.

Beyond speed-up, there is an additional reason why the coarser search may be more desirable. 
We hope that our discovered rules will be examined (and perhaps edited) by the domain experts. 
Such experts are likely to feel more comfortable dealing with rules such as ``\emph{If you see this one hour-long valley in the temp reading...}", than the spuriously precise ``\emph{If you see this fifty-nine minute, thirty-seven second-long valley…}"~\cite{johansson2004accuracy}.

\subsection{Using the Learned Dictionary}
\label{ch4secusedic}
Having learned the dictionary, applying it is straightforward; however, in Algorithm~\ref{ch4algusedic}, we outline the details of its application for completeness.

\begin{algorithm}[htb]
\caption{Prediction Algorithm.}
\label{ch4algusedic}
\begin{algorithmic}[1]
\Statex \textbf{Procedure} $Predict(T', \mathbf{S}, H)$ \vspace{-1em} 
\Statex \textbf{Input:} Time series $T'$ and dictionary (set of shape features $\mathbf{S}$ and thresholds $H$). \vspace{-1em} 
\Statex \textbf{Output:} $B'$ predicted annotation \vspace{-1em}
\State $B\gets$ vector of zeros\vspace{-1em} 
\For{\textbf{each} $(S, h)$ \textbf{in} $(\mathbf{S}, H)$}\vspace{-1em} 
\State $D \gets MASS(S, T')$ \Comment{see Section~\ref{ch2secmass}}\vspace{-1em} 
\For{$i$ \textbf{from} $0$ \textbf{to} $Length(D)-1$}\vspace{-1em} 
\If{$D[i]<h$} \vspace{-1em}
\State $B'[i]\gets 1$\vspace{-1em} 
\EndIf \vspace{-1em}
\EndFor \vspace{-1em}
\EndFor \vspace{-1em}
\State \textbf{return} $B'$
\end{algorithmic}
\end{algorithm}

In line 1, the predicted annotation $B'$ is initialized as a zero vector of the same size as the input time series $T'$. 
From line 2 to line 9, we test each shape feature in the dictionary on $T'$. 
First, we compute the $z$-normalized Euclidean distance between a shape feature and each subsequences of the same length by using the MASS algorithm (see Section~\ref{ch2secmass}). 
Next, from line 4 to line 8, we check each value in the distance vector $D$, and flag the subsequence as positive if its value is below the associated threshold $h$. 
Lastly, the predicted annotation $B'$ is returned in line 10. 
The time complexity of the prediction algorithm is $O(|\mathbf{S}| n' \log n')$ as we perform MASS algorithm $|\mathbf{S}|$ times, and each MASS call takes $O(n' \log n')$, where $n'$ is the length of $T'$. 

The extension of the prediction algorithm to streaming time series monitoring is trivial. 
In line 3, instead of computing the $z$-normalized Euclidean distance between a shape feature to all subsequence in $T'$, we simply compute the $z$-normalized Euclidean distance between the shape feature and the newly observed subsequence, and check the newly computed distance with the associated threshold.
Na\"ively, this operation takes $O(m)$ each time the algorithm ingests a new point (where $m$ is the length of shape feature). 
However, since the goal is to determine whether the resulting distance is below the threshold, techniques such as early abandoning and lower bounding~\cite{rakthanmanon2012searching} can be applied to speed up the computation. 
To concretely ground the computational demands, even if the dictionary contained one hundred shape features, each of length 1,000, it would be trivial to process a stream arriving at 500Hz, using off-the-shelf hardware.

\section{Experimental Evaluation}
\label{ch4secexp}
We begin by stating our experimental philosophy. 
We have designed all experiments in a manner such that they are easily reproducible. 
To this end, we have built a web page~\cite{weakwebsite} that contains all datasets and code used in this chapter, together with spreadsheets which contain the raw numbers.

Throughout the experiment section, we report the $F_1$-score as the single number measurement of success. 
We also report the wall clock time required, running on a desktop computer with Intel Core i7‑6700K 4 GHz Quad‑Core Processor. 
It is difficult to overstate the utility of the MASS algorithm in accelerating our learning algorithm. 
Where appropriate below, we will report the time taken if we eschew MASS, and resort to the second fastest known algorithm for Euclidean search~\cite{rakthanmanon2012searching}.  
Such times are necessarily estimated.

We begin our experiments with a synthetic dataset. 
Such tests are less compelling than the four diverse real-world case studies that follow it. 
However, the synthetic dataset allows us to ``stress test" our algorithm, by varying the factors that make the task-at-hand challenging.

\subsection{Stress Testing on a Synthetic Dataset}
\label{ch4secstress}
The TRACE dataset~\cite{roverso2000multivariate} is a synthetic dataset designed to model industrial processes that are ``\emph{..characterized by long periods of steady-state operation, intercalated by occasional shorter periods of a more dynamic nature in correspondence of either normal events, such as minor disturbances, planned interruptions or transitions to different operation states, or abnormal events, such as major disturbances, actuator failures, instrumentation failures, etc.}".
These are all data characteristics that have been echoed back to Oracle by its IOT customers in the manufacturing and the oil-and-gas industries~\cite{oracle2016driving}. 
Testing on such synthetic data offers us the possibility of studying how the properties of the data and the domain affect our ability to learn.

We begin by performing a single experiment on a particular instantiation of the problem space; then, having calibrated our expectations, we vary each factor of the problem space one-by-one, while holding everything else constant to see how much that factor matters. 
The factors in question are:

\begin{itemize}
\item \textbf{Occurrence of positive events}:
Varying this factor is similar to varying skewness between classes (and number of training example) in traditional binary classification.
\item \textbf{Fraction of false positive labels}:
Some segments without repeated shape features are marked as positive. 
Varying this factor allows us to examine the algorithm's robustness against noisy labels.
\item \textbf{Fraction of false negative labels}:
Some sections of the raw time series with conserved shape features are marked as negative. 
Similarly to the last factor, varying this factor allows us to examine the algorithm's robustness against noisy labels.
\item \textbf{Amount of label slop}:
This factor is unique to weakly labeled time series classification, and is measured by the fraction of each positive segment being irrelevant time series (i.e., time series other than embedded shape feature). 
Varying this factor allows us to examine the algorithm's ability to work against imprecise labels in time.
\end{itemize}

The default setup is as follows: 100 occurrences of positive events, 0 false positives, 0 false negative, and 0.7 label slop.

\begin{figure}[htb]
\centering
\includegraphics[trim={12.5cm 7.5cm 12.5cm 7.5cm}, clip, width=0.8\textwidth,page=7]{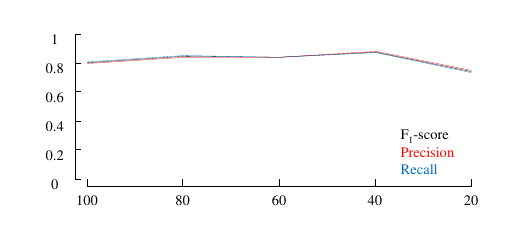}
\caption{
The performance of SDTS versus various settings of positive events occurrence.
SDTS's performance suffers slightly when the number of positive events decreases.
}
\label{ch4fignpos}
\end{figure}

\begin{figure}[htb]
\centering
\includegraphics[trim={12.5cm 7.5cm 12.5cm 7.5cm}, clip, width=0.8\textwidth,page=8]{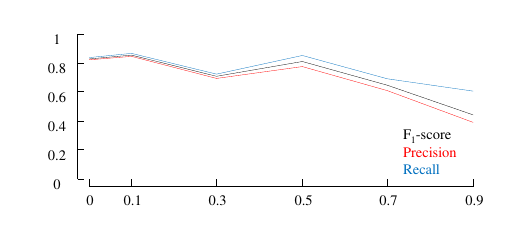}
\caption{
The performance of SDTS versus various settings of false positive fraction. 
Unsurprisingly, the performance decreases as the false positive fraction increases, but the degradation is slow and graceful.
}
\label{ch4figfalsepos}
\end{figure}

\begin{figure}[htb]
\centering
\includegraphics[trim={12.5cm 7.5cm 12.5cm 7.5cm}, clip, width=0.8\textwidth,page=9]{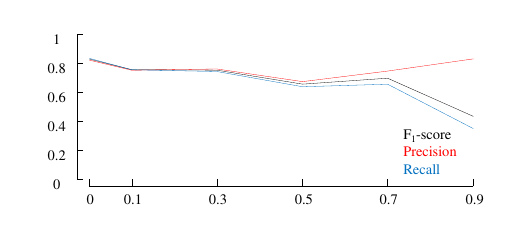}
\caption{
The performance of SDTS versus various settings of false negative fraction. 
Interestingly, the precision increases when the false negative fraction increases.
}
\label{ch4figfalseneg}
\end{figure}

\begin{figure}[htb]
\centering
\includegraphics[trim={12.5cm 7.5cm 12.5cm 7.5cm}, clip, width=0.8\textwidth,page=10]{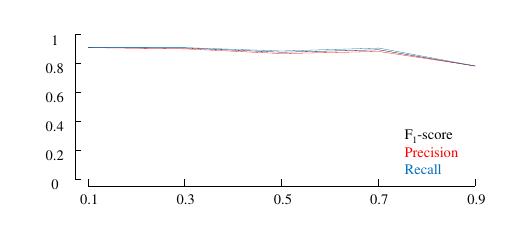}
\caption{
The performance of SDTS versus various settings of label slop amount. 
SDTS is not sensitive to increasing amounts of label slop.
}
\label{ch4figslop}
\end{figure}

We have summarized the $F_1$-score, precision, and recall versus various settings in each factor in Figure~\ref{ch4fignpos}, Figure~\ref{ch4figfalsepos}, Figure~\ref{ch4figfalseneg}, and Figure~\ref{ch4figslop}. 
Note that in each plot, only a single factor is varied while all the other factors are kept fixed. 
The synthetic data was generated by embedding TRACE patterns to random walk. 
Each set of experiments was repeated 16 times (with random walks generated by different seed), and the reported performances averaged 16 trials. 
Since the random walk for each set of experiments was generated independently, the performance of the default setups in each figure is slightly different, but within each plot, the numbers are commensurate as we vary the factors.

As shown in Figure~\ref{ch4fignpos}, increasing the number of positive events benefits SDTS, since the number of shape feature candidates is directly proportional to the occurrence of positive events, and SDTS benefits from larger set of candidates to search over.

Moreover, increasing the number of positive events can mitigate the issues associated with class imbalance. 
Since the length of training data is fixed, increasing the ratio of positive events reduces the preponderance of negative events; thus nudging the positive to negative ratio is closer to 1.

In contrast, SDTS's $F_1$-score, precision, and recall all suffer from the increase of false positives (i.e. mislabeled data) in the training data as demonstrated in Figure~\ref{ch4figfalsepos}.

It is unsurprising that the performance of the system degrades with increasing false positive labels. 
However, the performance of SDTS offers graceful degradation and does not fall dramatically, even when the fraction of false positive is as high as 0.5.

As shown in Figure~\ref{ch4figfalseneg}, SDTS's $F_1$-score and recall suffer from the increase of false negatives. 
Yet, the precision is maintained at a relatively high value compared to the other two performance metrics.

One possible explanation is that the false negatives force the dictionary learning algorithm to learn a tighter threshold, because the algorithm is trying to separate a captured (true) shape feature from an embedded shape feature (which is very similar to the captured shape feature) in negative segment. 
Similar to Figure~\ref{ch4figfalsepos}, the $F_1$-score of SDTS does not drastically decrease until the fraction of false negative is 0.7.

The experiment shown in Figure~\ref{ch4figslop} suggests that SDTS's performance is only slightly impaired by large amounts of label slop. 
One possible reason is that SDTS is shift invariant. 
In other words, as long as the embedded shape feature is within the positive segment, SDTS would find the shape feature even if the ratio between noise and signal (the shape feature) is as large as 0.9.

\subsection{Insect EPG Case Study}
\label{ch4seccase1}
\begin{figure}[htb]
\centering
\includegraphics[trim={12cm 5cm 12cm 5cm}, clip, width=0.7\textwidth,page=11]{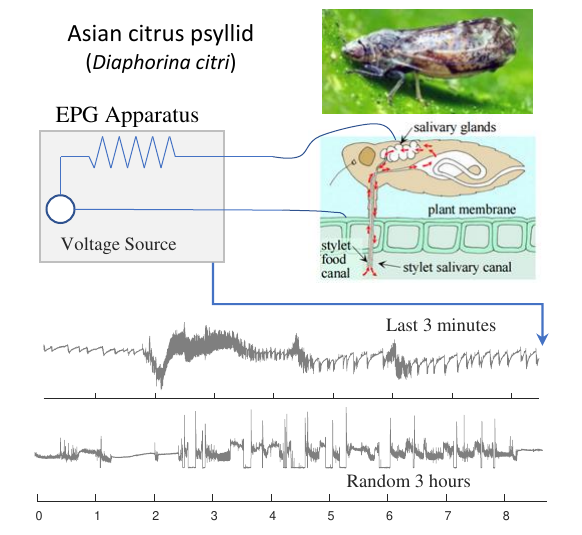}
\caption{
\emph{top-to-bottom}) The Asian citrus psyllid can be connected to an EPG (Electrical Penetration Graph) apparatus, and have its behavior recorded. 
As the three minute, and three hour snippets show, this behavior is suggestive of structure, but nosily and complex.
}
\label{ch4figinsect}
\end{figure}

Insects that feed by ingesting plant fluids cause devastating damage to agriculture worldwide, primarily by transmitting pathogens of plants. 
As a concrete example, the Asian citrus psyllid (\emph{Diaphorina citri}) shown in Figure~\ref{ch4figinsect}.\emph{top} a is vector of the pathogen causing citrus greening disease, and has already caused billions of dollars of damage to Florida's citrus industry in the last decade and is poised to do this same in California.

As shown in Figure~\ref{ch4figinsect}, the feeding processes required for successful pathogen transmission by psyllids can be recorded by monitoring voltage changes across an insect-food source feeding circuit. 
However, as~\cite{willett2016machine} notes ``\emph{The output from such monitoring has traditionally been examined manually, a slow and onerous process}''. 
While we do not wish to makes any claims of entomological significance, it is natural to ask if our ideas can be applied to such datasets.

\begin{figure}[htb]
\centering
\includegraphics[trim={8cm 7cm 8cm 7cm}, clip, width=0.8\textwidth,page=12]{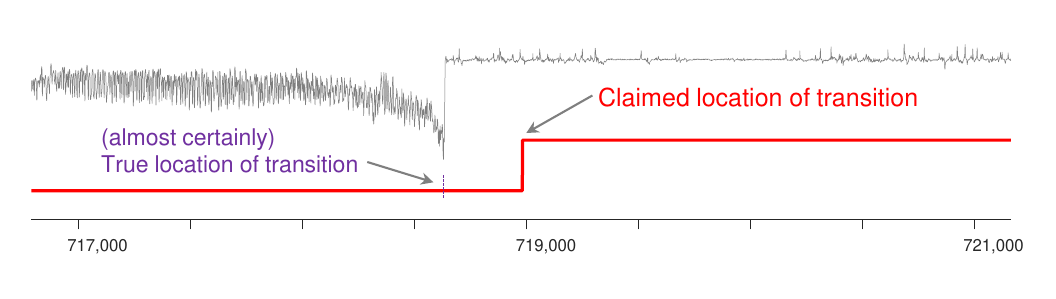}
\caption{
An original annotation of a transition from \texttt{stylet passage} to \texttt{non-probing} behavior~\cite{willett2016machine}. 
Although we do not have access to the original data, it is virtually certain that this is an example of label slop.
}
\label{ch4figinsectslop}
\end{figure}

We obtained a dataset recently made publicly available by the United States Department of Agriculture. 
While this dataset has been labelled by domain experts, as we show in Figure~\ref{ch4figinsectslop}, it contains significant label slop, and is thus an ideal dataset to test SDTS robustness to that issue. 

We learn the model from one EPG recoding section of an insect feeding on \emph{Corrizo} (a rootstock for citrus) and verify the learned model on another EPG recoding of feeding on the same citrus variation. 
While both experiments consider the same \emph{species}, the Asian citrus psyllid, and the \emph{individual} insects where different, thus we are testing the generalization ability of our algorithm.

\begin{figure}[htb]
\centering
\includegraphics[trim={12.5cm 7.5cm 12.5cm 7.5cm}, clip, width=0.8\textwidth,page=13]{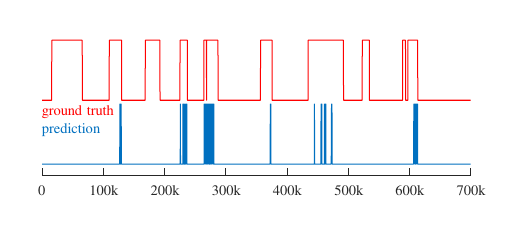}
\caption{
The annotation predicted by SDTS versus the ground truth annotation. 
The prediction of SDTS is not perfect, but it has no false positives.
}
\label{ch4figinsectpredict}
\end{figure}

To demonstrate our algorithm's ability to capture shape features from multiple classes, we treat both \emph{phloem ingestion} and \emph{xylem ingestion} as the positive class. 
SDTS is able to achieve a $F_1$-score of 0.78, a precision of 1.00, and a recall of 0.64. 
Figure~\ref{ch4figinsectpredict} shows the prediction result with the ground truth label. We can see that SDTS is capable of learning a model that gives no false positives despite some false negative.

\begin{figure}[htb]
\centering
\includegraphics[trim={12.5cm 7.5cm 12.5cm 7.5cm}, clip, width=0.8\textwidth,page=14]{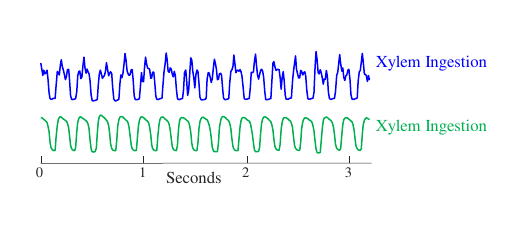}
\caption{
The first two patterns in the learned EPG model show that a single class can be highly polymorphic.    
}
\label{ch4figinsectpattern}
\end{figure}

Beyond the high accuracy achieved, we wonder if the dictionary learned is itself useful and/intuitive. 
We showed the model to Dr. Gregory Walker (Figure~\ref{ch4figinsectpattern}), who has not involved in collecting the data, but who has decades of experience in manually exploring EPG data. 
He noted ``\emph{(the first two waveforms) represent ingestion from two different apoplastic compartments such as xylem versus other extracellular space.}''~\cite{walker2017personal}, which confirms that the two patterns are indeed polymorphic variants of a single behavior.

\subsection{Neuroscience Case Study}
\label{ch4seccase2}
\begin{figure}[htb]
\centering
\includegraphics[trim={12.5cm 7.5cm 12.5cm 7.5cm}, clip, width=0.8\textwidth,page=15]{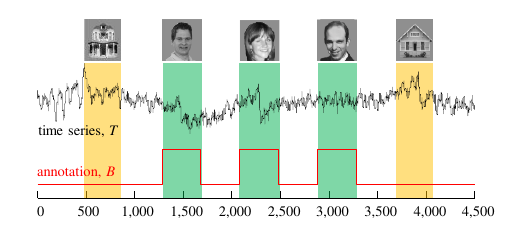}
\caption{
A small snippet of the electrocorticographic data used in our face discrimination experiment. 
The positive class is when the patient can see a \texttt{face}, and the negative class is when the patient is seeing either a \texttt{house} or \texttt{nothing}. 
}
\label{ch4figecog}
\end{figure}

The connection between visual perception of objects and neural activity in the visual cortical areas is a fundamental problem in neuroscience~\cite{miller2016spontaneous}. 
Recent work has shown that that electrical potential from the temporal lobe in humans contains sufficient information for spontaneous and near-instantaneous identification of a subject's perceptual state~\cite{miller2016spontaneous}. 
However, such efforts require an extraordinary amount of domain knowledge, data preprocessing, and algorithm tuning. 
Here, we will attempt to duplicate some fraction of the recent achievements, with our \emph{completely} domain agnostic algorithm. 
To be clear, we are not claiming any medical significance or utility in this section. 
We are merely showing that, in a real-word, noisy, complex and massive electrocorticographic (ECoG) dataset (see Figure~\ref{ch4figecog}) created out of our control, we can robustly learn models that capture true structure in the data and allow (much) better-than-random guessing predictions on unseen data.

The ECoG data we consider was collected from an epileptic patient. 
Electrodes were placed directly on the patient's occipital lobe (the visual processing center for the mammalian brain). 
Fifty images of faces and fifty images of houses were shown to the patient in a random order, with 0.4-second pauses in-between. 
While fifty 1,000 Hz traces where recorded from various parts of the brain, for simplicity we consider only a single trace.
Our task is to examine the traces to find patterns that indicate that the patient is seeing a \texttt{face}.

As noted in~\cite{miller2016spontaneous} ``\emph{face-selective} (time series patterns) \emph{may have wide structural variation, with `peaks' and `troughs' that are very different in shape, latency, and duration,}'' making this a challenging task. 
It particular, we see this uncertainty in \emph{latency} and \emph{duration} as label slop.

We performed our experiment on subject 2. 
We partitioned the original time series into three sections (each section corresponding to different experiment runs) and performed three-fold cross validation similarly to~\cite{miller2016spontaneous}. 
To confirm that SDTS performs better than the \emph{default rate} (random guessing in proportion to the prior probability of events), we repeated the experiment on the same data using a permutation test~\cite{ojala2010permutation}. 
We generated the permuted labels by randomly shuffling the temporal location of the positive segments. 
In other words, a positive segment may or may not correspond to \texttt{face} in the false label. 
The experimental results suggest that SDTS is significantly better than random guessing ($F_1$-score of 0.47 vs. 0.21). 
This is a huge difference, and it is unsurprising that a two-sample t-test confirms the difference at a 5\% level.

\newpage
The time it takes SDTS to learn a model from the ECoG dataset was 41 minutes. 
If we replace MASS with the standard Euclidean distance subsequence-search technique, a sliding window that exacts the subsequences, $z$-normalizes them, and then compares the distance, this time grows to a few days. 
Interestingly, (as also noted in~\cite{rakthanmanon2012searching}), we found the time needed to $z$-normalize the subsequences dominates the time required for this operation.  

\subsection{Traffic Loop Sensor Case Study}
\begin{figure}[htb]
\centering
\includegraphics[trim={12.5cm 7.5cm 12.5cm 7.5cm}, clip, width=0.8\textwidth,page=16]{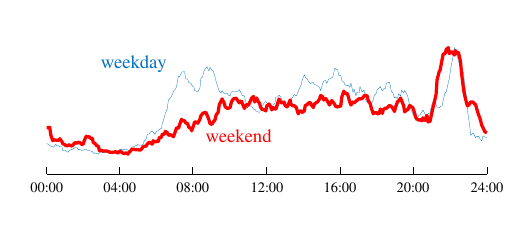}
\caption{
The major differences between weekend ({\color{red}red}/bold) and weekday ({\color{blue}blue}/fine) patterns are the morning and evening rush hour `bumps'.
}
\label{ch4figtraffic}
\end{figure}

To demonstrate that SDTS does not produce an unnecessarily complex dictionary for simple problems, we have applied SDTS on the much-studied Dodgers loop sensor dataset~\cite{freewaypems,ihler2006adaptive}. 
This dataset records the number of vehicles on the 101 North freeway off-ramp near Dodgers Stadium in Los Angeles for 25 weeks. 
The research community has performed a wide variety of time series data miming experiments on the dataset. 
The particular experiment we performed was weekend detection. 
In other words, in the accompany annotation of the training data, all the weekends were marked as positive while the weekdays were marked as negative. 
While this is a contrived problem, it is not trivial. 
As Figure~\ref{ch4figtraffic} suggests, the data is noisy. 
Moreover, there are dropouts (random occasions when the sensor was offline), and several weekday holidays that might act as pseudo weekends. 
The data exhibits ``bursts'' when the Dodgers played a home game, which could be any day of the week. 
Finally, as the data spans a half year, and we learn from only the first twelve weeks, there is the possibly of concept drift as the seasons change.

\newpage
Nevertheless, as Figure~\ref{ch4figtraffic} shows, in general, a typical weekday traffic pattern \emph{does} look different than typical weekend traffic, suggesting a simple model should suffice for accurately distinguishing the weekend from a weekday.

\begin{figure}[htb]
\centering
\includegraphics[trim={12.5cm 7.5cm 12.5cm 7.5cm}, clip, width=0.8\textwidth,page=17]{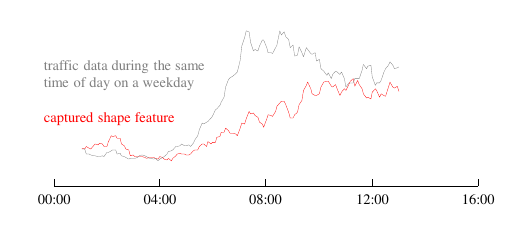}
\caption{
The capture shape feature is corresponding to weekend morning traffic.
}
\label{ch4figtrafficmodel}
\end{figure}

To perform such an experiment, we trained the SDTS model on the first half of data. 
Then, we used the learned model on the second half. 
The result model is surprisingly simple. 
The model only contains one shape feature, corresponding to traffic density in the morning (more precisely, midnight to noon) of a Saturday. 
The captured feature is relevant as it can be used to differentiate a (relatively) quiet weekend morning from a busy weekday morning. 
Figure~\ref{ch4figtrafficmodel} shows the captured shape feature and how different it is from the traffic data from a weekday.

\begin{figure}[htb]
\centering
\includegraphics[trim={12.5cm 7.5cm 12.5cm 7.5cm}, clip, width=0.8\textwidth,page=18]{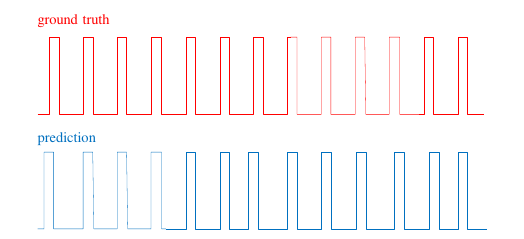}
\caption{
The ground-truth vs. prediction of the SDTS model.  
Of 13 weekends, the learned model perfectly annotates 9 of them. 
The other 4 weekends are slightly mislabeled in terms of their temporal locations (falsely skipping the Saturday or mistakenly labeling Friday as a weekend day).
}
\label{ch4figtrafficpredict}
\end{figure}

Despite the learned model being sample, it can accurately detect weekend from the traffic data. 
Figure~\ref{ch4figtrafficpredict} shows how similar the predicted annotation is to the ground truth.

\newpage
This is a good place to revisit one of our assumptions. 
Recall that we are searching for subsequences in the $z$\emph{-normalized space}. 
Here, it might be imagined that we should not normalize the data, as the \emph{absolute} values offer clues, with higher traffic volumes on weekdays. 
If this is really desired, it is trivial to achieve, as the MASS algorithm, and the matrix profile that is built upon it, can trivially be converted to an amplitude/offset sensitive algorithm by simply commenting-out some lines of code~\cite{masspage}. 
However, we claim that this is unlikely to ever be appropriate. 
Recall that in our motivating example shown in Figure~\ref{ch4figtoy}, the \emph{shape} was informative, but the change in \emph{offset} (in this domain, ``\emph{wandering baseline}'') was not. 
We argue that this is generally true, even in this apparent counterexample. 
For example, the absolute volume of cars could change due to nearby road maintenance, or even because of changes in the price of fuel; however, the overall shapes will remain near constant. 
In~\cite{rakthanmanon2012searching}, the authors make a more detailed argument that virtually every task, in almost every dataset requires the normalization of subsequences.

\subsection{Predicting the Future: A Tentative Case Study}
\label{ch4secfuture}
As the experiments in the previous sections suggest, the ability of SDTS to predict the \emph{current} state of the world can be useful in many domains. 
However, in many situations it is clearly more desirable and actionable to predict the \emph{future} state of the world. 
Such shape features are called sometimes called ``precursors'' or ``precursors signatures'' (although the literature is inconsistent in its nomenclature~\cite{cheong2011extracting, janakiraman2016discovery}).

\begin{figure}[htb]
\centering
\includegraphics[trim={11.5cm 6.5cm 11.5cm 6.5cm}, clip, width=0.8\textwidth,page=19]{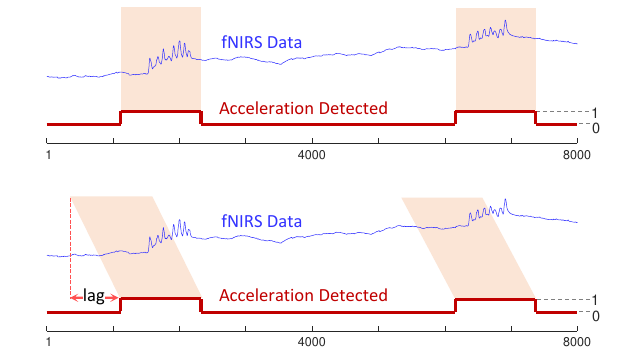}
\caption{
\emph{top}) A visual reminder of the original setup for the weakly labeled classification problem (recall Figure~\ref{ch4figtoy}). 
\emph{bottom}) Generalizing the problem to a precursor setting simply requires compensating for the lag between the binary time series $B$ and real-valued time series $T$.
}
\label{ch4figfuture}
\end{figure}

As Figure~\ref{ch4figfuture} suggests, it is trivial to generalize SDTS to allow the discovery of precursors.

All we need do to generalize from our typical consideration of ``co-cursors'' to precursors, is create a lag between the binary time series $B$ and real-valued time series $T$ (conversely, it may sometimes be more natural to speak of the \emph{lead} time between $T$ and $B$). 
As a practical matter, we can achieve this by simply removing the first $L$ data points of $B$, where $L$ is the length of the desired lag.

We may have some ideas of a reasonable value for $L$ based on the domain. 
For example, for a small distillation column a lag of five minutes might be ambitious, but for a large distillation column, the inertia of the system may allow a lag of a few hours~\cite{jain2001book}. 
As it happens, this discussion of a domain dependent constraints may be moot. 
We will always want as much lead time as possible, and our proposed algorithm is fast enough to test expanding values of $L$ until the scoring function is unable to find predictive patterns.

To test this idea, we have adapted a real dataset. 
This contrived experiment is not as interesting as the propriety real-world customer problem that inspired this study, but has the advantage that we can share all the data with the community.

\begin{figure}[htb]
\centering
\includegraphics[trim={12.5cm 7.5cm 12.5cm 7.5cm}, clip, width=0.7\textwidth,page=20]{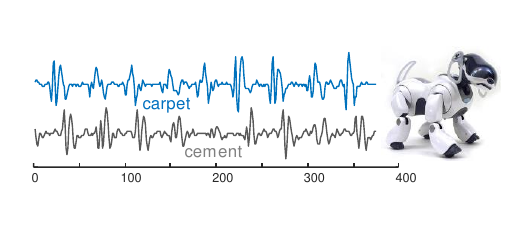}
\caption{
\emph{left}) Two three-second snippets extracted from a Sony AIBO robot dog (\emph{right}). 
The snippets show about three gait cycles.
}
\label{ch4figdog}
\end{figure}

As shown in Figure~\ref{ch4figdog}.\emph{right}, the Sony AIBO is a small quadruped robot that comes equipped with a tri-axial accelerometer and a (very) low-resolution camera. 
This accelerometer measures data at a rate of 125 Hz. 
In Figure~\ref{ch4figdog}.\emph{left}, we show two snippets of telemetry from the accelerometer's z-axis (the direction pointing skyward) as the robot walks on two different surfaces. 
As the reader will appreciate, the differences in gait due to the surface makes are non-obvious, even after careful visual inspection, and seem swamped by natural viability and noise.

The onboard camera and limited processing power do not lend themselves to complex image processing, but we can simply `snap' a targeted color to the positive class. 
Finally, if we task our dog to walk backwards across the lab, we will produce a dataset that exactly models the setup in Figure~\ref{ch4figfuture}.\emph{bottom}, with series $T$ being the accelerometer value, and $B$ being \texttt{cement=TRUE} extracted from the video feed. 
The exact amount of lag depends on the angle of the robot's head. 
Again, while we acknowledge that this toy experiment is highly contrived, it is non-trivial, and is an excellent proxy for real-word problems in prognostics for manufacturing and transport.

Our dataset was created by interleaving the $z$-axis accelerometer time series of Sony AIBO surface recognition dataset~\cite{vail2004learning}. 
The original dataset consists of accelerometer time series, corresponding to the robot walking on different surfaces (i.e., carpet, field, and cement)~\cite{vail2004learning}. 
The goal of our experiment is to show that SDTS is capable of discovering \emph{precursors} for an event of interest. 
Among the three classes provided by the Sony AIBO dataset~\cite{vail2004learning}, we picked ``walking on cement'' as the targeted event of interest.

We begin by splitting each of the time series into disjoint training and test splits. 
Then, we apply the following three steps independently to the training and test data. 

\begin{enumerate}
\item \texttt{carpet} and \texttt{field} are concatenated together to make the problem more challenging. 
\item \texttt{cement} is sliced into segments of various lengths. 
\item The segments of \texttt{cement} are embedded into the \texttt{carpet-field} time series at multiple randomly selected locations.
\end{enumerate}

\begin{figure}[htb]
\centering
\includegraphics[trim={12.5cm 7.5cm 12.5cm 7.5cm}, clip, width=0.8\textwidth,page=21]{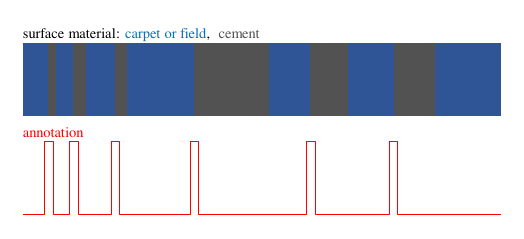}
\caption{
\emph{top}) A representation of the surface walked upon by the robot. 
\emph{bottom}) The annotation used to train SDTS for precursors of walking on ``\texttt{cement}'' has a slight lag, due to the delay between the robot experiencing the real-value stimulus, and seeing the positive label. 
}
\label{ch4figdoglabel}
\end{figure}

Figure~\ref{ch4figdoglabel}.\emph{top} illustrate how the time series from various classes are put together. 
Note that the duration of the positive events, and the amount of interstitial time between them, are random and highly variable.

In order to generate the accompanying annotation for precursor discovery training data, we flag a small chunk (about 2 seconds in time) of the annotation time series as \emph{positive} at the beginning of each \texttt{cement} regions. 
To make the situation conform to our assumptions, the positive segments for each \texttt{cement} regions has a lag relative to the actual starting point of \texttt{cement}, because our robot experiences a slight change in gait (due to walking on a different material), before visually confirming the change of surface.  
Figure~\ref{ch4figdoglabel}.\emph{bottom} shows an example of such annotation.

\begin{figure}[htb]
\centering
\includegraphics[trim={12.5cm 7.5cm 12.5cm 7.5cm}, clip, width=0.8\textwidth,page=21]{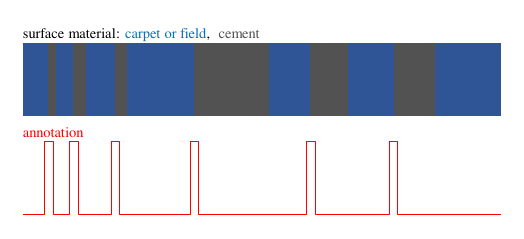}
\caption{
\emph{top}) The surface walked on by the robot the ground truth for our predictions. 
\emph{bottom}) The predictions made by our precursor model, found using SDTS.
}
\label{ch4figdogpredict}
\end{figure}

With the annotation for training data prepared in this fashion, we can simply apply the SDTS algorithm without any modification, to discover the precursor(s) for ``walking on cement''.
Figure~\ref{ch4figdogpredict} shows the predicted annotation against the ground truth. 
The corresponding $F_1$-score is 0.63. 
There are a handful of false negatives, but all regions predicted as positive are indeed just prior to the robot seeing \texttt{cement}.

In essence, this experiment shows that in principle, we can use SDTS to gain a little ``lead-time'' to predict upcoming events.

\section{Discussion and Conclusions}
\label{ch4secconclusion}
Of the four case studies we considered, we believe that only \emph{Traffic Loop Sensor} would be solvable by ``eye'', by the average person. 
The \emph{Insect EPG} dataset appears to be at least partially solvable by humans, but only fully solvable by expert entomologists with decades of experience examining such data~\cite{walker2017personal}. 
For both the \emph{Robot Gait} and \emph{Neuroscience} datasets, our algorithm offers truly superhuman performance. 
Even if we ``cheat'' by examining various sources of extra information, the differences discovered by our algorithm are too subtle for us to appreciate, much less duplicate or improve upon with human-coded rules.

In conclusion, we have introduced SDTS, a parameter-free domain agnostic algorithm for learning from weakly supervised datasets\footnote{
See Appendix~\ref{appendixa} for  information on applying SDTS algorithm to conventional time series classification problem.}. 
We have made all code and data freely available to the community, to confirm, extend, and exploit our work~\cite{weakwebsite}.

Future work includes consideration of the multidimensional time series case, and allowing humans to interactively edit the learned models. 
We are also interested in the ``cold-start'' problem~\cite{gao2015item}. 
Could a model learned on one domain be used on similar domain, at least until enough data has been observed to allow relearning the model?
In the industrial domain, this problem can arise if the production run for one object finishes, and a new production run for a similar device begins.

\chapter{Matrix Profile for Representation Learning}
\label{ch5}
Unsupervised representation learning has been shown effective in tasks such as dimension reduction, clustering, visualization, information retrieval, and semi-supervised learning~\cite{goodfellow2016book}.
Learned representations have been shown to achieve better performance on individual tasks than domain-speciﬁc handcrafted features, and different tasks can use the same learned representation~\cite{goodfellow2016book}.
For example, the embedding obtained by methods like word2vec~\cite{mikolov2013arxiv, mikolov2013nips} has been exploited in many different text mining systems~\cite{catherine2017transnets, zheng2017joint}.
Moreover, to help a user extract knowledge from a dataset, a data exploration system can first learn the representation without supervision for each item in the dataset; then display both the clustering (e.g., $k$-means~\cite{lloyd1982tit}) and visualization (e.g., $2D$ projection with $t$-Distributed Stochastic Neighbor Embedding/$t$-SNE~\cite{maaten2008jmlr}) results produced from the representation.

\newpage
There are two types of unsupervised representation learning methods: \emph{domain-specific} unsupervised representation learning methods and \emph{general} unsupervised representation learning methods.
While domain-specific unsupervised representation learning methods like word2vec~\cite{mikolov2013arxiv, mikolov2013nips} and video-based methods~\cite{agrawal2015iccv, jayaraman2015iccv, wang2015iccv, pathak2017cvpr} have been widely adopted in their respective domains, their success cannot be directly transferred to other domains because their assumptions do not hold for other types of data.
In contrast, general unsupervised representation learning methods, such as autoencoder~\cite{bengio2007nips, huang2007cvpr, vincent2008icml, vincent2010jmlr}, can be effortlessly applied to data from various domains, but the performance of general methods is usually inferior to those that utilize domain knowledge~\cite{agrawal2015iccv, jayaraman2015iccv, mikolov2013arxiv, mikolov2013nips, pathak2017cvpr, wang2015iccv}.

\begin{figure}[htb]
\centering
\includegraphics[trim={7cm 4.5cm 7cm 4.5cm}, clip, width=0.90\textwidth,page=1]{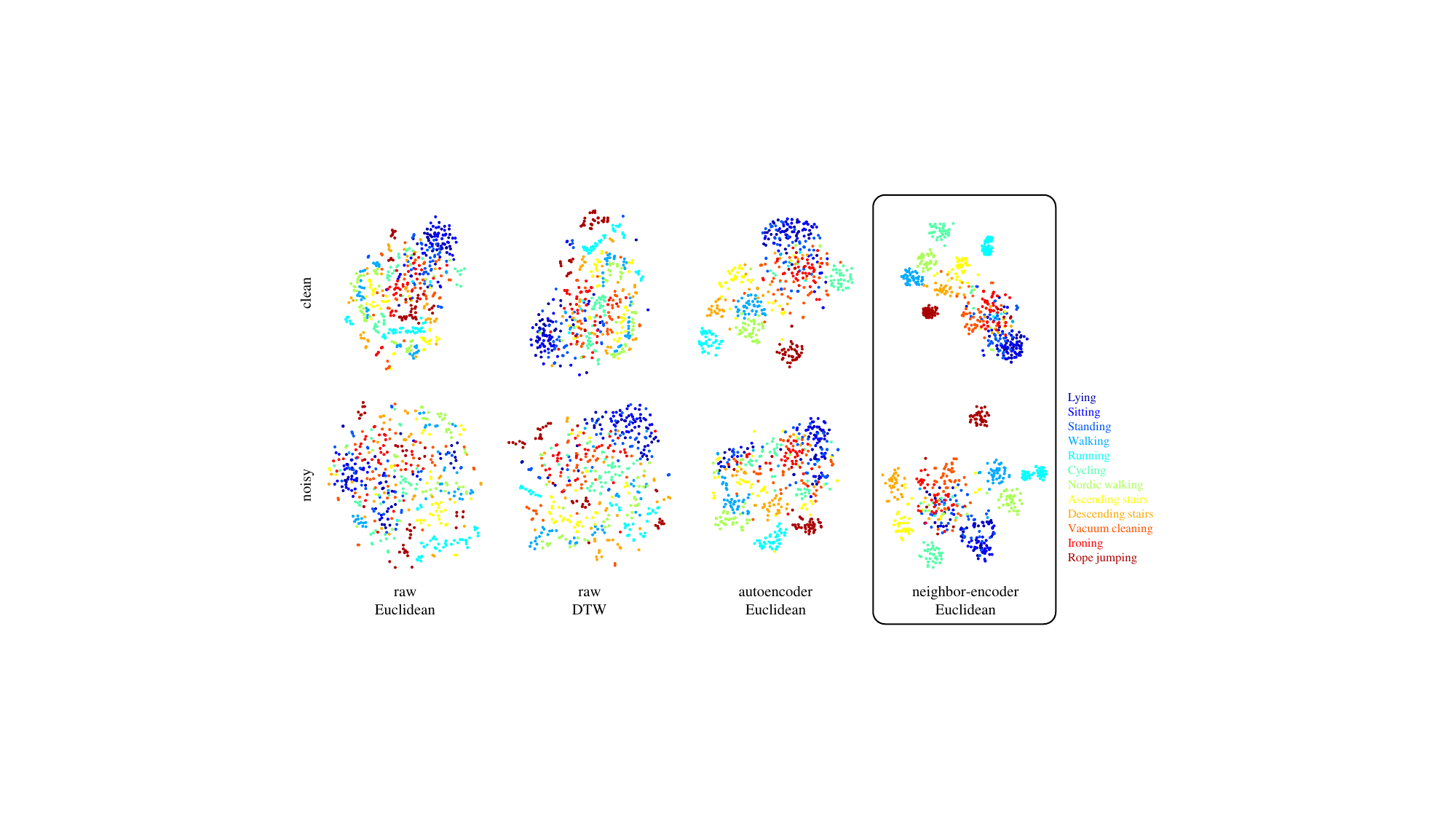}
\caption{
Visualizing the learned representation versus the raw time series on the PAMAP2 (human physical activity) dataset~\cite{reiss2012abra, reiss2012iswc} using $t$-SNE~\cite{maaten2008jmlr} with either Euclidean or dynamic time warping (DTW) distance~\cite{nguyen2017arxiv}.
If we manually select $27$ dimensions of the time series that are \emph{clean} and relevant (acceleration, gyroscope, magnetometer, etc.), the representation learned by both autoencoder and neighbor-encoder achieves better class separation than raw data. 
However, if the data includes \emph{noisy} and/or irrelevant dimensions (heart rate, temperature, etc.), neighbor-encoder outperforms autoencoder noticeably.
}
\label{fig_tsne_scatter}
\end{figure}

In this chapter, we propose an unsupervised representation learning framework (i.e., neighbor-encoder) which is \emph{general}, as it can be applied to various types of data, and \emph{versatile} since domain knowledge can be added by adopting various ``off-the-shelf'' data mining algorithms for finding neighbors.
Figure~\ref{fig_tsne_scatter} previews the $t$-SNE~\cite{maaten2008jmlr} visualization produced from a human physical activity dataset (see Section~\ref{exp_activity} for details).
The embedding is generated by projecting representation learned by neighbor-encoder, autoencoder, and raw data respectively to $2D$.
By using a suitable neighbor finding algorithm, the representation learned by neighbor-encoder provides a more meaningful visualization than its rival methods.

In summary, our major contributions include:
\begin{itemize}
\item We propose a \emph{general} and \emph{versatile} framework, neighbor-encoder, which incorporates domain knowledge into unsupervised representation learning by leveraging a large family of off-the-shelf similarity search techniques.
\item We demonstrate that the \emph{performance} of the representations learned by neighbor-encoder is superior to representations learned by autoencoder.
\item We showcase the \emph{applicability} of neighbor-encoder in a diverse set of domains (i.e., handwritten digit data, text, and human physical activity data) for various data mining tasks (i.e., classification, clustering, and visualization).
\end{itemize}

To allow reproducibility, all the codes and models associated with this chapter can be downloaded from~\cite{nnwebsite}.
The rest of this chapter is organized as follows:
In Section~\ref{relatedwork} we consider related work. 
Section~\ref{framework} we introduce the propose neighbor-encoder framework. 
We perform a comprehensive evaluation in Section~\ref{exp} before offering conclusions and directions for future research in Section~\ref{conclusion}.

\section{Background and Related Work}
\label{relatedwork}
\textbf{Unsupervised representation learning} is usually achieved by optimizing either domain-specific objectives or general unsupervised objectives.
For example, in the domain of computer vision and music processing, unsupervised representation learning problem is formulated as a supervised learning problem with surrogate labels, generated by exploiting the temporal coherence in videos and music~\cite{agrawal2015iccv, huang2017arxiv, jayaraman2015iccv, pathak2017cvpr, wang2015iccv}. 
In the case of natural language processing, word embedding can be achieved by optimizing an objective function that ``pushes'' words occurring in a similar context (i.e., surrounded by similar words) closer in the embedding space~\cite{mikolov2013arxiv, mikolov2013nips}.
Alternatively, general unsupervised objectives are also useful for unsupervised representation learning.
For example, both autoencoder~\cite{bengio2007nips, huang2007cvpr, vincent2008icml, vincent2010jmlr} and dictionary learning~\cite{mairal2009icml, su2014tmm} are based on minimizing the self-reconstruction error, while optimizing the $k$-means objective is shown effective in~\cite{coates2012nn} and~\cite{yang2017icml}.
Other objectives, such as self-organizing map criteria~\cite{bojanowski2017icml, kohonen1982bc} and adversarial training~\cite{donahue2016arxiv, goodfellow2014nips, larsen2015arxiv,  radford2015arxiv}, are also effective objectives for unsupervised representation learning.

\textbf{Autoencoder} is a decade-old unsupervised learning framework for dimension reduction, representation learning, and deep hierarchical model pre-training; many variants have been proposed since its initial introduction~\cite{bengio2007nips, goodfellow2016book}.
For example, the denoising autoencoder reconstructs the input data from its corrupted version; such modification improves the robustness of the learned representation~\cite{vincent2010jmlr}.
Variational autoencoder (VAE) regularizes the learning process by imposing a standard normal prior over the latent variable (i.e., representation), and such constraints help the autoencoder learn a valid generative model~\cite{kingma2013arxiv, rezende2014arxiv}.
Larsen et al.~\cite{larsen2015arxiv} and Makhzani et al.~\cite{makhzani2015arxiv} further improves generative model learning by combining VAE with adversarial training.
Sparsity constraints on the learned representation are another form of regularization for autoencoder to learn a more discriminating representation for classification; both the $k$-sparse autoencoder~\cite{makhzani2013arxiv, makhzani2015nips} and $k$-competitive autoencoder~\cite{chen2017kdd} incorporate such ideas.

\section{Neighbor-encoder Framework}
\label{framework}
\begin{figure}[htb]
\centering
\begin{minipage}[b]{0.30\columnwidth}
\includegraphics[trim={14.8cm 8.25cm 14.9cm 8.55cm}, clip, width=\textwidth,page=2]{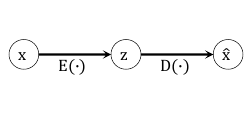}
\subcaption{}
\label{fig_scheme_auto}
\end{minipage}\hskip 4.5em
\begin{minipage}[b]{0.30\columnwidth}
\includegraphics[trim={14.8cm 8.25cm 14.9cm 8.55cm}, clip, width=\textwidth,page=3]{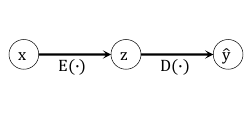}
\subcaption{}
\label{fig_scheme_neighbor}
\end{minipage}\hskip 4.5em
\begin{minipage}[b]{0.30\columnwidth}
\includegraphics[trim={14.8cm 8.1cm 14.9cm 8.4cm}, clip, width=\textwidth,page=4]{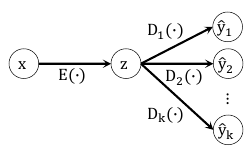}
\subcaption{}
\label{fig_scheme_kneighbor1}
\end{minipage}
\caption{
Various encoder-decoder configurations for training autoencoder and neighbor-encoder:
\subref{fig_scheme_auto}) autoencoder, \subref{fig_scheme_neighbor}) neighbor-encoder, and \subref{fig_scheme_kneighbor1}) $k$-neighbor-encoder with $k$ decoders.
}
\label{fig_scheme}
\end{figure}

In this section, we introduce the proposed neighbor-encoder framework and make a comparison with autoencoder.
Figure~\ref{fig_scheme} shows different  encoder-decoder configurations for both neighbor-encoder and autoencoder.
In the following sections, we discuss the motivation and design of each encoder-decoder configuration in detail.

\setcounter{subfigure}{0}

\subsection{Autoencoder} 
\label{framework_auto}
The overall architecture of autoencoder (AE) consists of two components: an \emph{encoder} and a \emph{decoder}.
Given input data $x$, the encoder $E(\cdot)$ is a function that encodes $x$ into a latent representation $z$ (usually in a lower dimensional space), and the decoder $D(\cdot)$ is a function that decodes $z$ in order to reconstruct $x$.
Figure~\ref{fig_scheme}.\ref{fig_scheme_auto} shows the feed-forward path of an autoencoder where $z = E(x)$ and $\hat{x} = D(z)$.
We train the autoencoder by minimizing the difference between the input data $x$ and the reconstructed data $\hat{x}$. 
Formally, given a set of training data $X$, the parameters in $E(\cdot)$ and $D(\cdot)$ are learned by minimizing the objective function $\sum_{x \in X}loss(x,\hat{x})$, where $\hat{x}=D(E(x))$.
The particular loss function we used in this chapter is cross entropy, but other loss function, like mean square error or mean absolute error can also be applied.
Once the autoencoder is learned, any given data can be projected to the latent representation space with $E(\cdot)$.
Both the encoder and the decoder can adopt any existing neural network architecture, such as multilayer perceptron~\cite{bengio2007nips}, convolutional net~\cite{huang2007cvpr}, or long short-term memory~\cite{hochreiter1997neural, srivastava2015icml}.

\subsection{Neighbor-encoder} 
Similar to the autoencoder, neighbor-encoder (NE) also consists of an encoder and a decoder.
Both the encoder and the decoder in neighbor-encoder work similarly to their counterparts in autoencoder; the major difference is in the objective function.
Given input data $x$ and the neighborhood function $N(\cdot)$ (which returns the neighbor $y$ of $x$), the encoder $E(\cdot)$ is a function that encodes $x$ into a latent representation $z$, and the decoder $D(\cdot)$ is a function that reconstructs $x$'s neighbor $y$ by decoding $z$.
Figure~\ref{fig_scheme}.\ref{fig_scheme_neighbor} shows the feed-forward path of a neighbor-encoder where $z = E(x)$ and $\hat{y} = D(z)$.
Formally, given a set of training data $X$ and a neighborhood function $N(\cdot)$, the neighbor-encoder is learned by minimizing the objective function $\sum_{x \in X}loss(y, \hat{y})$, where $y=N(x)$ and $\hat{y}=D(E(x))$.
Neighbor-encoder can be considered as a generalization of autoencoder as the input data can be treated as the nearest neighbor of itself with zero distance.
Note that here \emph{neighbor} can be defined in a variety of ways.
We will introduce examples of different neighbor definitions later in Section~\ref{nnfun}.

\begin{figure}[htb]
\centering
\begin{minipage}[b]{0.21\textwidth}
\includegraphics[trim={10cm 6cm 12cm 5cm}, clip, width=\textwidth,page=5]{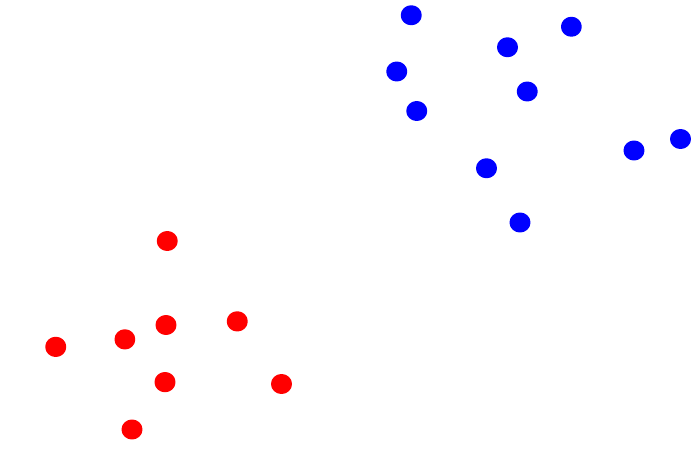}
\subcaption{}
\label{fig_intuition_auto}
\end{minipage}\hskip 4.5em
\begin{minipage}[b]{0.21\textwidth}
\includegraphics[trim={10cm 6cm 12cm 5cm}, clip, width=\textwidth,page=6]{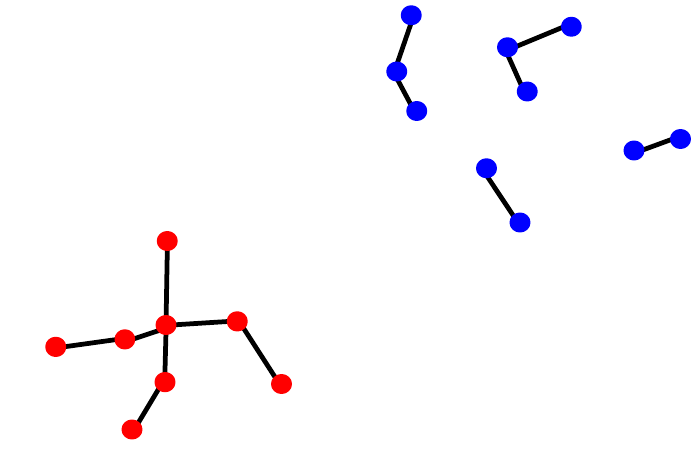}
\subcaption{}
\label{fig_intuition_neighbor0}
\end{minipage}\hskip 4.5em
\begin{minipage}[b]{0.21\textwidth}
\includegraphics[trim={10cm 6cm 12cm 5cm}, clip, width=\textwidth,page=7]{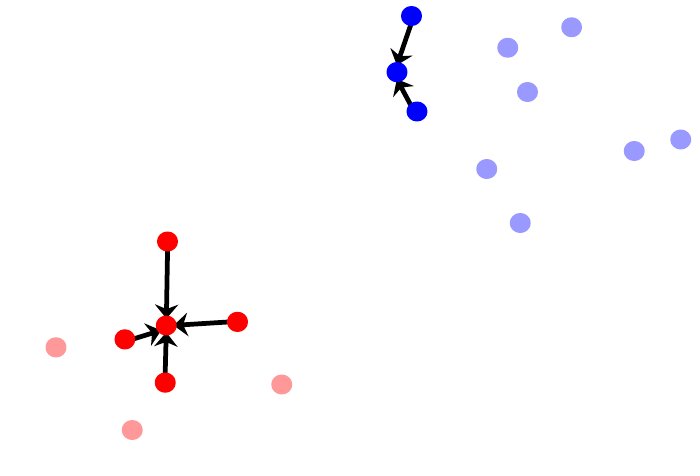}
\subcaption{}
\label{fig_intuition_neighbor1}
\end{minipage}
\caption{
Intuition behind neighbor-encoder compared to autoencoder.
\subref{fig_intuition_auto}) A simple $2D$ dataset with two classes, 
\subref{fig_intuition_neighbor0}) the nearest neighbor graph constructed for the dataset (arrowheads are removed for clarity), and
\subref{fig_intuition_neighbor1}) an example of how neighbor-encoder would generate representation, with smaller intra-class variation for highlighted data points.
}
\label{fig_intuition_comparison}
\end{figure}

Compared to autoencoder, we argue that neighbor-encoder can better retain the similarity between data samples in the latent representation space.
Figure~\ref{fig_intuition_comparison} builds a case for this claim.
As shown in Figure~\ref{fig_intuition_comparison}.\ref{fig_intuition_auto}, we assume the dataset of interest consists of samples from two classes (i.e., blue class and red class, and each class forms a cluster) in $2D$ space.
Since the autoencoder is trained by mapping each data point to itself, the learned representation for this dataset would most likely be a rotated and/or re-scaled version of Figure~\ref{fig_intuition_comparison}.\ref{fig_intuition_auto}.
In contrast, the neighbor-encoder (trained with nearest neighbor relation, as shown in Figure~\ref{fig_intuition_comparison}.\ref{fig_intuition_neighbor0}) would learn a representation with much less intra-class variation.
As Figure~\ref{fig_intuition_comparison}.\ref{fig_intuition_neighbor1} shows, when several similar data points share the same nearest neighbor, the objective function will force the network to generate exactly the same output for these similar data points, thus forcing their latent representation (which is the input of the decoder) to be very similar. 

\setcounter{subfigure}{0}

Alternatively, neighbor-encoder can be understood as a non-parametric way of generating corrupted data for denoising autoencoder.
Instead of being trained to remove arbitrary noise (e.g., Gaussian noise) from the corrupted data (which is the norm), the neighbor-encoder is trained to remove more meaningful noise from the corrupted data.
For example, a pair of nearest neighbors found using Euclidean distance in MNIST database~\cite{lecun1998ieee} usually reflects different writing styles of the same numeric digit (see Figure~\ref{fig_mnist_pair}.\ref{fig_mnist_pair_0}).
By training the neighbor-encoder with such nearest neighbor pairs, the learning process would push the encoder network to ignore or ``remove" the writing style aspect from the handwritten digits.

Since we are using neighbor finding algorithms to guide the representation learning process, one may argue that we could instead construct a graph using the neighbor finding algorithm, then apply various graph-based representation learning methods like the ones proposed in~\cite{dong2017kdd, grover2016kdd, perozzi2014kdd, ribeiro2017kdd, tang2015www}.
Graph-based methods are indeed valid alternatives to neighbor-encoder; however, they have the following two limitations:
1) If one wishes to encode a newly obtained data, the out-of-sample problem would bring about additional complexity, as these methods are not designed to handle such a scenario. 
2) It will be impossible to learn a generative model, as graph-based methods learn the representation by modeling the relationship between examples in a dataset, rather than modeling the example itself.
As a result, whenever the above limitations are crucial, the proposed neighbor-encoder is preferred over the graph-based methods.

\subsection{$k$-neighbor-encoder}
\label{framework_kne} 
Similar to the idea of generalizing the $1$-nearest neighbor classifier to a $k$-nearest neighbor classifier, neighbor-encoder can also be extended to the $k$-neighbor-encoder by reconstructing $k$ neighbors of the input data (see Figure~\ref{fig_scheme}.\ref{fig_scheme_kneighbor1}).
We train $k$ decoders to simultaneously reconstruct all $k$ neighbors of the input. 
Given an input data $x$ and the neighborhood function $N(\cdot)$ (which returns the $k$ neighbors $[y_i | \forall i \in \mathbb{Z}:0<i \leq k]$ of $x$), the encoder $E(\cdot)$ is a function that encodes $x$ into the latent representation $z$.
Then, we have a set of $k$ decoders $[D_i(\cdot) | \forall i \in \mathbb{Z}:0<i \leq k]$, in which each individual function $D_i(\cdot)$ decodes $z$ in order to reconstruct $x$'s $i$th neighbor $y_i$.

The $k$-neighbor encoder learning process is slightly more complicated than the neighbor-encoder (i.e., $1$-neighbor-encoder).
\sloppy Given a set of training data $X$ and a neighborhood function $N(\cdot)$, the $k$-neighbor-encoder can be learned by minimizing $\sum_{x \in X}\sum_{y_i \in N(x)}loss(y_i, \hat{y_i})$ where $\hat{y_i}=D_i(E(x))$ and $0<i\leq k$.
Note that since there are $k$ decoders, we need to assign each $y_i$ to one of the decoders.
If there are ``naturally'' $k$ types of neighbors, we can train one decoder for each type of neighbor.
Otherwise, one possible decoder assignment strategy is choosing the decoder that provides the lowest reconstruction loss for each $y_i \in N(x)$.
This decoder assignment strategy will work if each training example has less than $k$ ``modes" of neighbors.

\subsection{Neighborhood Function}
\label{nnfun} 
To use any of the introduced neighbor-encoder configurations, we need to properly define the term neighbor.
In this section, we discuss several possible neighborhood functions for the neighbor-encoder framework.
Note that the functions listed in this section are just a small subset of all the available functions, and were chosen because they demonstrate the versatility of our approach.

\textbf{Simple Neighbor}
is defined as the objects that are closest to a given object in Euclidean distance or other distances, assuming the distance between every two objects is computable.
For example, given a set of objects $[x_1, x_2, x_3, ..., x_n]$ where each object is a real-value vector, the neighboring relationship among the objects under Euclidean distance can be approximately identified by construing a $k$-$d$ tree.

\textbf{Feature Space Neighbor}
is very similar to \emph{simple neighbor}, except that instead of computing the distance between objects in the space where the reconstruction is performed (e.g., the raw-data space), we compute the distance in an alternative representation or feature space.
To give a more concrete example, suppose we have a set of objects $[x_1, x_2, x_3, ..., x_n]$ where each object is an audio clip in mel-frequency spectrum space.
Instead of finding neighbors directly in the mel-frequency spectrum space, we transform the data into the Mel-Frequency Cepstral Coefficient (MFCC) space, as neighbors discovered in MFCC space are semantically more meaningful and searching in MFCC space is more efficient.

\textbf{Time Series Subspace Neighbor}
, as defined for multidimensional time series data, is the similarity between two objects measured by using only a subset of all dimensions.
By ignoring some dimensions, a time series could find higher quality neighbors since it is very likely that some of the dimensions contain irrelevant or noisy information (e.g., room temperature in human physical activity data).  
Given a multidimensional time series, we can use $m$STAMP/$m$STOMP~\cite{yeh2017icdm} (See Section~\ref{ch2secmstamp}) to evaluate the neighboring relationship between all the subsequences within the time series.

\textbf{Spatial or Temporal Neighbor}
defines the neighbor based on the spatial or temporal closeness of objects. 
Specifically, given a set of objects $[x_1, x_2, x_3, ..., x_n]$ where the subscript denotes the temporal (or spatial) arrival order, $x_i$ and $x_j$ are neighbors when $|i - j| < d$, where $d$ is the predefined size of the neighborhood.
The skip-gram model in word2vec~\cite{mikolov2013arxiv, mikolov2013nips} is an example of spatial neighbor-encoder, as the skip-gram model can be regarded as reconstructing the spatial neighbors (in the form of one-hot vector) of a given word.

\textbf{Side Information Neighbor}
defines the neighbor with side information, which could be more semantically meaningful than the aforementioned functions.
For example, images shown in the same eCommerce webpage (e.g., Amazon~\cite{amazon}) would most likely belong to the same merchandise, but they can reflect different angles, colors, etc., of the merchandise.
If we select a random image from a webpage and assign it as the nearest neighbor for all the other images in the same page, we could train a representation that is invariant to view angles, lighting conditions, product variations (e.g., different color of the same smart phone), and so forth.
One may consider that using such side information implies a supervised learning system instead of an unsupervised learning system.
However, note that we only have the information regarding similar pairs while the information regarding dissimilar pairs (i.e., negative examples) is missing\footnote{
We can construct a $1$-nearest-neighbor graph by treating each image as a node and connecting each image with its nearest neighbor. 
One may sample pairs of disconnected nodes as negative examples, but such sampling method may produce false negatives, as disconnected nodes may or may not be semantically dissimilar.
}; compared to the information required by a conventional supervised learning system, this information is very limiting.

\section{Experimental Evaluation}
\label{exp}
In this section, we show the effectiveness and versatility of neighbor-encoder compared to autoencoder by performing experiments on handwritten digits, texts, and human physical activities with different neighborhood functions.
As the neighbor-encoder framework is a generalization of autoencoder, all the variants of autoencoder (e.g., denoising autoencoder~\cite{vincent2010jmlr}, variational autoencoder~\cite{kingma2013arxiv, rezende2014arxiv}, $k$-sparse autoencoder~\cite{makhzani2013arxiv, makhzani2015nips}, or adversarial autoencoder~\cite{larsen2015arxiv, makhzani2015arxiv}) can be directly ported to the neighbor-encoder framework.
As a result, we did not exhaustively test all variants of autoencoder/neighbor-encoder, but instead selected the three most popular variants (i.e., vanilla, denoising, and variational).
We leave the exhaustive comparison of the other variants for future work.

\subsection{Handwritten Digits Case Study}
\label{exp_mnist}
\begin{figure}[htb]
\centering
\includegraphics[trim={6.5cm 3.5cm 6.5cm 3.5cm}, clip, width=0.6\columnwidth,page=8]{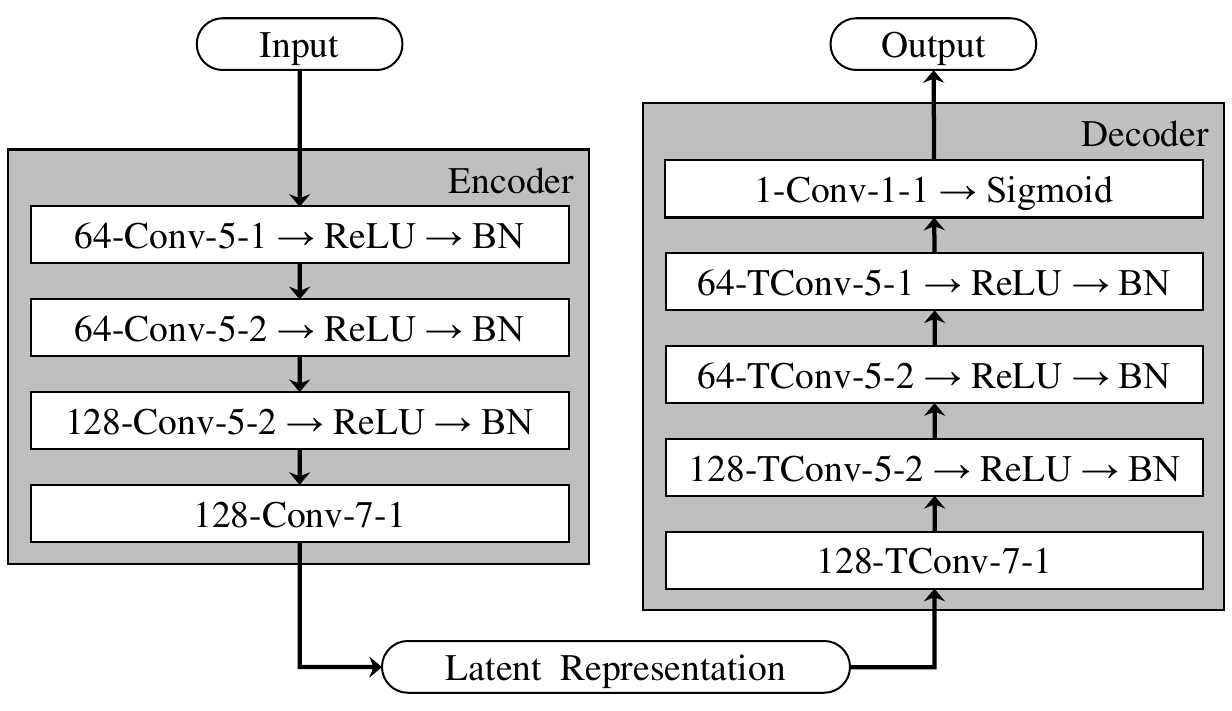}
\caption{
Network architecture for the encoder and the decoder.
64-Conv-5-1 denotes $2D$ convolutional layer with $64$ $5\times5$ kernels and stride of $1$.
ReLU denotes rectified linear unit.
BN denotes batch normalization.
TConv denotes transposed $2D$ convolutional layers.
}
\label{mnist_network}
\end{figure}

The MNIST database is commonly used in the initial study of newly proposed methods due to its simplicity~\cite{lecun1998ieee}.
It contains $70,000$ images of handwritten digits (one digit per image); $10,000$ of these images are test data, and the other $60,000$ are training data.
The original task for the dataset is multi-class classification.
Since the proposed method is not a classifier but a representation learner (i.e., an encoder), we have evaluated our method using the following procedure: 
1) we train the encoder with all the training data, 
2) we encode both training data and test data into the learned representation space, 
3) we train a simple classifier (i.e., linear support vector machine/SVM) with various amounts of labeled training data in the representation space, then apply the classifier to the representation of test data and report the classification error (i.e.,  semi-supervised classification problem), and
4) we also apply a clustering method (i.e., $k$-means) to the representation of test data and report the adjusted Rand index.
As a proof of concept, we did not put much effort in optimizing the structure of the encoder/decoder.
We simply used a $4$-layer $2D$ convolutional net as the encoder and a $4$-layer transposed $2D$ convolutional net as the decoder.
The detailed setting of the network architecture is summarized in Figure~\ref{mnist_network}.
We have tried several other convolutional net architectures as well; we draw the same conclusion from the experimental results with these alternative architectures.

\begin{figure}[htb]
\centering
\includegraphics[trim={8.5cm 5.1cm 8.0cm 5.0cm}, clip, width=0.65\columnwidth,page=9]{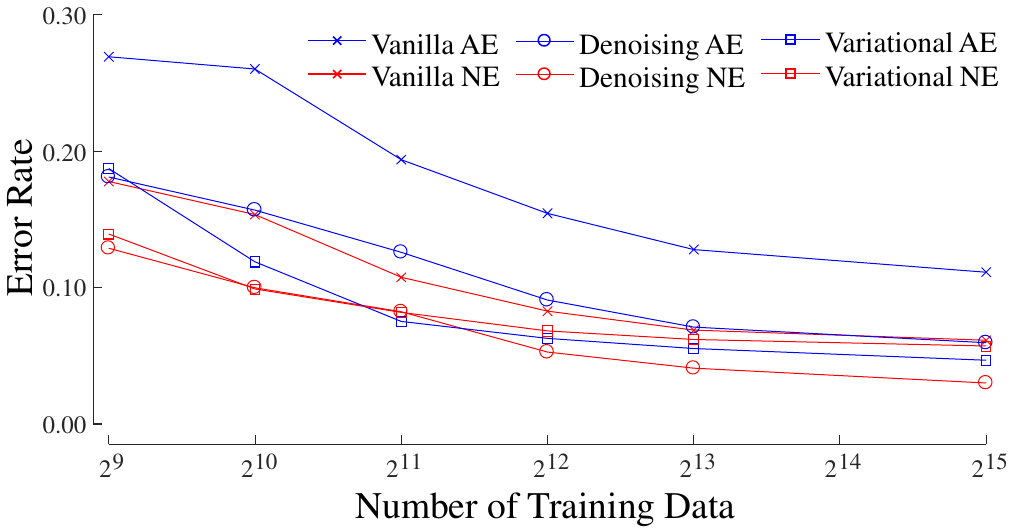}
\caption{
The classification error rate with linear SVM versus various training data sizes using different variants (i.e., vanilla, denoising, variational) of autoencoder and neighbor-encoder.
}
\label{fig_mnist_svm}
\end{figure}

Here we use the neighbor-encoder configuration (
Figure~\ref{fig_scheme}.\ref{fig_scheme_neighbor}) with the simple neighbor definition for our neighbor-encoder.
We compare the performance of three variants (vanilla, denoising, and variational) of neighbor-encoder and the same three variants of autoencoder.
Figure~\ref{fig_mnist_svm} shows the classification error rate as we change the number of labeled training data for linear SVM.
All neighbor-encoder variants outperform their corresponding autoencoder variants, except the variational neighbor-encoder when the number of labeled training data is larger.
Overall, denoising neighbor-encoder produces the most discriminating representations.

Besides the semi-supervised learning experiment, we also performed a purely unsupervised clustering experiment with $k$-means.
Table~\ref{tab_mnist_kmeans} summarizes the experiment's result.
The overall conclusion is similar to that of the semi-supervised learning experiment, where all neighbor-encoder variants outperformed their corresponding autoencoder variants.
Unlike the semi-supervised experiment, variational neighbor-encoder produces the most clusterable representations in this particular experiment, but all three variants of neighbor-encoder are comparable with each other.

\begin{table}[htb]
\centering
\caption{
The clustering adjust Rand index with $k$-means. 
}
\label{tab_mnist_kmeans}
\begin{tabular}{l|ccc}
& Vanilla & Denoising & Variational \\ \hline
AE & 0.3005 & 0.3710 & 0.4492 \\
NE & 0.4926 & 0.5039 & 0.5179 \\
\end{tabular}
\end{table}

\newpage
In the previous two experiments, we define the neighbor of an object as its nearest neighbor under Euclidean distance. 
With this definition, the visual difference between an object and its neighbor is usually small, given that we have sufficient data.
To allow for more visual discrepancy between the objects and their neighbors, we could change that neighbor definition to the $i$th nearest neighbor under Euclidean distance ($i>1$).
We have repeated the clustering experiment under different settings of $i$ to examine the effect of increasing discrepancy between the objects and their neighbors.
We chose to perform the clustering experiment instead of the semi-supervised learning experiment because 1) clustering is unsupervised and 2) it is easier to present the clustering result in a single figure, as semi-supervised learning requires varying both the amount of training data and~$i$.

\setcounter{subfigure}{0}
\begin{figure}[htb]
\centering
\begin{minipage}[b]{0.12\columnwidth}
\includegraphics[trim={1.2cm 0.8cm 0.35cm 0.4cm}, clip, width=\textwidth]{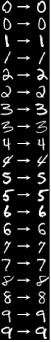}
\subcaption{$2^0$}
\label{fig_mnist_pair_0}
\end{minipage}~
\begin{minipage}[b]{0.12\columnwidth}
\includegraphics[trim={1.2cm 0.8cm 0.35cm 0.4cm}, clip, width=\textwidth]{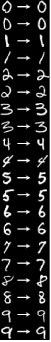}
\subcaption{$2^4$}
\label{fig_mnist_pair_16}
\end{minipage}~
\begin{minipage}[b]{0.12\columnwidth}
\includegraphics[trim={1.2cm 0.8cm 0.35cm 0.4cm}, clip, width=\textwidth]{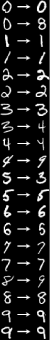}
\subcaption{$2^8$}
\label{fig_mnist_pair_256}
\end{minipage}~
\begin{minipage}[b]{0.12\columnwidth}
\includegraphics[trim={1.2cm 0.8cm 0.35cm 0.4cm}, clip, width=\textwidth]{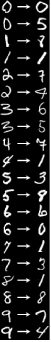}
\subcaption{$2^{12}$}
\label{fig_mnist_pair_4096}
\end{minipage}
\caption{
Neighbor pairs under different proximity setting.
}
\label{fig_mnist_pair}
\end{figure}

\begin{figure}[htb]
\centering
\includegraphics[trim={7.0cm 5.1cm 6.6cm 5.0cm}, clip, width=0.75\columnwidth,page=10]{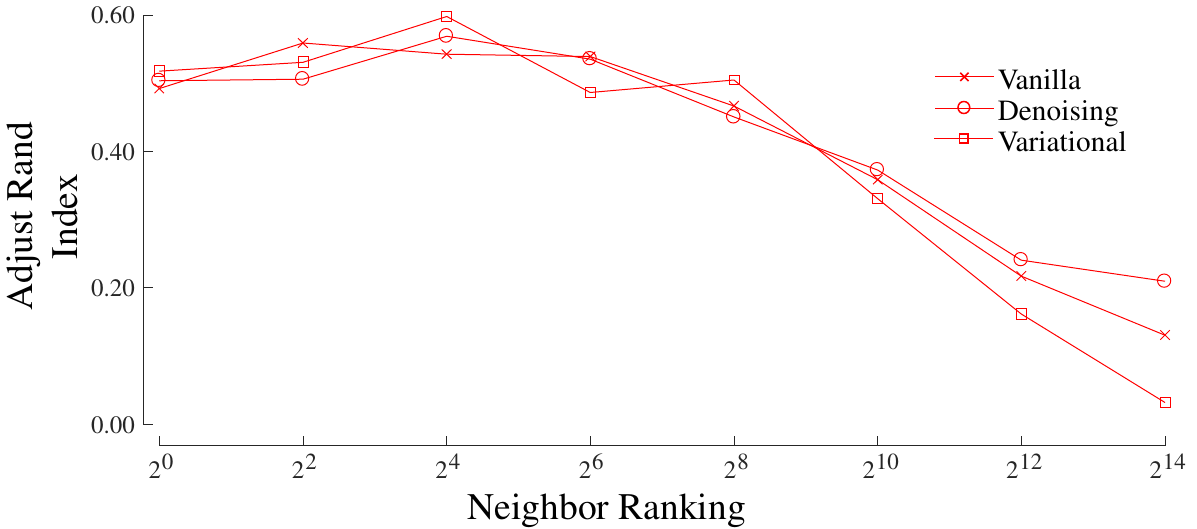}
\caption{
The clustering adjust Rand index versus the proximity of the neighbor using various neighbor-encoder variations (i.e., vanilla, denoising, variational).
The proximity of a neighbor is defined as its ranking when queried with the input.
}
\label{fig_mnist_kth_nebr}
\end{figure}

Figure~\ref{fig_mnist_kth_nebr} summarizes the result, and Figure~\ref{fig_mnist_pair} shows a randomly selected set of object-neighbor pairs under different settings of $i$.
The performance peaks around $i=2^4$ and decreases as we increase $i$; therefore, choosing the $2^4$th nearest neighbor as the reconstruction target for neighbor-encoder would create enough discrepancy between the object-neighbor pair for better representation learning.
When neighbor-encoder is used in this fashion, it can be regarded as a non-parametric way of generating noisy objects (similar as the principle of denoising autoencoder), and the settings of $i$ controls the amount of noise added to the object.
Note that neighbor-encoder is not equivalent to denoising autoencoder, as several objects can share the same $i$th nearest neighbor (recall Figure~\ref{fig_intuition_comparison}.\ref{fig_intuition_neighbor1}), but denoising autoencoder would most likely generate different noisy inputs for different objects.

\begin{figure}[htb]
\centering
\begin{minipage}[b]{0.20\columnwidth}
\includegraphics[trim={1.1cm 0.8cm 0.5cm 0.4cm}, clip, width=\textwidth]{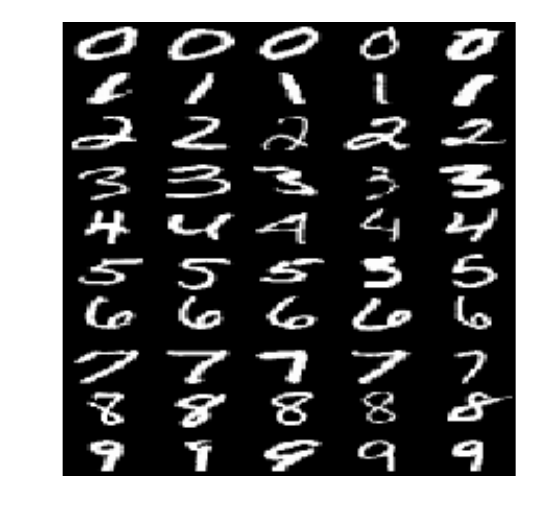}
\subcaption{Input}
\label{fig_mnist_input}
\end{minipage}~~~
\begin{minipage}[b]{0.20\columnwidth}
\includegraphics[trim={1.1cm 0.8cm 0.5cm 0.4cm}, clip, width=\textwidth]{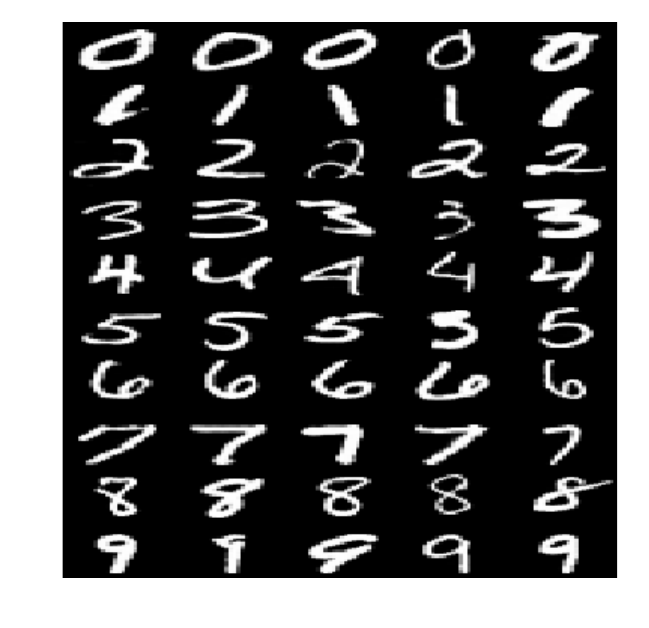}
\subcaption{Vanilla AE}
\label{fig_mnist_self_ae}
\end{minipage}~
\begin{minipage}[b]{0.20\columnwidth}
\includegraphics[trim={1.1cm 0.8cm 0.5cm 0.4cm}, clip, width=\textwidth]{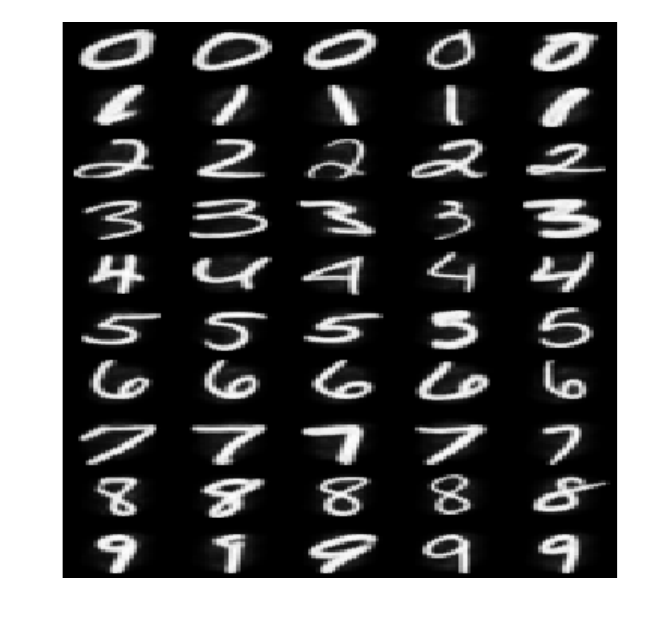}
\subcaption{Denoising AE}
\label{fig_mnist_self_dae}
\end{minipage}~
\begin{minipage}[b]{0.20\columnwidth}
\includegraphics[trim={1.1cm 0.8cm 0.5cm 0.4cm}, clip, width=\textwidth]{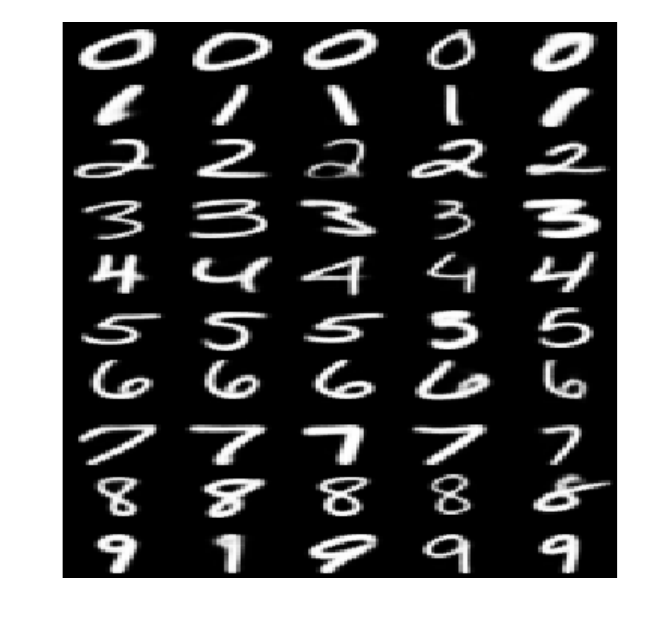}
\subcaption{Variational AE}
\label{fig_mnist_self_vae}
\end{minipage}\\ \hspace{0.20\columnwidth}~~~
\begin{minipage}[b]{0.20\columnwidth}
\includegraphics[trim={1.1cm 0.8cm 0.5cm 0.4cm}, clip, width=\textwidth]{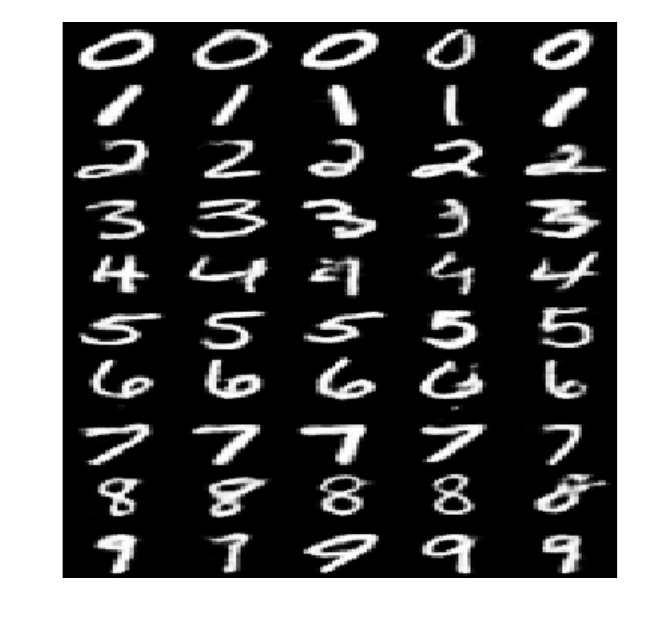}
\subcaption{Vanilla NE}
\label{fig_mnist_nebr_ae}
\end{minipage}~
\begin{minipage}[b]{0.20\columnwidth}
\includegraphics[trim={1.1cm 0.8cm 0.5cm 0.4cm}, clip, width=\textwidth]{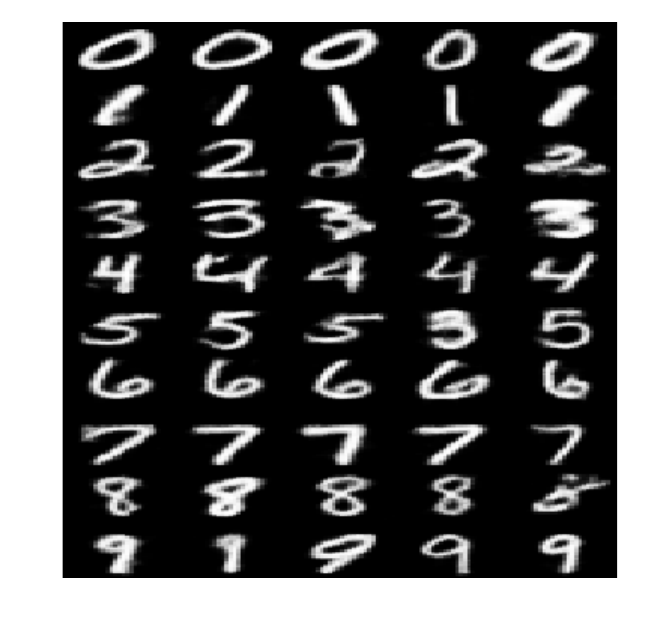}
\subcaption{Denoising NE}
\label{fig_mnist_nebr_dae}
\end{minipage}~
\begin{minipage}[b]{0.20\columnwidth}
\includegraphics[trim={1.1cm 0.8cm 0.5cm 0.4cm}, clip, width=\textwidth]{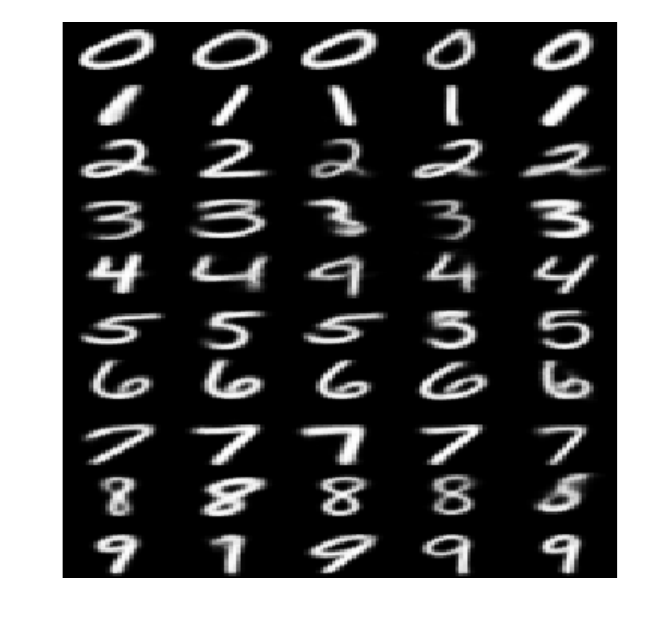}
\subcaption{Variational NE}
\label{fig_mnist_nebr_vae}
\end{minipage}\\
\caption{
Outputs of the decoders for different autoencoder (AE) and neighbor-encoder (NE) variations.
}
\label{fig_mnist_recon}
\end{figure}

To explain the performance difference between autoencoder and neighbor-encoder, we randomly selected five test examples from each class (see Figure~\ref{fig_mnist_recon}.\ref{fig_mnist_input}) and fed them through both the autoencoder and the neighbor-encoder trained in the previous experiments.  
The outputs are shown in Figure~\ref{fig_mnist_recon}, where the top row and bottom row are autoencoder and neighbor-encoder respectively.
As expected, the output of autoencoder is almost identical to the input image.
Although the output of neighbor-encoder is still very similar to the input image, the intra-class variation is less than the output of autoencoder. 
This is because neighbor-encoder tends to reconstruct the same neighbor image from similar input data points (recall Figure~\ref{fig_intuition_comparison}.\ref{fig_intuition_neighbor1}). 
As a result, the latent representation learned by neighbor-encoder is able to achieve better performances.

\setcounter{subfigure}{0}

\subsection{Text Documents Case Study}
\label{exp_texts}
The 20Newsgroup\footnote{downloaded from~\cite{cardoso2007phd}} dataset contains nearly 20,000 newsgroup posts grouped into 20 (almost) balanced newsgroups/classes.
It is a popular dataset for experimenting with machine learning algorithms on text documents.
We follow the clustering experiment setup presented in~\cite{yang2017icml}, wherein each document is represented as a tf-idf vector (using the 2,000 most frequent words in the corpus), and the performance of a method is measured by the normalized mutual information (NMI)~\cite{cai2011ltkde}, adjusted Rand index (ARI)~\cite{yeung2001bioinformatics}, and clustering accuracy (ACC)~\cite{cai2011ltkde}.
To ensure the fairness of the comparison, we use an identical network structure (250-100-20 multilayer perceptron~\cite{yang2017icml}) for the encoder.

We test three different autoencoder variants (vanilla autoencoder/AE, denoising autoencoder/DAE, and variational autoencoder/VAE) as the baselines, and enhance the best variant with the neighbor-encoder objective function (denoising neighbor-encoder/DNE).
The neighbor definition adopted in this set of experiments is the feature space neighbor, where we find the nearest neighbor of each document in the current encoding space at each epoch.
We use $k$-means (KM) to cluster the learned representation.
Table~\ref{tab_texts_20news} shows our experiment results accompanied by the experiment result reported in~\cite{yang2017icml}.
The proposed method (neighbor-encoder), when combined with the best variant of autoencoder, outperforms all other methods.

\begin{table}[htb]
\centering
\begin{threeparttable}
\scriptsize
\caption{
The results of the experiment on 20Newsgroup. 
}
\label{tab_texts_20news}
\renewcommand\TPTminimum{\linewidth}
\begin{tabular}{p{0.2\linewidth}l|cccp{0.2\linewidth}}
& Methods & NMI & ARI & ACC & \\ \cline{2-5}
& JNKM\tnote{*}~\cite{yang2017tsp} & 0.40 & 0.10 & 0.24 & \\
& XARY\tnote{*}~\cite{kumar2013icml} & 0.19 & 0.02 & 0.18 & \\
& SC\tnote{*}~\cite{ng2002nips} & 0.40 & 0.17 & 0.34 & \\
& KM\tnote{*}~\cite{lloyd1982tit} & 0.41 & 0.15 & 0.30 & \\
& NMF+KM\tnote{*}~\cite{yang2017icml} & 0.39 & 0.17 & 0.33 & \\
& LCCF\tnote{*}~\cite{cai2011ltkde} & 0.46 & 0.17 & 0.32 & \\
& SAE+KM\tnote{*}~\cite{yang2017icml} & 0.47 & 0.28 & 0.42 & \\
& DCN\tnote{*}~\cite{yang2017icml} & 0.48 & 0.34 & 0.44 & \\ \cline{2-5}
& AE+KM & 0.44 & 0.29 & 0.43 & \\
& DAE+KM & 0.52 & 0.38 & 0.53 & \\
& VAE+KM & 0.41 & 0.18 & 0.31 & \\ \cline{2-5}
& DNE+KM & \textbf{0.56} & \textbf{0.41} & \textbf{0.57} & \\
\end{tabular}
\begin{tablenotes}
\item [\hspace{0.25\linewidth}*] Experiment results reported by~\cite{yang2017icml}.
\end{tablenotes}
\end{threeparttable}
\end{table}

The most similar systems (to our baselines) examined by~\cite{yang2017icml} is the stacked autoencoder with $k$-means (SAE+KM).
When comparing our baselines with SAE+KM, AE+KM unsurprisingly performs similar to SAE+KM, as they are almost identical.
Out of our three baselines, the denoising autoencoder outperforms the other two variants considerably, with the variational autoencoder being the worst system. 
Because the denoising is the best autoencoder variant, we decided to extend it with the neighbor reconstruction loss function.
The resulting system (DNE+KM) outperforms all other systems, including the previous state-of-the-art deep clustering network (DCN).

Finally, we apply DNE+KM to a larger dataset with imbalanced classes, RCV1-v2~\cite{lewis2004jmlr}, following the experiment/encoder setup with 20 clusters outlined in~\cite{yang2017icml}. 
Table~\ref{tab_texts_rcv1} summarizes the results. 
The performance of DNE+KM is similar to DCN in terms of NMI, while outperforming DCN in terms of ARI/ACC.

\begin{table}[htb]
\centering
\begin{threeparttable}
\scriptsize
\caption{
The result of the experiment on RCV1-v2 with 20 clusters. 
}
\label{tab_texts_rcv1}
\renewcommand\TPTminimum{\linewidth}
\begin{tabular}{p{0.2\linewidth}l|cccp{0.2\linewidth}}
& Methods & NMI & ARI & ACC & \\ \cline{2-5}
& XARY\tnote{*}~\cite{kumar2013icml} & 0.25 & 0.04 & 0.28 & \\
& DEC\tnote{*}~\cite{xie2016icml} & 0.08 & 0.01 & 0.14 & \\
& KM\tnote{*}~\cite{lloyd1982tit} & 0.58 & 0.29 & 0.47 & \\
& SAE+KM\tnote{*}~\cite{yang2017icml} & 0.59 & 0.33 & 0.46 & \\
& DCN\tnote{*}~\cite{yang2017icml} & \textbf{0.61} & 0.33 & 0.47 & \\ \cline{2-5}
& DNE+KM & 0.60 & \textbf{0.40} & \textbf{0.49} & \\
\end{tabular}
\begin{tablenotes}
\item [\hspace{0.25\linewidth}*] Experiment results reported by~\cite{yang2017icml}.
\end{tablenotes}
\end{threeparttable}
\end{table}

\subsection{Human Physical Activities Case Study}
\label{exp_activity}
In Section~\ref{framework}, we introduced the $k$-neighbor-encoder in addition to the neighbor-encoder.
Here we test the $k$-neighbor-encoder on the PAMAP2 dataset~\cite{reiss2012abra, reiss2012iswc} using the time series subspace neighbor definition~\cite{yeh2017icdm}.
We chose the subspace neighbor definition because 
1) it addresses one of the commonly seen multidimensional time series problem scenarios (the existence of irrelevant/noisy dimensions),  
2) it is able to extract meaningful repeating patterns, and
3) it na\"{\i}vely gives multiple ``types'' of neighbors to each object.

The PAMAP2 dataset was collected by mounting three inertial measurement units and a heart rate monitor on nine subjects, and recording them performing $18$ different physical activities (e.g., walking, running, playing soccer), with one session per subject, each ranging from $0.5$ hours to $1.9$ hours.
The subjects performed one activity for a few minutes, took a short break, then continued performing another activity.
In order to transfer the dataset into a format that we can use for evaluation (i.e., a training/test split), 
for each subject (or recording session) we cut the data into segments according to their corresponding physical activities; 
then, within each activity segment, we generated training data from the first half, and test data from the second half with a sliding window length of $100$ and a step size of one.
We make sure that there is no overlap between training data and test data.
After the reorganization, we end up with none datasets (one pair of training/test set per subject). 
We ran experiments on each dataset independently, and report averaged performance results.

\begin{figure}[htb]
\centering
\includegraphics[trim={5.5cm 3.5cm 5.5cm 3.5cm}, clip, width=0.6\columnwidth,page=11]{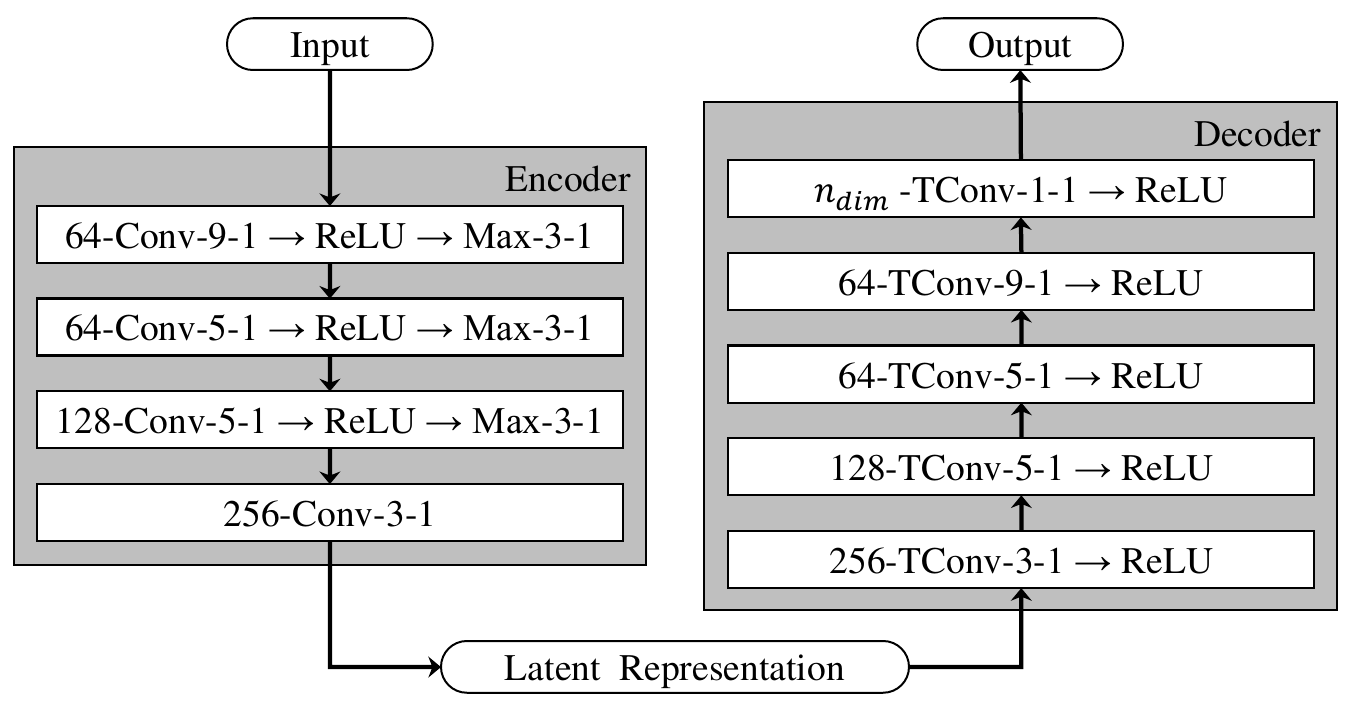}
\caption{
Network architecture for the encoder and the decoder.
64-Conv-9-1 denotes a $1D$ convolutional layer with $64$ sized $9$ kernels and sized $1$ stride.
ReLU denotes rectified linear layers.
Max-3-1 denotes a max pooling layer with sized $3$ pooling window and sized $1$ stride.
TConv denotes transposed $1D$ convolutional layers.
$n_{dim}$ is the number of dimensions for the multidimensional time series.
}
\label{pamp2_network}
\end{figure}

The experiment procedure is very similar to the one presented in Section~\ref{exp_mnist}.
We perform the experiments under two different scenarios: ``clean" and ``noisy."
In the ``clean'' scenario, we manually deleted some dimensions of the data that are irrelevant (or harmful) to the classification/clustering tasks,
while in the ``noisy'' scenario, all dimensions of the data are retained.  
The encoder-decoder network architecture we used is summarized in Figure~\ref{pamp2_network}.
Here we use a $5$-layer $1D$ convolutional net as the encoder, and a $5$-layer transposed $1D$ convolutional net as the decoder.
Similar to Section~\ref{exp_mnist}, we did not put much effort in optimizing the structure of this network architecture.
We have tried modifying the convolutional net architectures in various ways, such as adding batch normalization, changing the number of layers, or varying the number of filters for each layer, etc., and the conclusion drawn from the experimental results remains virtually unchanged. 
During test time, we applied the global average pooling to the output of the encoder to obtain the latent representation.

\begin{figure}[htb]
\begin{minipage}[b]{\columnwidth}
\centering
\includegraphics[trim={9.3cm 5.5cm 8.0cm 5.0cm}, clip, width=0.65\textwidth,page=12]{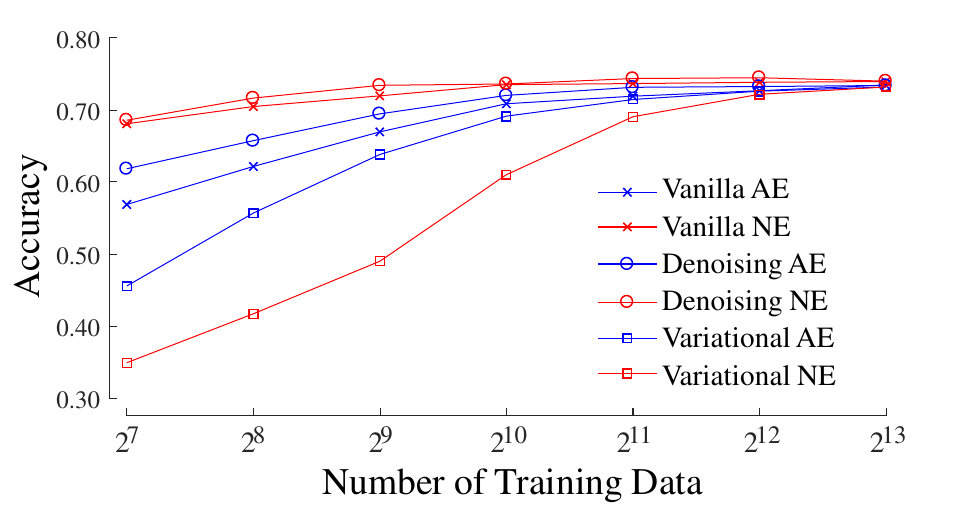}
\subcaption{Clean scenario}
\label{fig_pamap2_svm_clean}
\end{minipage}\\
\begin{minipage}[b]{\columnwidth}
\centering
\includegraphics[trim={9.3cm 5.5cm 8.0cm 5.0cm}, clip, width=0.65\textwidth,page=13]{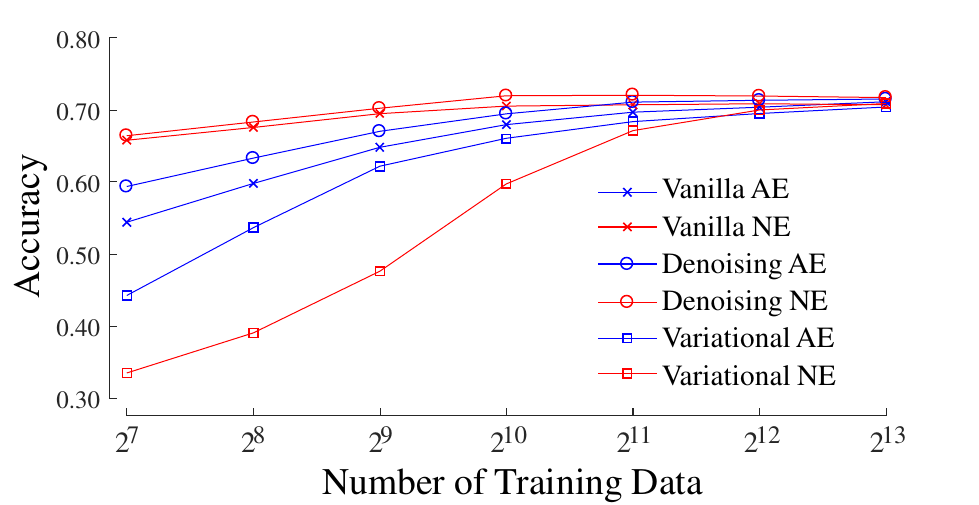}
\subcaption{Noisy scenario}
\label{fig_pamap2_svm_noisy}
\end{minipage}
\caption{
The classification accuracy with linear SVM versus various labeled training data size using different variants (i.e., vanilla, denoising, variational) of either autoencoder and $k$-neighbor-encoder.
}
\label{fig_pamap2_svm}
\end{figure}

In Figure~\ref{fig_pamap2_svm}, we compare the semi-supervised classification capability of vanilla, denoising, and variational autoencoder/$k$-neighbor-encoder under both the``clean" scenario and the ``noisy" scenario.
Both vanilla and denoising $k$-neighbor-encoder outperforms their corresponding autoencoder in all scenarios.
The performance difference is more notable when the number of training data is small.
On the contrary, variational autoencoder outperforms the corresponding $k$-neighbor-encoder; however, the performance of both variational autoencoder and $k$-neighbor-encoder are considerably worse than their vanilla and denoising counterparts.
Overall, both the vanilla and denoising $k$-neighbor-encoders work relatively well for this problem.

\setcounter{subfigure}{0}

Table~\ref{tab_pamap2_kmeans} shows the clustering experiment with $k$-means.
For the vanilla encoder-decoder system, $k$-neighbor-encoder surpasses autoencoder in both scenarios, especially in the noisy scenario.
When the denoising mechanism is added to the encoder-decoder system, it greatly boosts the performance of autoencoders, but the performance of $k$-neighbor-encoder still greatly exceeds autoencoder.  
Similar to the semi-supervised learning experiment, the variational encoder-decoder system performs poorly for this dataset.
In general, both the vanilla and denoising $k$-neighbor-encoders outperform their autoencoder counterparts for the clustering problem on PAMAP2 dataset.

\begin{table}[htb]
\centering
\caption{
The clustering adjust Rand index with $k$-means.
}
\label{tab_pamap2_kmeans}
\begin{tabular}{l|l|ccc}
& & Vanilla & Denoising & Variational \\ \hline
\multirow{2}{*}{Clean} & AE & 0.3815 & 0.4159 & 0.1597 \\
& NE & 0.4203 & 0.4272 & 0.1192 \\ \hline
\multirow{2}{*}{Noisy} & AE & 0.1844 & 0.2336 & 0.1034 \\
& NE & 0.3832 & 0.3948 & 0.1081 \\
\end{tabular}
\end{table}

Figure~\ref{fig_tsne_scatter} further demonstrates the advantage of neighbor-encoder over autoencoder. 
Here we use $t$-SNE to project various representations of the data of subject $1$ into $2D$ space. 
The representations include the raw data itself, the latent representation learned by denoising autoencoder, and the latent representation learned by denoising $k$-neighbor-encoder. 
Despite the clustering experiment suggests that autoencoder is comparable with $k$-neighbor-encoder, we can see that the latent representation learned by $k$-neighbor-encoder provides a much more meaningful visualization of different classes than the rival methods do (includes autoencoder) in the face of noisy/irrelevant dimensions.

\section{Discussion and Conclusions}
\label{conclusion}
In this chapter, we have proposed an unsupervised learning framework called neighbor-encoder that is both \emph{general}, in that it can easily be applied to data in various domains, and \emph{versatile} as it can incorporate domain knowledge by utilizing different neighborhood functions. 
We have showcased the effectiveness of neighbor-encoder compared to autoencoder in various domains, including images, text, time series, and so forth. 
In future work, we plan to either
1) explore the possibility of applying neighbor-encoder to problems like one-shot learning or 
2) demonstrate the usefulness of the neighbor-encoder in more practical and applied tasks, including information retrieval.
We made all the codes/models available at~\cite{nnwebsite}, to allow others to confirm and expand our work. 

\chapter{Conclusions}
\label{ch6}
In this thesis, we have introduced a near universal time series data mining tool called matrix profile~\cite{yeh2016icdm, yeh2017dmkd}.
By storing the solution to the all-pairs-similarity-search problem (of time series subsequences) in a convenient fashion, many time series data mining tasks (e.g., motif discovery, discord discovery shapelet discovery, semantic segmentation, and clustering) can be trivially achieved once such information is pre-computed~\cite{yeh2016icdm, yeh2017dmkd}.
Several matrix profile algorithms, for both single-dimensional and multidimensional time series, are presented in Chapter~\ref{ch2}.
Detailed studies on applying matrix profile on problem such as motif discovery, weakly labeled time series classification, and representation learning are reported in Chapter~\ref{ch3}, Chapter~\ref{ch4}, and Chapter~\ref{ch5} respectively.

This thesis covers only a small part of the matrix profile research as substantial endeavor has already been made on improving various aspects of matrix profile since its original introduction~\cite{dau2017kdd, gharghabi2017icdm, gharghabi2018dmkd, linardi2018matrix, yeh2017vldb, yeh2017icdm, yeh2016icdm2, zhu2017icdm, zhu2016icdm, zhu2018exploiting}.
We expect researchers with diverse backgrounds will adopt the matrix profile in their own research and continue improving upon previous matrix profile findings.
Indeed, there is already evidence of researchers using matrix profile in Arabic dialect identification~\cite{moftah2018arabic}, analysis of geoscience time series~\cite{barbosa2018exploiting}, and music information retrieval~\cite{silva2018summarizing,silva2016ismir,silva2018fast}.
This thesis, which serves as an introductory summary for the matrix profile, is expected to remain relevant for researches interested in the future development of matrix profile and time series data mining.

There are many possibilities for future work and we anticipate that researchers from different research communities will find many uses for (and properties of) the matrix profile that did not occur to us.

\bibliographystyle{plain}
\bibliography{reference_mike}

\bookmarksetupnext{level=part}
\appendix
\chapter{Matrix Profile for Conventional Time Series Classification}
\label{appendixa}
As shown in Chapter~\ref{ch4}, matrix profile can help us solve the \emph{weakly labeled time series classification problem}.
Since matrix profile is a near universal tool for time series data mining, it is unsurprising that matrix profile can also be used to solve the \emph{conventional time series classification problem} (e.g., UCR archive~\cite{ucrarchive}) with simple modification to the Scalable Dictionary learning for Time Series ($SDTS$) algorithm proposed in Chapter~\ref{ch4}.
We call the modified algorithm Scalable Dictionary learning for Conventional Time Series Classification ($SDCTS$).
In this appendix, we will outline the $SDCTS$ algorithm for conventional time series classification problem in Section~\ref{appasecsdcts} and provide the experiment result on UCR archive~\cite{ucrarchive} in Section~\ref{appasecexp}.

\section{The SDCTS Algorithm}
\label{appasecsdcts}
The learning algorithm of the $SDTS$ method can be roughly summarized into three steps:
1) enumerates a large set of shape candidates using matrix profile,
2) selects a smaller set of shape (i.e., the dictionary) from the candidate set, and
3) builds a classifier using the shape dictionary.
When apply the same strategy on conventional time series classification problem, the shapes in the dictionary can be understand as \emph{shapelets}~\cite{mueen2011logical, rakthanmanon2013fast, ye2009time}, and the way we use these shapelets is via \emph{shapelet transformation}~\cite{hills2014classification}.
Since we are striving for a enumerate-and-refine strategy for dictionary building, the ``spirit'' of the method is also similar to \emph{EAST}~\cite{renard2016east}, but instead of randomly enumerate the shape candidates as in EAST~\cite{renard2016east}, we enumerate the shape candidates using the matrix profile\footnote{
We extract both the \emph{motifs} and \emph{discords} from the data to form the initial shape candidate set.}.
The dictionary learning algorithm of $SDTS$ for conventional time series classification (i.e., $SDCTS$) is outlined in Algorithm~\ref{appaalglearndic}.

\begin{algorithm}[htbp]
\caption{Dictionary Learning Algorithm.}
\label{appaalglearndic}
\begin{algorithmic}[1]
\Statex \textbf{Procedure} $Train(X, Y, m)$ \vspace{-1em} 
\Statex \textbf{Input:} Time series dataset $\mathbf{X}$, label $Y$, and subsequence length $m$ \vspace{-1em} 
\Statex \textbf{Output:} Dictionary $\mathbf{S_{refine}}$ and classifier model $C$) \vspace{-1em}
\State $n_{class} \gets GetNumberOfClass(Y)$ \vspace{-1em}
\State $\mathbf{S} \gets \emptyset$ \vspace{-1em}
\For{$i \gets 0$ \textbf{to} $n_{class}$}\vspace{-1em} 
\State $\mathbf{X_i} \gets ExtractExamplesOfAClass(\mathbf{X}, Y, i)$ \vspace{-1em}
\State $T \gets ConcatenateTimeSeries(\mathbf{X_i})$ \vspace{-1em}
\State $P \gets ComputeMP(T, m)$ \Comment{see Chapter~\ref{ch2}}\vspace{-1em} 
\State $\mathbf{S} \gets \mathbf{S}\cup GetMotif(P)$\vspace{-1em} 
\State $\mathbf{S} \gets \mathbf{S}\cup GetDiscord(P)$\vspace{-1em} 
\EndFor \vspace{-1em}
\State $\mathbf{X_{ST}} \gets$ zero matrix of size $|\mathbf{X}| \times 2|\mathbf{S}|$  \vspace{-1em}
\For{$i \gets 0$ \textbf{to} $|\mathbf{X}|$}\vspace{-1em}
\For{$j \gets 0$ \textbf{to} $|\mathbf{S}|$}\vspace{-1em} 
\State $D_{ED}, D_{Corr} \gets MASS(\mathbf{X}[i], \mathbf{S}[j])$ \Comment{see Algorithm~\ref{ch2algmass}}\vspace{-1em} 
\State $\mathbf{X_{ST}}[i, 2j] \gets Max(D_{Corr})$ \vspace{-1em} 
\State $\mathbf{X_{ST}}[i, 2j + 1] \gets Min(D_{ED})$ \vspace{-1em} 
\EndFor \vspace{-1em}
\EndFor \vspace{-1em}
\State $\mathbf{S_{refine}}, C \gets TrainL1RegularizedLinearClassifier(\mathbf{X_{ST}}, Y, \mathbf{S})$  \vspace{-1em}
\State \textbf{return} $\mathbf{S_{refine}}, C$
\end{algorithmic}
\end{algorithm}

In line 1, the number of class for the given dataset $(\mathbf{X}, Y)$ is identified.
Note we assume that the examples are labeled with integer values range from 0 to $n_{class}$ where $n_{class}$ is the number of class in the dataset.
From lines 3 to 9, the initial shape candidate is generated.
In line 4, the time series example associated with the $i$th class is extracted from the dataset and stored in $\mathbf{X_i}$.
Then, in line 5, the time series stored in $\mathbf{X_i}$ are concatenated into a new time series $T$.
In line 6, the matrix profile $P$ is extracted from $T$ (see Chapter~\ref{ch2}).
Similar to Algorithm~\ref{ch4alglearndic}, we exclude nonexistent subsequences (i.e., the subsequence that cross the discontinuity) from the matrix profile computation, and the use of multiple subsequence lengths is also possible.
Next, in lines 7 and 8, the motifs and discords are extracted using the matrix profile, then added to the initial shape candidate set $\mathbf{S}$.

The effectiveness of each shape is determined when used in shapelet transformation~\cite{hills2014classification}.
In order to select a subset of shapes from the candidate shapes, the shapelet transformation is applied to all the time series in the data set with all the candidate shapes.
In lines 10 to 16, such shapelet transformation is performed with the aid of MASS algorithm (see Algorithm~\ref{ch2algmass}).
In line 13, the distance profile using both $z$-normalized Euclidean distance (i.e., $D_{ED}$) and Pearson correlation coefficient (i.e., $D_{Corr}$) are computed using MASS since it is capable of computing both type of distance profile at the same time~\cite{masspage}.
Then, the minimum value of $D_{ED}$ and the maximum value of $D_{Corr}$ are extracted and store in $\mathbf{X_{ST}}$ to complete the shapelet transformation in lines 14 and 15.
Because $D_{Corr}$ contains similarity instead of distance, the maximum value is extracted instead of the minimum value.
Both distance and similarity are extracted because some shapes are better used distance-based shapelet transformation while others are better used in similarity-based shapelet transformation.

Lastly, we perform feature selection on $\mathbf{X_{ST}}$ with a $L1$ regularized linear classification algorithm.
We choose $L1$ regularized linear classification algorithm because this family of classification is capable of doing feature selection and training classification model at the same time.
The particular linear classifier we used is $L1$ regularized linear support vector machine implemented in LIBLINEAR~\cite{fan2008liblinear}.
The use of other classifiers (e.g., $L1$ regularized ridge regression~\cite{fan2008liblinear}) or other feature selection methods (e.g., FRESH algorithm~\cite{christ2018time, christ2016distributed}) is also possible.
To predict the class of a given test example, we simply apply the shapelet transformation using the shapes in $\mathbf{S_{refine}}$, then feed the transformation result to $C$.

\section{Experimental Evaluation}
\label{appasecexp}
We evaluate the proposed $SDCTS$ method on UCR archive~\cite{ucrarchive} against four rival methods: $1NN_{ED}$, $1NN_{DTW\text{-}r}$, $1NN_{DTW}$, and $ST_{random}$, where the first three methods are the standard one-nearest-neighbor baselines\footnote{
one-nearest-neighbor with $z$-normalized euclidean distance, one-nearest-neighbor with $z$-normalized dynamic time warping with warping window~$r$, and one-nearest-neighbor with $z$-normalized dynamic time warping.} for UCR archive~\cite{ucrarchive} while the last one is shapelet transformation with random initial shape candidate set.
The performance of $SDCTS$ comparing to the four rival methods is shown in Table~\ref{appatabsummary}.

\begin{table}[htb]
\centering
\caption{The Performance of four rivals compared to $SDCTS$.}
\begin{tabular}{|c|c|c|c|c|} \hline
& $1NN_{ED}$ & $1NN_{DTW\text{-}r}$ & $1NN_{DTW}$ & $ST_{random}$ \\ \hline
 {\color{red}win} $\mid$ {\color{blue}lose} $\mid$ {\color{green}draw} & \multirow{2}[2]{*}{${\color{red}22} \mid {\color{blue}61} \mid {\color{green}2}$} & \multirow{2}[2]{*}{${\color{red}30} \mid {\color{blue}52} \mid {\color{green}3}$} & \multirow{2}[2]{*}{${\color{red}27} \mid {\color{blue}55} \mid {\color{green}3}$} & \multirow{2}[2]{*}{${\color{red}35} \mid {\color{blue}42} \mid {\color{green}8}$} \\
over $SDCTS$ &       &       &       &  \\ \hline
\end{tabular}
\label{appatabsummary}
\end{table}

According to Table~\ref{appatabsummary}, $SDCTS$ outperforms all the rivals in majority of UCR archive~\cite{ucrarchive} datasets.
$SDCTS$ is not only more powerful than the $1NN$ baselines, but also surpasses the shapelet transformation with random initial shape candidates (i.e., $ST_{random}$).
As the only difference between $SDCTS$ and $ST_{random}$ is the procedural used to generate the initial set of candidate, we can contribute the success of $SDCTS$ to the matrix profile based candidate selection method.
Table~\ref{appataball} shows the full experimental result on UCR archive~\cite{ucrarchive}.
All the codes associated with this section can be downloaded from~\cite{weakwebsite}.

\begin{table}[htbp]
\centering
\caption{Detailed experimental result on UCR archive~\cite{ucrarchive}.}
\begin{adjustbox}{width=\textwidth,totalheight=0.98\textheight,keepaspectratio}
\begin{tabular}{l|cccc|cccc|c}
\multirow{2}[1]{*}{Dataset Name} & Number of & Size of & Size of & Length of & \multirow{2}[1]{*}{$1NN_{ED}$} & \multirow{2}[1]{*}{$1NN_{DTW\text{-}r}$} & \multirow{2}[1]{*}{$1NN_{DTW}$} & \multirow{2}[1]{*}{$ST_{random}$} & \multirow{2}[1]{*}{$SDCTS$} \\
 & Classes & Training & Test & Time Series & & & & & \\  \hline
Synthetic Control & 6     & 300   & 300   & 60    & 0.120 & 0.017 & 0.007 & 0.017 & 0.013 \\
Gun-Point &  2    &  50   &  150  &  150  & 0.087 & 0.087 & 0.093 & 0.053 & 0.013 \\
CBF   &  3    &  30   &  900  &  128  & 0.148 & 0.004 & 0.003 & 0.006 & 0.011 \\
Face (all) &  14   &  560  &  1690 &  131  & 0.286 & 0.192 & 0.192 & 0.221 & 0.221 \\
OSU Leaf &  6    &  200  &  242  &  427  & 0.479 & 0.388 & 0.409 & 0.029 & 0.062 \\
Swedish Leaf &  15   &  500  &  625  &  128  & 0.211 & 0.154 & 0.208 & 0.053 & 0.051 \\
50Words &  50   &  450  &  455  &  270  & 0.369 & 0.242 & 0.310 & 0.244 & 0.251 \\
Trace &  4    &  100  &  100  &  275  & 0.240 & 0.010 & 0.000 & 0.000 & 0.020 \\
Two Patterns &  4    &  1000 &  4000 &  128  & 0.090 & 0.002 & 0.000 & 0.445 & 0.445 \\
Wafer &  2    &  1000 &  6174 &  152  & 0.005 & 0.005 & 0.020 & 0.346 & 0.347 \\
Face (four) &  4    &  24   &  88   &  350  & 0.216 & 0.114 & 0.170 & 0.011 & 0.011 \\
Lightning-2 &  2    &  60   &  61   &  637  & 0.246 & 0.131 & 0.131 & 0.541 & 0.525 \\
Lightning-7 &  7    &  70   &  73   &  319  & 0.425 & 0.288 & 0.274 & 0.712 & 0.726 \\
ECG   & 2     & 100   & 100   & 96    & 0.120 & 0.120 & 0.230 & 0.150 & 0.120 \\
Adiac & 37    & 390   & 391   & 176   & 0.389 & 0.391 & 0.396 & 0.220 & 0.238 \\
Yoga  & 2     & 300   & 3000  & 426   & 0.170 & 0.155 & 0.164 & 0.118 & 0.142 \\
Fish  & 7     & 175   & 175   & 463   & 0.217 & 0.154 & 0.177 & 0.017 & 0.029 \\
Plane & 7     & 105   & 105   & 144   & 0.038 & 0.000 & 0.000 & 0.000 & 0.000 \\
Car   & 4     & 60    & 60    & 577   & 0.267 & 0.233 & 0.267 & 0.150 & 0.150 \\
Beef  & 5     &  30   & 30    & 470   & 0.333 & 0.333 & 0.367 & 0.233 & 0.133 \\
Coffee & 2     & 28    & 28    & 286   & 0.000 & 0.000 & 0.000 & 0.000 & 0.000 \\
OliveOil & 4     & 30    & 30    & 570   & 0.133 & 0.133 & 0.167 & 0.167 & 0.067 \\
CinC\_ECG\_torso & 4     & 40    & 1380  & 1639  & 0.103 & 0.070 & 0.349 & 0.085 & 0.249 \\
ChlorineConcentration & 3     & 467   & 3840  & 166   & 0.350 & 0.350 & 0.352 & 0.247 & 0.222 \\
DiatomSizeReduction & 4     & 16    & 306   & 345   & 0.065 & 0.065 & 0.033 & 0.108 & 0.141 \\
ECGFiveDays & 2     & 23    & 861   & 136   & 0.203 & 0.203 & 0.232 & 0.001 & 0.000 \\
FacesUCR & 14    & 200   & 2050  & 131   & 0.231 & 0.088 & 0.095 & 0.076 & 0.059 \\
Haptics & 5     & 155   & 308   & 1092  & 0.630 & 0.588 & 0.623 & 0.461 & 0.455 \\
InlineSkate & 7     & 100   & 550   & 1882  & 0.658 & 0.613 & 0.616 & 0.545 & 0.529 \\
ItalyPowerDemand & 2     & 67    & 1029  & 24    & 0.045 & 0.045 & 0.050 & 0.049 & 0.040 \\
MALLAT & 8     & 55    & 2345  & 1024  & 0.086 & 0.086 & 0.066 & 0.014 & 0.031 \\
MedicalImages & 10    & 381   & 760   & 99    & 0.316 & 0.253 & 0.263 & 0.308 & 0.312 \\
MoteStrain & 2     & 20    & 1252  & 84    & 0.121 & 0.134 & 0.165 & 0.072 & 0.083 \\
SonyAIBORobot SurfaceII & 2     & 27    & 953   & 65    & 0.141 & 0.141 & 0.169 & 0.083 & 0.112 \\
SonyAIBORobot Surface & 2     & 20    & 601   & 70    & 0.305 & 0.305 & 0.275 & 0.235 & 0.106 \\
StarLightCurves & 3     & 1000  & 8236  & 1024  & 0.151 & 0.095 & 0.093 & 0.024 & 0.023 \\
Symbols & 6     & 25    & 995   & 398   & 0.100 & 0.062 & 0.050 & 0.217 & 0.164 \\
TwoLeadECG & 2     & 23    & 1139  & 82    & 0.253 & 0.132 & 0.096 & 0.002 & 0.003 \\
Cricket\_X & 12    & 390   & 390   & 300   & 0.423 & 0.228 & 0.246 & 0.174 & 0.179 \\
Cricket\_Y & 12    & 390   & 390   & 300   & 0.433 & 0.238 & 0.256 & 0.197 & 0.192 \\
Cricket\_Z & 12    & 390   & 390   & 300   & 0.413 & 0.254 & 0.246 & 0.187 & 0.197 \\
uWaveGestureLibrary\_X & 8     & 896   & 3582  & 315   & 0.261 & 0.227 & 0.273 & 0.696 & 0.702 \\
uWaveGestureLibrary\_Y & 8     & 896   & 3582  & 315   & 0.338 & 0.301 & 0.366 & 0.738 & 0.741 \\
uWaveGestureLibrary\_Z & 8     & 896   & 3582  & 315   & 0.350 & 0.322 & 0.342 & 0.724 & 0.719 \\
Non-Invasive Fetal ECG Thorax1 & 42    & 1800  & 1965  & 750   & 0.171 & 0.185 & 0.209 & 0.060 & 0.055 \\
Non-Invasive Fetal ECG Thorax2 & 42    & 1800  & 1965  & 750   & 0.120 & 0.129 & 0.135 & 0.091 & 0.082 \\
InsectWingbeatSound & 11    & 220   & 1980  & 256   & 0.438 & 0.422 & 0.645 & 0.368 & 0.362 \\
ECG5000 & 5     & 500   & 4500  & 140   & 0.075 & 0.075 & 0.076 & 0.054 & 0.058 \\
ArrowHead & 3     & 36    & 175   & 251   & 0.200 & 0.200 & 0.297 & 0.183 & 0.229 \\
BeetleFly & 2     & 20    & 20    & 512   & 0.250 & 0.300 & 0.300 & 0.100 & 0.150 \\
BirdChicken & 2     & 20    & 20    & 512   & 0.450 & 0.300 & 0.250 & 0.250 & 0.250 \\
Ham   & 2     & 109   & 105   & 431   & 0.400 & 0.400 & 0.533 & 0.276 & 0.248 \\
Herring & 2     & 64    & 64    & 512   & 0.484 & 0.469 & 0.469 & 0.391 & 0.344 \\
PhalangesOutlinesCorrect & 2     & 1800  & 858   & 80    & 0.239 & 0.239 & 0.272 & 0.219 & 0.239 \\
ProximalPhalanxOutlineAgeGroup & 3     & 400   & 205   & 80    & 0.215 & 0.215 & 0.195 & 0.302 & 0.293 \\
ProximalPhalanxOutlineCorrect & 2     & 600   & 291   & 80    & 0.192 & 0.210 & 0.216 & 0.220 & 0.223 \\
ProximalPhalanxTW & 6     & 205   & 400   & 80    & 0.292 & 0.263 & 0.263 & 0.200 & 0.203 \\
ToeSegmentation1 & 2     & 40    & 228   & 277   & 0.320 & 0.250 & 0.228 & 0.039 & 0.110 \\
ToeSegmentation2 & 2     & 36    & 130   & 343   & 0.192 & 0.092 & 0.162 & 0.100 & 0.038 \\
DistalPhalanxOutlineAgeGroup & 3     & 139   & 400   & 80    & 0.218 & 0.228 & 0.208 & 0.373 & 0.335 \\
DistalPhalanxOutlineCorrect & 2     & 276   & 600   & 80    & 0.248 & 0.232 & 0.232 & 0.390 & 0.348 \\
DistalPhalanxTW & 6     & 139   & 400   & 80    & 0.273 & 0.272 & 0.290 & 0.373 & 0.495 \\
Earthquakes & 2     & 139   & 322   & 512   & 0.326 & 0.258 & 0.258 & 0.373 & 0.385 \\
MiddlePhalanxOutlineAgeGroup & 3     & 154   & 400   & 80    & 0.260 & 0.253 & 0.250 & 0.360 & 0.358 \\
MiddlePhalanxOutlineCorrect & 2     & 291   & 600   & 80    & 0.247 & 0.318 & 0.352 & 0.310 & 0.308 \\
MiddlePhalanxTW & 6     & 154   & 399   & 80    & 0.439 & 0.419 & 0.416 & 0.366 & 0.411 \\
ShapeletSim & 2     & 20    & 180   & 500   & 0.461 & 0.300 & 0.350 & 0.000 & 0.000 \\
Wine  & 2     & 57    & 54    & 234   & 0.389 & 0.389 & 0.426 & 0.204 & 0.148 \\
WordSynonyms & 25    & 267   & 638   & 270   & 0.382 & 0.252 & 0.351 & 0.329 & 0.320 \\
Computers & 2     & 250   & 250   & 720   & 0.424 & 0.380 & 0.300 & 0.500 & 0.384 \\
ElectricDevices & 7     & 8926  & 7711  & 96    & 0.450 & 0.376 & 0.399 & 0.518 & 0.534 \\
FordA & 2     & 1320  & 3601  & 500   & 0.341 & 0.341 & 0.438 & 0.066 & 0.071 \\
FordB & 2     & 810   & 3636  & 500   & 0.442 & 0.414 & 0.406 & 0.100 & 0.078 \\
HandOutlines & 2     & 370   & 1000  & 2709  & 0.199 & 0.197 & 0.202 & 0.162 & 0.156 \\
LargeKitchenAppliances & 3     & 375   & 375   & 720   & 0.507 & 0.205 & 0.205 & 0.605 & 0.635 \\
Meat  & 3     & 60    & 60    & 448   & 0.067 & 0.067 & 0.067 & 0.100 & 0.033 \\
Phoneme  & 39    & 214   & 1896  & 1024  & 0.891 & 0.773 & 0.772 & 0.688 & 0.694 \\
RefrigerationDevices & 3     & 375   & 375   & 720   & 0.605 & 0.560 & 0.536 & 0.520 & 0.515 \\
ScreenType & 3     & 375   & 375   & 720   & 0.640 & 0.589 & 0.603 & 0.632 & 0.635 \\
ShapesAll & 60    & 600   & 600   & 512   & 0.248 & 0.198 & 0.232 & 0.108 & 0.103 \\
SmallKitchenAppliances & 3     & 375   & 375   & 720   & 0.659 & 0.328 & 0.357 & 0.488 & 0.480 \\
Strawberry & 2     & 370   & 613   & 235   & 0.062 & 0.062 & 0.060 & 0.033 & 0.033 \\
UWaveGestureLibraryAll & 8     & 896   & 3582  & 945   & 0.052 & 0.034 & 0.108 & 0.676 & 0.672 \\
Worms & 5     & 77    & 181   & 900   & 0.635 & 0.586 & 0.536 & 0.431 & 0.414 \\
WormsTwoClass & 2     & 77    & 181   & 900   & 0.414 & 0.414 & 0.337 & 0.260 & 0.260 \\
\end{tabular}
\end{adjustbox}
\label{appataball}
\end{table}

\end{document}